\documentclass[11pt]{article}

\PassOptionsToPackage{table}{xcolor}

\usepackage[preprint]{acl}

\usepackage{times}
\usepackage{latexsym}

\usepackage[T1]{fontenc}

\usepackage[utf8]{inputenc}

\usepackage{microtype}

\usepackage{inconsolata}

\usepackage{graphicx}
\graphicspath{{./}{../}}
\usepackage[textsize=tiny]{todonotes}
\usepackage{comment}

\usepackage{amsmath}
\usepackage{hyperref}
\usepackage{multirow}
\usepackage{bbold}
\RequirePackage[inline,shortlabels]{enumitem}
\setlist{nosep}
\usepackage{enumitem}
\usepackage{xcolor}
\usepackage{array}
\usepackage{colortbl}
\usepackage{booktabs}
\usepackage{tabularx}

\usepackage{float}
\usepackage{subcaption}
\RequirePackage[inline,shortlabels]{enumitem}
\setlist{nosep}

\definecolor{ECEcol}{RGB}{225,235,255}    
\definecolor{Briercol}{RGB}{255,247,225}  
\newcommand{\xhdr}[1]{\vspace{1mm}\noindent{{\bf #1.}}}
\newcommand{\faintmidrule}{%
  \arrayrulecolor{gray!60}\midrule\arrayrulecolor{black}%
}
\newcommand{\cM}{\mathcal{M}}
\newcommand{\sysname}{{\sc{Calibformer}}}
\newcommand{\Permute}{ICP}
\newcommand{\ICTemp}{TS-AR}
\newcommand{\FSICTemp}{TS-FS}
\newcommand{\ICC}{ICC}

\newcommand{\xtest}{x^*}
\newcommand{\cC}{\mathcal{C}}
\newcommand{\cY}{\mathcal{Y}}
\newcommand{\calP}{\tilde{P}}

\newcommand{\argmax}{\text{argmax}}

\newcommand{\amazonc}{Amazon\_CF}
\newcommand{\commons}{CS\_QA}

\title{Towards Universal Calibration for In-Context Classification}

\author{Awik Dhar \and Sunita Sarawagi \and Arpit Agarwal \\
        Indian Institute of Technology Bombay \\ 
        \texttt{awikdhar@alumni.iitm.ac.in}, \texttt{sunita@iitb.ac.in}, \texttt{aarpit@iitb.ac.in}}

\begin{document}
\maketitle
\begin{abstract}
  In-Context Learning (ICL) has received a lot of interest recently due to its ability to improve classification accuracy for new tasks without having to fine-tune the model over these tasks. However, the impact of ICL on calibration has been far less studied. Recent work has shown that ICL can have a negative impact on calibration due to biased selection of in-context examples. The existing methods for improving ICL calibration require additional task and model-specific training, detracting from the no-fine-tuning convenience of ICL.
In this work, we introduce \sysname, a \emph{universal} ICL calibrator for classification tasks, that generalizes to new models and tasks without further fine-tuning on these tasks. Specifically, we design a transformer-based calibrator which uses the model predictions on in-context examples to predict a confidence score for the test input. We show that the calibrator trained on three models and ten tasks, is able to generalize/calibrate ICL predictions on several new tasks and four LLMs unseen during training.  Even for frontier models like Gemini, \sysname\ is able to obtain over 50\% reduction in calibration error\footnote{\url{https://github.com/AwikDhar/icl-calibration}}.
\end{abstract}

\section{Introduction}
\label{sec:intro}

In-Context Learning (ICL) allows large language models (LLMs) to adapt to new classification tasks at inference time by conditioning on a small set of labeled demonstrations, without any parameter updates \cite{BrownMRSKDNSSAA20}. This capability has found wide application in settings where labeled data is scarce, including question answering \cite{WuWYK23}, low-resource NLP \cite{Alves2023SteeringLL,li-etal-2025-context-learning}, tool calling \cite{zhang2025understanding,dang-etal-2025-improving}, and risk assessment \cite{padayachy2026incontextlearning}.
\begin{figure*}[t]
    \centering
    \begin{subfigure}{0.495\textwidth}
        \centering
        \includegraphics[width=\linewidth]{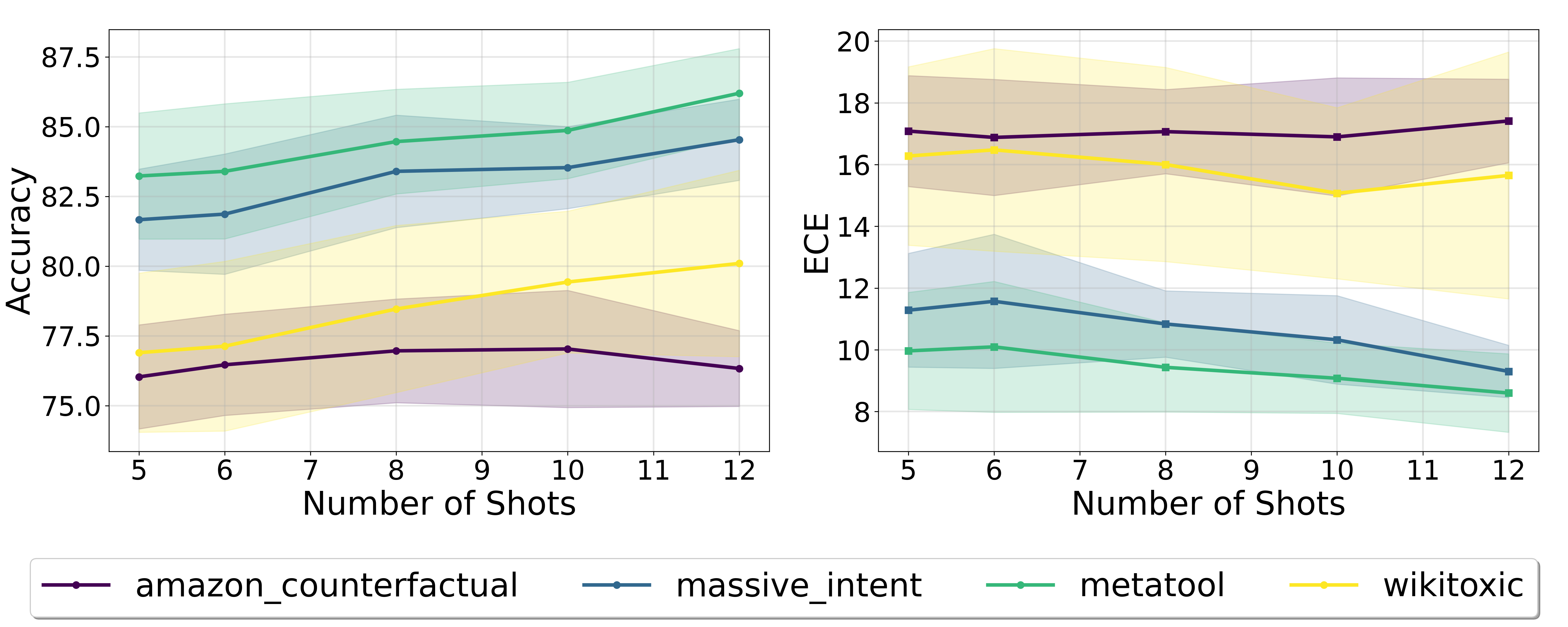}
        \caption{Similarity sampling}
    \end{subfigure}
    \begin{subfigure}{0.495\textwidth}
        \centering
        \includegraphics[width=\linewidth]{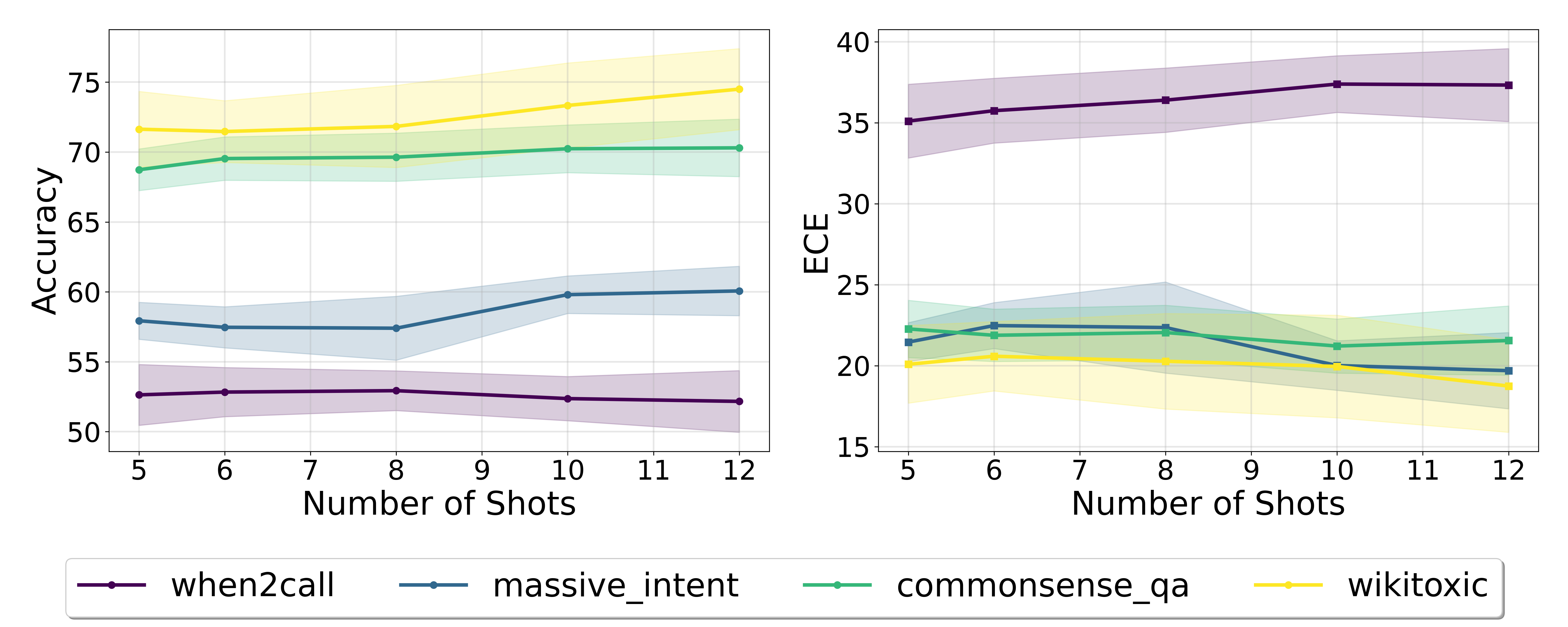}
        \caption{Random sampling}
    \end{subfigure}
    \caption{Accuracy and ECE with varying number of shots. While accuracy typically increases with the number of shots, the model does not significantly improve in aligning its confidence with true likelihood of correctness. Sometimes unpredictable trends are observed - wikitoxic ECE in (a) is turbulent at the end; when2call in (b) experiences worsening ECE despite stable accuracy}
    \label{fig:miscalibration-demonstration}
\vspace{-5mm}
\end{figure*}
However, for ICL to be deployed reliably the model's predicted probabilities must be \emph{calibrated}: a prediction made with 90\% confidence should be correct 90\% of the time.  A well-calibrated predictor enables trust in downstream decision-making as the predicted probabilities provide a good measure of the \emph{confidence} in prediction.
For instance, in risk assessment or sequential reasoning, if the system identifies that it is not confident in its prediction, then it can delegate control to a human expert or invoke external verification tools. 
On the other hand, a miscalibrated predictor can
be detrimental to reliable and safe
decision-making, especially in autonomous or semi-autonomous systems.
Overconfidence in ``agentic'' loops can propagate errors to downstream tasks, leading to compounding failures in autonomous workflows.

Unfortunately, ICL often degrades calibration. The very mechanism that makes ICL effective—conditioning on a small, context-dependent set of demonstrations—introduces variance in the model's confidence that does not track its actual accuracy. As illustrated in Figure~\ref{fig:miscalibration-demonstration}, while accuracy generally improves with additional demonstrations, calibration error behaves erratically, sometimes worsening as more examples are added.

\xhdr{Related Work} Existing approaches to improving ICL classification calibration each carry significant limitations. Bias-correction methods~\cite{Zhao2021CalibrateBU,zhou-etal-2024-batch,Jang2024RectifyingDS-ICC} subtract an estimated label prior from the logits, but they target accuracy rather than calibration, assume randomly sampled demonstrations, and require task-specific estimation. Temperature scaling can be fitted on the in-context examples themselves, but treats predictions at different context positions as independent whereas predictions on ICL examples are correlated. Methods that train auxiliary calibrators \cite{apricot,pmlr-shen-thermo} require retraining for each target LLM, and approaches based on hidden-state features \cite{pmlr-shen-thermo} additionally cannot be applied to black-box models. In all cases, the user must invest effort per model/task, undermining the convenience that makes ICL attractive. See Appendix~\ref{sec:related} for further discussion of related work. 

In this work, we ask: \emph{can a single calibrator, trained once, improve the calibration of any LLM on any classification task, using only the in-context examples?} Towards that goal we develop \sysname, a lightweight (0.61M parameter) transformer-based meta-model that predicts the probability that an LLM's prediction is correct, serving as a calibrated confidence score.

The key idea is that in-context examples constitute a \emph{latent validation set}. As the LLM processes the demonstrations autoregressively, its predictions on these examples reveal the extent and pattern of its miscalibration in the current prompt context. \sysname\ is a causal transformer that attends over a compact set of task-agnostic features derived from these prediction--label discrepancies (top-class probabilities, running correctness, ground-truth probability at each position) and learns to output a calibrated confidence for the test prediction. Because these features capture only the \emph{statistical signature} of the LLM's behavior rather than any task-specific or model-specific representation, a single calibrator trained on $3$ open-weight LLMs and $10$ classification tasks generalizes to unseen models and tasks without further fine-tuning.

We evaluate \sysname\ on eight held-out classification tasks spanning tool use, toxicity detection, intent classification, and medical question answering, using four LLMs not seen during training—including the black-box frontier model Gemini-2.5-flash. Our main findings are:
\begin{enumerate}
    \item \sysname\ reduces average ECE by 41--73\% relative to uncalibrated ICL  (e.g., from 18.3\% to 6.1\% on Phi-4-mini-instruct under similarity sampling; from 12.2\% to 4.5\% on Gemini-2.5-flash under random sampling).
    \item \sysname\ consistently outperforms strong baselines including in-context temperature scaling and context permutation averaging, while requiring only a single forward pass through a sub-million-parameter model.
    \item The calibrator transfers to LLMs with substantially different architectures and confidence profiles, including models that are already reasonably well-calibrated (Qwen-3.5-9B) and black-box models where only token-level log-probabilities are available (Gemini-2.5-flash).
\end{enumerate}


\section{Problem Formulation}
\label{sec:prelims}
We study the calibration of LLM predictions on text classification tasks defined through in-context learning.
A task $\tau$ is specified by $k$ input–output pairs $\mathcal{C}_k = \{(x_i, y_i)\}_{i=1}^{k}$, where each $x_i$ is a text string and $y_i \in \cY_\tau$ is a class label drawn from a task-specific label space.
We assume these examples are sampled from a latent distribution $P_\tau(X,Y)$, with $P_\tau(y \mid x)$ denoting the (unknown) true conditional label distribution.

Given the $k$ demonstrations and a test input $\xtest$, an LLM $\cM$ produces a predictive distribution $P_\cM(y \mid \xtest, \mathcal{C}_{k})$ over $\cY_\tau$.
Section~\ref{sec:llmpd} describes how we extract such a distribution from a generative LLM for an arbitrary label set.
We write $\hat{y} = \argmax_y P_\cM(y \mid \xtest, \mathcal{C}_{k})$ for the predicted label and refer to the corresponding probability $P_\cM(\hat{y} \mid \xtest, \mathcal{C}_{k})$ as the model's \emph{confidence}.

A model is \emph{well-calibrated} when its confidence matches the true likelihood of correctness: among all predictions assigned confidence $p$, a fraction $p$ should be correct.
In practice, modern LLMs tend to be \emph{overconfident}---for instance, among predictions with confidence above 0.9, substantially fewer than 90\% may be correct.

Our goal is to transform the predicted distribution $P_\cM(\cdot \mid \cC_k, \xtest)$ into a better-calibrated distribution $\calP_\cM(\cdot \mid \cC_k, \xtest)$, using \emph{only} in-context examples $\cC_k$ and model's own outputs---without access to additional labeled data from the target task.

\section{Universal Calibrator for ICL}
\label{sec:calibrator}
We propose \sysname, a lightweight transformer-based calibrator that predicts, for any LLM and any classification task, the probability that the LLM's in-context prediction on a test input is correct. This predicted probability serves as the calibrated confidence score. The central design principle is that \sysname\ operates exclusively on \emph{task-agnostic features} derived from the LLM's own output probabilities and the ground-truth labels of the in-context examples, and requires no access to hidden states, attention patterns, or any task-specific metadata. This makes it applicable even to black-box LLMs that expose only token-level log-probabilities.

The key insight underlying our approach is that the in-context examples can be repurposed as a \emph{latent validation set}. As the LLM processes the demonstration sequence autoregressively, its predictions on the in-context examples with known true labels reveal how well-calibrated (or miscalibrated) the model is in the current prompt context. \sysname\ is a causal transformer that attends over this sequence of prediction–label discrepancies and learns to correct the confidence on the subsequent test prediction accordingly. Because the features capture only the \emph{statistical signature} of the LLM's behavior (prediction confidence, ranking among classes, running correctness), a single calibrator trained on a modest collection of models and datasets generalizes to unseen tasks and LLMs without further fine-tuning.  

To train such a calibrator, we leverage a diverse pool of publicly available text classification datasets $\{D_1, \ldots, D_n\}$ spanning tasks such as natural language inference, sentiment analysis, topic classification, and toxicity detection, paired with a set of open-weight LLMs $\{\cM_1, \ldots, \cM_m\}$ that differ in architecture, scale, and confidence profile. This diversity exposes the calibrator to a broad range of miscalibration patterns during training. 
We describe the feature representation (\S\ref{sec:feat}), architecture (\S\ref{sec:arch}), training procedure (\S\ref{sec:training}), 
and the mechanism for extracting label distributions from generative LLMs (\S\ref{sec:llmpd}).

\subsection{Features}
\label{sec:feat}

Since \sysname\ is designed to be agnostic to both the LLM being calibrated and the target task, we restrict its inputs to features that can be computed from any model's output probabilities and the in-context labels alone. In particular, we avoid features that presuppose access to internal representations (hidden states, attention maps) or prior knowledge of the task's difficulty or domain.

Recall that, at each position $t \in \{1, \ldots, k{+}1\}$, the LLM has access to the demonstrations $\cC_{t-1} = \{(x_i, y_i)\}_{i=1}^{t-1}$ and predicts a distribution $P_\cM(y \mid x_t, \cC_{t-1})$ over the label space $\cY_\tau$. From this distribution and the ground-truth label $y_{t-1}$ at the preceding position, we construct a feature vector $\mathbf{f}_t \in \mathbb{R}^d$ with $d=5$:

\begin{enumerate}[leftmargin=*,itemsep=0pt]
\item \textbf{Top-1 predicted probability.}
The confidence assigned to the most likely class:  
\newline
$p^{(1)}_t = P_\cM(y{=} \hat{y}_t \mid x_t, \cC_{t-1})$.

\item \textbf{Top-2 predicted probability.}
The probability of the second most likely class:
\newline
$p^{(2)}_t = \max_{j \neq \hat{y}_t} \; P_\cM(y{=}j \mid x_t, \cC_{t-1})$.

Together, $p^{(1)}_t$ and $p^{(2)}_t$ summarize how peaked or uncertain LLM's prediction is at position $t$.

\item \textbf{Previous correctness indicator.}
A binary signal indicating whether the LLM's prediction at the preceding position matched the ground-truth label: $r_{t} = \mathbb{1}\!\left[\hat{y}_{t-1} = y_{t-1}\right]$, set to $0.5$ at $t=1$.

\item \textbf{Previous ground-truth probability.}
The probability that the LLM assigned to the true label at the preceding position:
$g_{t} = P_\cM(y{=}y_{t-1} \mid x_{t-1}, \cC_{t-2})$,
with dummy value of $0.5$ at $t=1$.

\item \textbf{Previous ground-truth squared loss.}
The squared error on the true-label probability at the preceding position: $\ell_{t} = (1 - g_{t})^2$,
set to $0.25$ at $t=1$ for consistency.

\end{enumerate}

Features 3–5 collectively provide a local measure of how well the LLM is performing on the previous  example and provides the signal  to judge whether the model's confidence at the current position should be trusted. Beyond the necessary confidence and correctness features, the additional three features help in lower shot and high class count regimes where more information may be needed to learn calibration from. Lower ground-truth class probability may be indicative of lower likelihood of correctness on the task.

The full feature vector at each position is the concatenation $\mathbf{f}_t = [p^{(1)}_t,\; p^{(2)}_t,\; r_t,\; g_t,\; \ell_t]$. Note that all five components are bounded in $[0,1]$, and are invariant to the size of the label space properties, which promotes cross-task generalization.

\subsection{\sysname\ Architecture}
\label{sec:arch}

\sysname\ is a causal transformer that maps the feature sequence $\mathbf{F} = (\mathbf{f}_1, \ldots, \mathbf{f}_{k+1}) \in \mathbb{R}^{(k+1) \times d}$ to a calibrated confidence at each position. It consists of a linear input projection, $L{=}12$ transformer layers with hidden dimension $h{=}64$ and $A{=}8$ attention heads, and a scalar sigmoid head---0.61M parameters in total.

The causal attention mask makes the output at position $t$ depend only on $(\mathbf{f}_1, \ldots, \mathbf{f}_t)$, so the calibrator observes ground-truth labels of preceding demonstrations but never the label at position $t$ itself, mirroring the LLM's own autoregressive structure. At the test position $t{=}k{+}1$ it attends to the full demonstration sequence, using cumulative prediction–label discrepancies to adjust the test confidence. We use sinusoidal positional encodings to avoid learning position-specific biases and allow generalization to unseen shot counts.

The final linear+sigmoid head of \sysname\ produces
$\hat{c}_t = \sigma\!\left(\mathrm{TF}_\phi(\mathbf{F})_t\right)$
where $\hat{c}_t \in (0,1)$ is the predicted probability that the LLM's prediction at position $t$ is correct, $\sigma$ is the sigmoid function and $\phi$ are the calibrator's parameters. 

\xhdr{From confidence to a calibrated distribution}
Using $\hat{c}_{k+1}$ as the LLM's calibrated confidence for the test input, we derive a calibrated distribution over all classes by finding the appropriate temperature $T > 0$ to adjust the LLM's original logit vector $\mathbf{z}_{k+1}$ such that:
$\max_{j \in \cY_\tau} \;\mathrm{softmax}\!\left(T \cdot \mathbf{z}_{k+1}\right)_j = \hat{c}_{k+1}$.
The unique $T$ satisfying this equation is found efficiently by binary search, and the resulting tempered softmax yields calibrated probabilities over all classes. Note that this is the reverse of standard temperature scaling: rather than learning a temperature to produce well-calibrated confidences, we predict the calibrated confidence directly and back out the temperature needed to achieve it.  Appendix~\ref{sec:temp-search} provides implementational details and Section~\ref{sec:expts} presents ablations that show that training the transformer to output confidence yielded much better results than training to output temperature.

\subsection{Characterizing Task Coverage}
\begin{figure*}[t]
    \begin{subfigure}{\textwidth}
        \centering
        \includegraphics[width=0.8\linewidth]{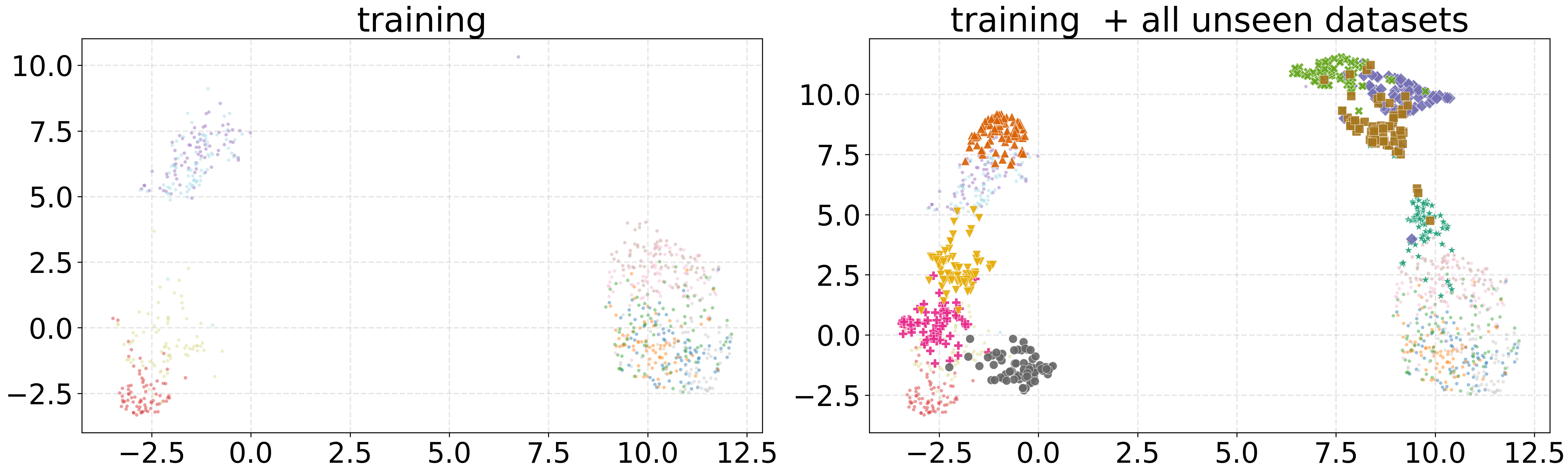}
        \caption{Coverage using partial train data (2 LLMs, 5 datasets)}
    \end{subfigure}
    \vspace{0.5em}
    \begin{subfigure}{\textwidth}
        \centering
        \includegraphics[width=0.8\linewidth]{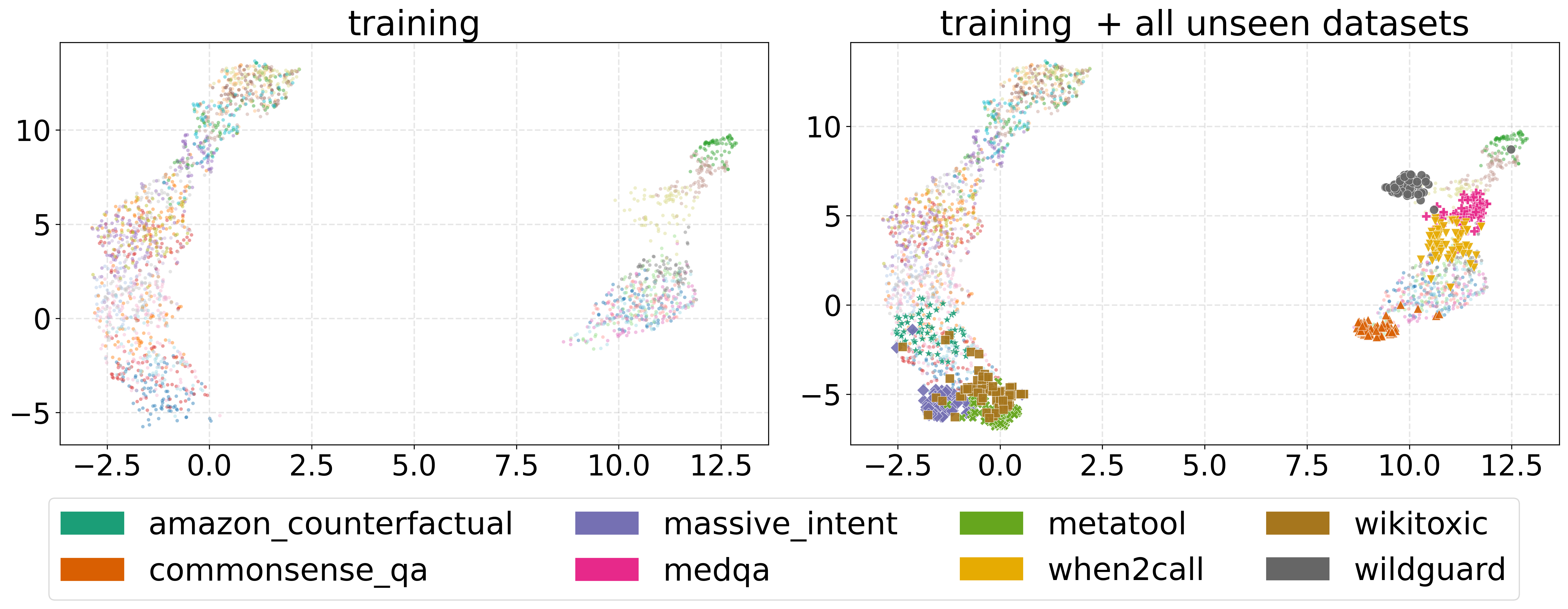}
        \caption{Coverage using full train data (3 LLMs, 10 datasets)}
    \end{subfigure}
    \caption{UMAP projections of embeddings derived from prediction trajectories of different LLM-dataset combinations for Phi-4-mini-instruct. 3 LLMs and 10 datasets achieves adequate coverage for generalizable calibration.}
    \label{fig:coverage}
\vspace{-4mm}
\end{figure*}

To understand why our features can generalize to unseen model-task combinations, we visualize how well the training distribution covers the space of unseen tasks. For each LLM-dataset combination, we sample batches of trajectories from the calibrator training data where each trajectory represents autoregressive predictions over the sequence of inputs (Section ~\ref{para:calib-data-gen}). We represent each batch by its shot-wise mean confidence and mean accuracy---a 2$T$-dimensional vector for a sequence of $T$ inputs---summarizing the miscalibration profile of that LLM-dataset pair. We fit a UMAP jointly on training and unseen embeddings: if unseen points fall within the training region, coverage is adequate. Figure~\ref{fig:coverage} shows that scaling from 2 LLMs $\times$ 5 datasets to 3 LLMs $\times$ 10 datasets substantially improves coverage, suggesting our features capture task-agnostic miscalibration properties that transfer across LLMs and datasets.

\subsection{Training Procedure}
\label{sec:training}
We partition datasets into three disjoint groups---training, validation, evaluation. The training mix spans diverse classification tasks such as NLI, sentiment, toxicity, topic classification, and others; while the evaluation set is held out from both training and checkpoint selection, providing an unbiased generalization test. Each training dataset is further split 80/20 into calibrator-training and calibrator-validation portions.

\xhdr{Training data generation}
\label{para:calib-data-gen}
For each dataset $D$ and model $\cM$ pair, we sample a test instance $(x_{\text{test}}, y_{\text{test}})$ and build a $k$-shot prompt ($k{=}20$) with demonstrations drawn either uniformly at random or by cosine similarity to $\xtest$; similarity-sampled demonstrations are randomly shuffled to add trajectory diversity. The prompt is processed autoregressively through $\cM$, yielding the per-position predictive distribution and the feature sequence $(\mathbf{f}_1, \ldots, \mathbf{f}_{k+1})$ (\S\ref{sec:feat}). We further augment each trajectory as described in Section~\ref{sec:data-aug}. Repeating across all training datasets and source LLMs produces a large pool that varies in label-space size, task difficulty, and model confidence profile.

\xhdr{Training objective}
We train with binary cross-entropy over predicted correctness:
\begin{align*}
\mathcal{L}(\phi) &= \sum_{d \in \mathcal{D}} \sum_{t=t_1}^{k+1} -a_t \log \hat{c}_t - (1 - a_t) \log (1 - \hat{c}_t),
\end{align*}
where $\hat{c}_t = \sigma(\mathrm{TF}_\phi(\mathbf{F})_t)$ and $a_t \in \{0,1\}$ indicates whether the LLM's prediction at position $t$ is correct.  The loss is applied only from $t_1{=}6$ onwards, forcing the calibrator to rely on in-context evidence---the pattern of preceding successes and failures---rather than memorizing task priors in its weights. Without this, the model learns shortcuts that generalize poorly.

\subsubsection{Data Augmentation}
\label{sec:data-aug}

The training LLMs exhibit similar overconfidence patterns, limiting the diversity of calibration regimes the calibrator sees during training. To broaden coverage, we apply two complementary augmentations, sampled at random per trajectory.

\xhdr{Prediction confidence augmentation}
We scale the LLM's logits at every trajectory position by a shared random temperature $T \in [1, T_{\max}]$ and recompute the prediction-derived features in Section ~\ref{sec:feat}. $T_{\max}$ grows with the trajectory's mean confidence so already-confident predictions are perturbed more aggressively, exposing the calibrator to a broader confidence spectrum than the training LLMs alone provide.

\xhdr{Label augmentation}
Even with confidence augmentation, some regimes---already well-calibrated trajectories, or severely miscalibrated ones with high confidence but poor accuracy---remain underrepresented. We replace ground-truth labels with samples from a temperature-scaled version of the LLM's predictive distribution and recompute features 3–5. A mild temperature produces labels mostly agreeing with the model (simulating near-calibrated behavior, teaching the calibrator to leave such sequences undisturbed); a high temperature flattens the distribution, producing labels that contradict the model's confidence and exposing it to severe miscalibration. Temperature schedules for both augmentations are detailed in Appendix~\ref{sec:train-data-aug}.

\subsection{Extracting LLM's Label Distributions}
\label{sec:llmpd}
The features defined in \S\ref{sec:feat} require access to the LLM's predictive distribution $P_\cM(y \mid x_t, \cC_{t-1})$ over the full label set $\cY_\tau$. For single-token labels with distinct first tokens, this distribution can be read directly from the next-token logits. However, many classification tasks have labels that span multiple tokens (e.g., \texttt{card\_in\_transit}) or share a common prefix (e.g., \texttt{card\_arrived} and \texttt{card\_declined}). 
We address this by organizing the label set into a \emph{label token trie}, where each path from the root to a leaf corresponds to the token sequence of one label, and calculating label probabilities via a sequence of forward passes that traverse this trie.
Appendix~\ref{sec:trie} provides a detailed description and example of this structure.

\section{Empirical Evaluation}
\label{sec:expts}
The goal of this evaluation is to test: (a) generalization of our calibrator to unseen model-dataset combinations including black-box models, compared to existing inference-time classification calibration methods (Tables~\ref{tab:open-llms-ece} and~\ref{tab:gemini-ece});
(b) whether the calibration performance is retained when noisy probability estimates are used (Table~\ref{tab:phi4-noisy-preds});
(c) effect of number of shots on calibration performance (Figure~\ref{fig:shotwise_calibration}); and
(d) feature importance, training strategy, and calibrator model choice - tested via ablations (Tables~\ref{tab:ablation} and~\ref{tab:calib_choice}).

\subsection{Baselines}
We compare \sysname\ against inference-time calibration strategies that do not require additional task-specific training. 
We provide a summary here; additional details are in Appendix~\ref{sec:baselines}.

\noindent
\textbf{In-Context Calibration (\ICC)} \cite{Jang2024RectifyingDS-ICC}\label{sec:icc} 
estimates a prompt-induced ``semantic prior'' by predicting on each demonstration while holding it out from the context. This semantic prior is used to scale the logits for inference-time calibration.

\noindent
\textbf{Temperature scaling from autoregressive demonstration predictions (\ICTemp)}\label{sec:ts-ar}
\ICTemp\ fits a single temperature per prompt by minimizing the negative log-likelihood over the autoregressive predictions on the demonstrations.

\noindent
\textbf{Temperature scaling from fixed $t$-shot demonstration predictions (\FSICTemp)}\label{sec:ts-fs}
fits a temperature using $t$-shot predictions for each demonstration (with $t=(k-1)/2$) where the $t$ IC examples are each sampled from the remaining IC examples.

\noindent
\textbf{In-Context Permutation (\Permute)} \cite{jung2025trust}\label{sec:permut}
averages predictions across $N=5$ permutations of the IC examples
to reduce sensitivity to order.

\noindent
\textbf{Discarded baselines.}
We could not compare with two recently proposed recalibration methods since they were either dataset or LLM-specific: (i) surprise-vector methods \cite{tan-etal-2025-surprise}, which require a fixed class dimension and did not generalize when padded across heterogeneous label spaces, and (ii) Thermometer \cite{pmlr-shen-thermo}, which relies on  the specific LLM's hidden states, and is ambiguous for multi-token label scoring.

\subsection{LLMs}
We train a single \sysname\ calibrator on three open-weight LLMs spanning different confidence profiles and instruction-tuning regimes---\texttt{Meta-Llama/Llama-3.1-8B-Instruct} \cite{grattafiori2024llama}, \texttt{Qwen/Qwen3-8B} \cite{qwen3}, and \texttt{OpenAI/Gpt-oss-20b} \cite{openai2025gptoss120bgptoss20bmodel}. We evaluate transfer on four held-out (unseen) LLMs (three open-weight, one API-based): \texttt{Microsoft/Phi-4-mini-instruct} \cite{Abouelenin2025Phi4MiniTR}, \texttt{Google/Gemma-4-E4B-IT} \cite{gemmateam2026gemma4}, \texttt{Qwen/Qwen3.5-9B}, and \texttt{Gemini-2.5-Flash} \cite{gemini25}.



\subsection{Datasets}

We evaluate on held-out (unseen) datasets where calibration matters for downstream decisions:
\begin{enumerate}[leftmargin=*, itemsep=0pt]
\item \textbf{Tool use}: When2call \cite{ross-etal-2025-when2call} and MetaTool \cite{Huang2023MetaToolBF}. Calibrated confidence reduces tool-calling cost by escalating only genuinely uncertain queries.
\item \textbf{Toxicity detection}: Wikitoxic \cite{wikitoxic} and Wildguard \cite{wildguard}. Safety filters trade off catching harmful prompts against refusing benign ones; miscalibration directly impacts this operating point.
\item \textbf{Intent classification}: Massive Intent \cite{fitzgerald-etal-2023-massive}. Calibrated confidence enables reliable fallback (clarification, human hand-off) in intent routing.
\item \textbf{Knowledge, logic, and NLU}: MedQA \cite{medqa}, Commonsense QA (\commons) \cite{talmor-etal-2019-commonsenseqa}, and Amazon Counterfactual (\amazonc) \cite{amazon-counterfactual}. Overconfident errors in high-stakes settings such as medical QA are costly \cite{bioengineering12060631}.
\end{enumerate}
Dataset and prompt details are in Appendix~\ref{sec:datasetdetails}.

\subsection{Experimental Setup}
We provide implementation details in Appendix~\ref{sec:implementation-details}.

\begin{table*}[t]
\centering
\small
\setlength\tabcolsep{2pt}
\resizebox{\textwidth}{!}{
\begin{tabular}{l | cccccc | cccc | cccc}
\toprule
& \multicolumn{6}{c|}{\texttt{Phi-4-mini-instruct}} & \multicolumn{4}{c|}{\texttt{Qwen-3.5-9B}} & \multicolumn{4}{c}{\texttt{Gemma-4-E4B-IT}} \\
\cmidrule(lr){2-7} \cmidrule(lr){8-11} \cmidrule(lr){12-15}
Dataset & Uncalib & \ICC & \Permute & \ICTemp & \FSICTemp & CF(Ours) & Uncalib & \ICTemp & \FSICTemp & CF(Ours) & Uncalib & \ICTemp & \FSICTemp & CF(Ours) \\
\midrule
\multicolumn{15}{c}{\textbf{Random Sampling}} \\
\midrule
\amazonc & 15.3 & 20.6 & 8.4 & 7.8 & \underline{7.6} & \textbf{4.7} & \textbf{4.9} & 7.0 & 7.2 & \underline{5.4} & 14.2 & 5.8 & \underline{5.7} & \textbf{3.4} \\
\commons & 21.8 & 23.7 & 19.3 & 7.9 & \underline{7.4} & \textbf{5.0} & 6.6 & \underline{6.1} & 6.3 & \textbf{5.0} & 20.7 & 7.3 & \underline{7.1} & \textbf{4.9} \\
massive\_intent & 21.2 & 28.9 & 15.3 & \underline{8.8} & 8.9 & \textbf{4.9} & 10.7 & \underline{7.0} & 7.1 & \textbf{4.5} & 17.2 & 8.0 & \underline{7.4} & \textbf{4.1} \\
medqa & 21.8 & 23.9 & 21.2 & \underline{8.6} & 8.8 & \textbf{6.6} & \underline{6.0} & 6.7 & 6.6 & \textbf{5.0} & 28.8 & \textbf{7.2} & \textbf{7.2} & \underline{7.6} \\
metatool & 23.0 & 27.4 & 17.9 & 7.1 & \underline{6.8} & \textbf{5.1} & 12.8 & \underline{6.2} & 6.4 & \textbf{3.9} & 20.0 & \underline{6.9} & \underline{6.9} & \textbf{4.9} \\
when2call & 36.4 & 34.8 & 33.5 & \underline{11.6} & \textbf{11.4} & 13.1 & 14.2 & \underline{7.6} & 8.0 & \textbf{4.7} & 25.6 & \underline{9.0} & \underline{9.0} & \textbf{7.9} \\
wikitoxic & 19.9 & 20.0 & 14.6 & 12.6 & \underline{12.3} & \textbf{9.0} & 10.4 & 7.0 & \underline{6.9} & \textbf{3.4} & 15.2 & 6.6 & \underline{6.5} & \textbf{5.9} \\
wildguard & 46.5 & 43.0 & 41.0 & \textbf{8.1} & 9.5 & \underline{9.3} & 16.3 & \underline{11.1} & 11.5 & \textbf{7.7} & 25.3 & 10.4 & \underline{10.2} & \textbf{5.4} \\
\midrule
Average & 25.7 & 27.8 & 21.4 & \underline{9.1} & \underline{9.1} & \textbf{7.2} & 10.2 & \underline{7.3} & 7.5 & \textbf{4.9} & 20.9 & 7.6 & \underline{7.5} & \textbf{5.5} \\
\midrule
\multicolumn{15}{c}{\textbf{Similarity Sampling}} \\
\midrule
\amazonc & 17.1 & - & 11.2 & 8.6 & \underline{8.2} & \textbf{6.0} & \textbf{4.3} & 7.3 & 7.3 & \underline{6.3} & 18.9 & 6.8 & \textbf{5.9} & \textbf{5.9} \\
\commons & 18.3 & - & 13.6 & \underline{9.7} & 9.9 & \textbf{4.8} & 6.9 & \underline{6.7} & \underline{6.7} & \textbf{4.9} & 19.3 & 7.0 & \underline{6.8} & \textbf{4.1} \\
massive\_intent & 10.7 & - & \underline{6.5} & 7.7 & 8.1 & \textbf{5.2} & 7.1 & \underline{6.8} & 7.2 & \textbf{4.8} & 11.3 & 8.0 & \underline{7.7} & \textbf{3.9} \\
medqa & 17.3 & - & 14.8 & \underline{8.8} & 8.9 & \textbf{5.0} & \textbf{4.4} & 6.3 & 6.5 & \underline{5.6} & 24.7 & 7.7 & \underline{7.5} & \textbf{6.5} \\
metatool & 9.4 & - & 5.9 & 6.0 & \underline{5.7} & \textbf{5.6} & 7.6 & \underline{4.4} & 4.5 & \textbf{3.8} & 11.6 & 5.6 & \underline{5.3} & \textbf{3.4} \\
when2call & 33.1 & - & 28.3 & \textbf{8.0} & \underline{8.5} & 10.5 & 7.2 & \underline{5.1} & 5.5 & \textbf{4.6} & 17.9 & \underline{6.3} & 6.6 & \textbf{4.4} \\
wikitoxic & 15.9 & - & 12.6 & 10.2 & \underline{9.3} & \textbf{6.5} & 10.4 & \underline{7.2} & 7.4 & \textbf{3.4} & 14.8 & 6.4 & \underline{6.3} & \textbf{5.8} \\
wildguard & 24.9 & - & 19.9 & \underline{8.0} & 9.9 & \textbf{4.9} & 18.4 & \underline{9.0} & 9.7 & \textbf{5.9} & 25.1 & \underline{8.0} & 8.6 & \textbf{7.3} \\
\midrule
Average & 18.3 & - & 14.1 & \underline{8.4} & 8.6 & \textbf{6.1} & 8.3 & \underline{6.6} & 6.8 & \textbf{4.9} & 18.0 & 7.0 & \underline{6.8} & \textbf{5.2} \\
\bottomrule
\end{tabular}
}
\caption{Across all three open weights LLMs, \sysname\ outperforms all baselines in 85\% of the cases}
\label{tab:open-llms-ece}
\vspace{-3mm}
\end{table*}

\xhdr{Demonstration selection}
For each test input, we construct a $k$-shot prompt using one of two ways:
\begin{enumerate}[leftmargin=*, itemsep=0pt]
    \item \textbf{Similarity sampling}: choose the $k$ nearest neighbors from the training split based on cosine similarity of $768$-dim embeddings obtained from \texttt{Google/EmbeddingGemma-300m}.
    \item \textbf{Random sampling}: sample $k$ demonstrations uniformly at random from the training split.
\end{enumerate}
\xhdr{Evaluation protocol}
For all open-weight test LLMs, we run 10 random seeds per dataset/setting and evaluate on 300 queries per seed (except for the Massive Intent dataset, where we use 5 seeds due to the higher cost of multi-token label scoring).
For \texttt{Gemini-2.5-flash}, we run 5 seeds and evaluate on 200 queries per seed for $k \in \{5,6,8\}$.
We only report averages across seeds in the main text; the standard errors are reported in Appendix~\ref{sec:add-exp-res}.

\subsection{Evaluation Metric}
We report smooth \emph{Expected Calibration Error (ECE)} \cite{smoothECE}, which kernel-smooths the standard binned ECE over the prediction–label pairs for more reliable estimates (see Appendix~\ref{sec:metrics} for the technical details). We additionally report \emph{Brier score} \cite{brier} in Appendix~\ref{sec:add-exp-res}. 

\begin{table}[t]
\centering
\footnotesize
\setlength\tabcolsep{4.0pt}
\begin{tabular}{l 
>{\columncolor{ECEcol}}c >{\columncolor{Briercol}}c
>{\columncolor{ECEcol}}c >{\columncolor{Briercol}}c
>{\columncolor{ECEcol}}c >{\columncolor{Briercol}}c
>{\columncolor{ECEcol}}c >{\columncolor{Briercol}}c
}
\toprule
& \multicolumn{2}{c}{Uncalib} 
& \multicolumn{2}{c}{\ICTemp} 
& \multicolumn{2}{c}{\FSICTemp} 
& \multicolumn{2}{c}{Ours} \\
\cmidrule(lr){2-3}\cmidrule(lr){4-5}\cmidrule(lr){6-7}\cmidrule(lr){8-9}
Datasets & R & S & R & S & R & S & R & S \\
\midrule
\amazonc               & 7.8  & 9.3  & \underline{4.9} & \underline{4.7} & 5.6 & 5.2 & \textbf{4.5} & \textbf{3.2} \\
\commons               & 11.9 & 11.4 & 8.3  & 8.2  & \underline{7.8} & \underline{7.7} & \textbf{4.5} & \textbf{4.3} \\
medqa                  & 11.3 & 9.0  & 12.3 & 12.7 & \underline{6.2} & \underline{7.0} & \textbf{4.1} & \textbf{4.4} \\
when2call              & 17.7 & 11.0 & 7.4  & \underline{6.3}  & \underline{7.1} & 6.5 & \textbf{4.7} & \textbf{3.9} \\
\midrule
Average                & 12.2 & 10.2 & 8.2  & 8.0  & \underline{6.7} & \underline{6.6} & \textbf{4.5} & \textbf{4.0} \\
\bottomrule
\end{tabular}
\caption{Comparison of ECE calibration on the \texttt{Gemini-2.5-flash} LLM. Values are reported for Random (R) and Similarity (S) sampled contexts. \sysname\ reduces the uncalibrated ECE by over 50\%.}
\label{tab:gemini-ece}
\vspace{-2mm}
\end{table}

\begin{table}[t]
\centering
\footnotesize
\setlength\tabcolsep{2.1pt}
\begin{tabular}{l
>{\columncolor{ECEcol}}c >{\columncolor{Briercol}}c
>{\columncolor{ECEcol}}c >{\columncolor{Briercol}}c
>{\columncolor{ECEcol}}c >{\columncolor{Briercol}}c
>{\columncolor{ECEcol}}c >{\columncolor{Briercol}}c
}
\toprule
& \multicolumn{2}{c}{Uncalib}
& \multicolumn{2}{c}{\ICTemp}
& \multicolumn{2}{c}{\FSICTemp}
& \multicolumn{2}{c}{Ours} \\
\cmidrule(lr){2-3}\cmidrule(lr){4-5}\cmidrule(lr){6-7}\cmidrule(lr){8-9}
Sampling & ECE & Brier & ECE & Brier & ECE & Brier & ECE & Brier \\
\midrule
Similarity & 18.3 & .480 & \underline{11.3} & \underline{.455} & \underline{11.3} & .456 & \textbf{6.6} & \textbf{.433} \\
Random     & 25.8 & .661 & \underline{10.9} & .591 & 11.0 & \underline{.590} & \textbf{7.7} & \textbf{.570} \\
\bottomrule
\end{tabular}
\caption{\sysname\ is robust to noisy probabilities on \texttt{Phi-4-mini-instruct}}
\label{tab:phi4-noisy-preds}
\vspace{-3mm}
\end{table}

\subsection{Main Results}

Tables \ref{tab:open-llms-ece},~\ref{tab:gemini-ece}  and~\ref{tab:phi4-noisy-preds} summarize the ECE comparisons for all the unseen models. 
\sysname\ attains the best average calibration across both sampling strategies and all evaluated LLMs. On \texttt{Phi-4-mini-instruct}, average ECE drops 18.3\% to 6.1\% (similarity) and 25.7\% to 7.2\% (random) (Table~\ref{tab:open-llms-ece}). On \texttt{Gemini-2.5-flash}, average ECE drops 10.2\% to 4.0\% (similarity) and 12.2\% to 4.5\% (random) (Table~\ref{tab:gemini-ece}).
While there are some rare instances where the base model (\texttt{Qwen-3.5-9B}) has better calibration than \sysname, the difference is generally within the standard error (see Appendix~\ref{sec:add-exp-res}).

\xhdr{Comparisons to baselines}
Temperature scaling (\ICTemp/\FSICTemp) is a strong baseline that improves ECE over uncalibrated ICL, but is worse than \sysname\ across datasets and LLMs. Permutation averaging (\Permute) is helpful but costly: it reduces ECE  on average but remains much worse than \sysname\ while requiring $N$ forward passes. \ICC, by contrast, is unstable in our setting — under random sampling on \texttt{Phi-4-mini-instruct} it \emph{worsens} average ECE (25.7\% $\to$ 27.8\%) suggesting that prompt-bias subtraction can be harmful when demonstrations are not representative.
We report \ICC/\Permute\ results only for one model because it is expensive, and did not show gains on small pilots.

\xhdr{Effect of similarity-based demonstrations}
Similarity sampling improves calibration for all methods relative to random sampling (e.g., for \texttt{Phi-4-mini-instruct} uncalibrated ECE drops from 25.7\% to 18.3\%). Nevertheless, even under similarity sampling, the uncalibrated model remains noticeably miscalibrated, and post-hoc calibration remains beneficial.

\xhdr{Effect of noisy or truncated predictions} 
We simulate noise in the LLM label distribution by estimating empirical probabilities based on $N=20$ label samples from the \texttt{Phi-4-mini-instruct} model.
We find that \sysname\ performance is quite robust relative to temperature scaling (Table ~\ref{tab:phi4-noisy-preds}).  
This is because with truncated/noisy probabilities, the ground truth class probability can be too low for some incorrect predictions and more strongly influence the learned temperature scaling.

\begin{figure*}[t]
    \centering 
    \begin{subfigure}{0.495\textwidth}
        \centering
        \includegraphics[width=\linewidth]{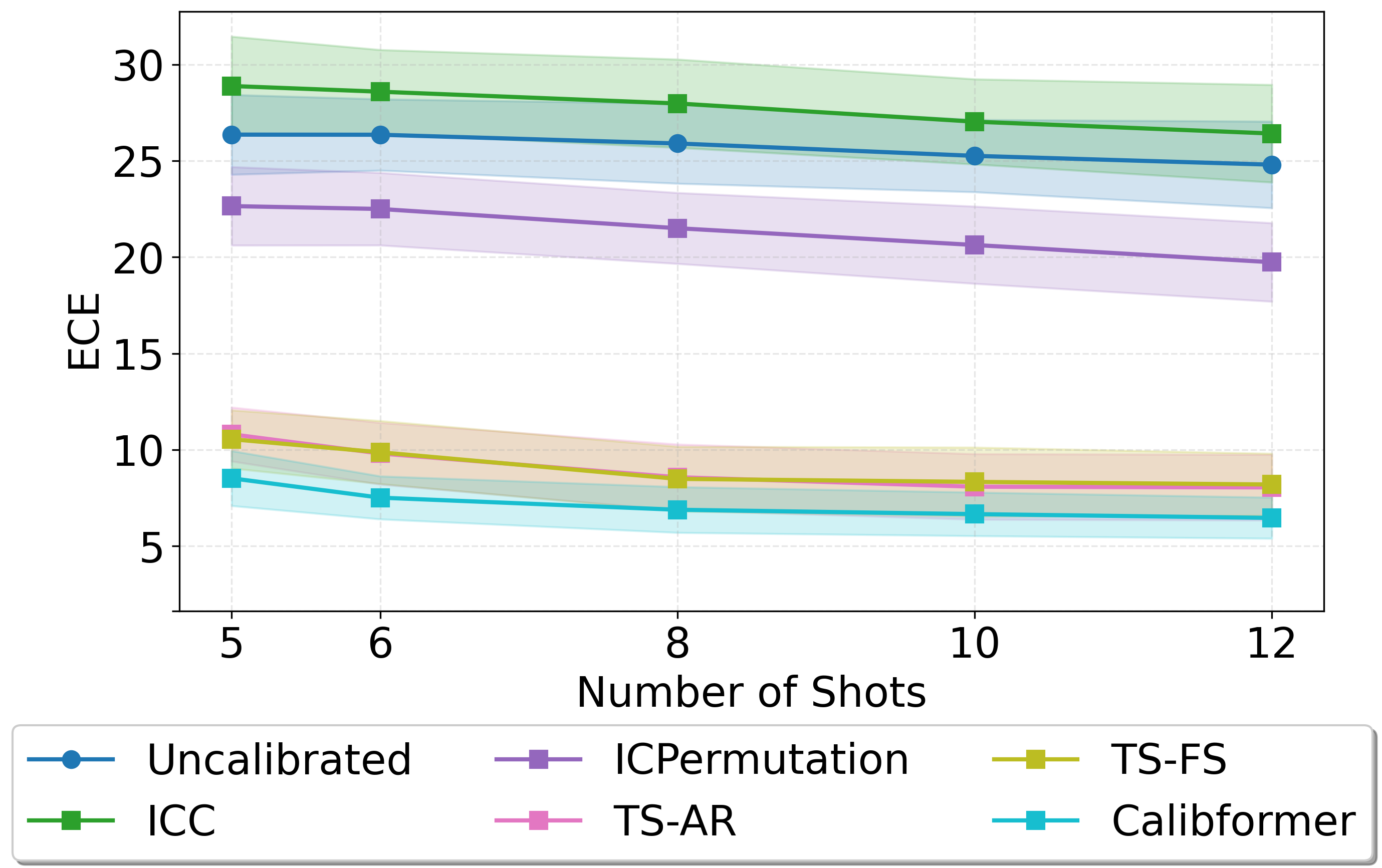}
        \caption{Random sampling}
        \label{fig:random_shotwise}
    \end{subfigure}
    \begin{subfigure}{0.495\textwidth}
        \centering
        \includegraphics[width=\linewidth]{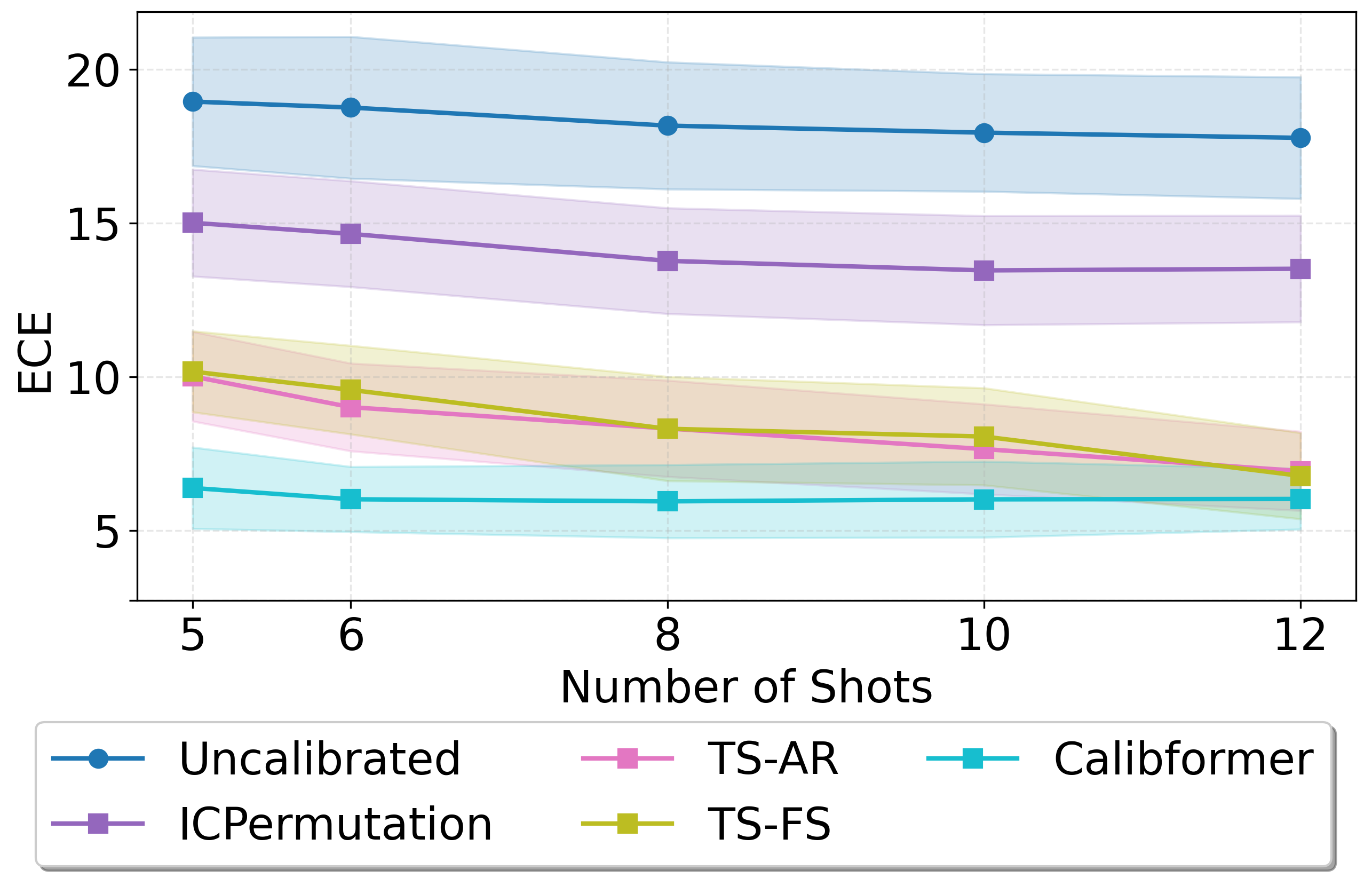}
        \caption{Similarity sampling}
        \label{fig:similarity_shotwise}
    \end{subfigure}
    \caption{For Phi-4-mini-instruct, calibration error tends to on average decrease with shots for all the methods, with \sysname\ consistently outperforming all baselines.}
    \label{fig:shotwise_calibration}
    \vspace{-3mm}
\end{figure*}

\xhdr{Calibration across shots}
Figure~\ref{fig:shotwise_calibration} reports ECE as a function of the number of shots, averaged across datasets (dataset-level plots in the Appendix). Although accuracy typically increases with more demonstrations, uncalibrated ECE only mildly improves, suggesting that additional in-context information primarily improves correctness rather than aligning confidence with correctness likelihood. In contrast, all recalibration methods improve with more shots, indicating that extra demonstrations provide a stronger signal for correcting miscalibration. Across the full range of shots, \sysname\ attains the lowest average ECE.

\begin{table}[h]
\centering
\small
\setlength{\tabcolsep}{4pt}
\begin{tabular}{lcc}

\toprule
Method                & ECE (\%)($\downarrow$) & Brier ($\downarrow$) \\
\midrule
Uncalibrated          & 18.12          & 0.4320           \\
\faintmidrule
Base(\sysname)                 & \textbf{5.41}     & \textbf{0.3807}  \\
\hyperref[itm:conf_correct_only]{Only confidence and correctness}     & \underline{5.47}$^{*}$          & 0.3823$^{***}$      \\
\hyperref[itm:halve_data]{Half Train Data}             & 5.83$^{***}$          & 0.3822$^{***}$    \\
\hyperref[itm:no_embeddings]{w/ embeddings}            & 8.46$^{***}$     & 0.3941$^{***}$    \\
\hyperref[itm:no_embeddings]{w/ permuted embeddings}                 & 5.70$^{***}$ & 0.3816$^{***}$  \\
\hyperref[itm:temp_output]{Temperature Output}   & 7.50$^{***}$          & 0.3837$^{***}$          \\
\hyperref[itm:data_aug]{No data augmentation} & 5.72$^{***}$ & \underline{0.3813}$^{***}$\\

\midrule
\multicolumn{3}{l}{\textbf{Lower shot regime}} \\ %
\midrule
Uncalibrated          & 19.00	& 0.4512           \\
\faintmidrule
Base(\sysname) & \textbf{5.97} & \textbf{0.3979} \\ %
Only confidence and Correctness & 6.10$^{**}$ & 0.4002$^{***}$ \\ %
\midrule

\multicolumn{3}{l}{\textbf{Lower shot and Higher Class Count Regime}} \\ %
\midrule
Uncalibrated          & 21.36	& 0.5016           \\
\faintmidrule
Base(\sysname) & \textbf{5.80} & \textbf{0.4415} \\ %
Only confidence and Correctness & 6.04$^{***}$ & 0.4438$^{***}$ \\ %
\bottomrule
\end{tabular}
\caption{Ablations demonstrating feature set and training strategy importance. While confidence and correctness carry most of the calibration signal, the full feature set provides larger gains in lower shot and higher class count regimes. Stat. significance relative to the Base is denoted by $^{***}$ ($p<.001$),$^{**}$ ($p<.01$) and $^{*}$ ($p<.05$).}
\label{tab:ablation}
\vspace{-3mm}
\end{table}

\begin{table}[ht]
\centering
\begin{tabular}{lc}
\toprule
Method & ECE(\%)($\downarrow$) \\
\midrule
Base(Transformer) & \textbf{5.41} \\
LSTM & 5.45$^{*}$ \\
RNN & 5.51$^{**}$ \\
MLP & 5.81$^{***}$ \\
Logistic Regression & 5.74$^{***}$ \\
Beta Calibration & 8.56$^{***}$ \\
Isotonic Calibration & 11.06$^{***}$ \\
\bottomrule
\vspace{-3mm}
\end{tabular}
\caption{Transformer as the calibrator performs slightly better than equal resource LSTM/RNN and significantly better than simpler approaches, demonstrating efficacy of sequence modeling on in-context predictions. Statistical significance relative to the Base is denoted by $^{***}$ ($p<.001$),$^{**}$ ($p<.01$) and $^{*}$ ($p<.05$).}
\label{tab:calib_choice}
\end{table}

\subsection{Ablations}
We ablate \sysname's design to identify components essential for cross-task, cross-model calibration. 
Tables ~\ref{tab:ablation} and ~\ref{tab:calib_choice} summarize the results, with statistical significance tests relative to the Base \sysname\ method.
Each variant is evaluated on \texttt{Phi-4-mini-instruct}, \texttt{Gemma-4-E4B-it}, and \texttt{Qwen3.5-9B}. Results are averaged over $k\in\{5,\dots,15\}$ shots, all unseen datasets and $30$ seeds.

\xhdr{Impact of feature set}
\phantomsection\label{itm:conf_correct_only}
Learning with just two minimal features---the LLM's predicted confidence and a correctness indicator---still outperforms temperature-scaling baselines, suggesting that a large fraction of the calibration signal is contained in these prediction-derived features. The three additional features provide larger gains in the lower shot regime ($k\in\{3,\dots,10\}$), especially with higher number of classes (5-70), since there is lesser information to learn calibration from.

\xhdr{Impact of training-data diversity}
\phantomsection\label{itm:halve_data}
Using half the training datasets leads to a noticeable increase in ECE, indicating that broad task diversity is important for learning generalizable calibration.


\xhdr{Role of embeddings}
\phantomsection\label{itm:no_embeddings}
We test two variants: concatenating input embeddings directly with base features, and permuting embeddings (via a random order shared across the input sequence) during training to prevent domain-specific memorization while preserving notions of similarity within the sequence. We observe, without permutation, generalization degrades substantially, and with permutation, memorization of embeddings is prevented but calibration is slightly worse than the base model. 

\xhdr{Predicting temperature vs. confidence}
\phantomsection\label{itm:temp_output}
Training \sysname\ to predict a temperature for rescaling logits, rather than calibrated confidence directly, increases ECE---suggesting direct correctness probability modeling is more stable and transferable.

\xhdr{Effect of data augmentation}
\phantomsection\label{itm:data_aug}
The average performance over the LLMs is meaningfully reduced in the absence of data augmentation. This is driven primarily by worse performance on \texttt{Qwen-3.5-9B}, which is reasonably calibrated out of the box.

\xhdr{Transformer as calibrator}
\phantomsection\label{itm:calib_choice}
Using a transformer as the calibrator provides small improvements over other equal resource (0.6M params) sequence models like LSTM and RNN - while all three sequence models significantly outperform simpler approaches like MLP and logistic regression over in-context prediction summary statistics, as well as beta and isotonic calibration. The prediction summary (described in \S~\ref{note:mlp-logistic}) loses the low level sequence information from the in-context examples, and beta/isotonic calibration are sample inefficient relative to temperature scaling. \sysname\ and other sequence models benefit from dedicated training and larger access to in-context information. 

\section{Conclusion}
\label{sec:concl_fut_work}

In this work, we introduced \sysname\, a lightweight universal calibrator for in-context learning classification, that generalizes across models and tasks without requiring task-specific fine-tuning. Our approach leverages a transformer-based architecture that treats in-context examples as a latent validation set, learning to predict calibrated confidence scores by attending to discrepancies between model predictions and ground-truth labels within the context window.
Our calibrator is trained on $3$ models and $10$ tasks, and shows significant reduction in ECE across (unseen) tasks/models.




\section{Limitations}
While we evaluate \sysname\ across a broad range of tasks, models, and seeds, several gaps remain.
First, our evaluation on proprietary models is limited to Gemini-2.5-Flash. Extending the study to more closed-source models would strengthen the paper, but is constrained both by inference cost and by the limited log-probabilities these APIs expose. Empirical logprob estimation via repeated sampling is in principle possible but requires many samples per partial prompt, inflating cost further.

Second, our method targets classification tasks. Extension to open-ended generation is not straightforward: the label space is huge, so the trie-based label-probability construction (\S\ref{sec:llmpd}) can be very costly, and the notion of a per-position ``correctness indicator'' that drives our features must be replaced by a continuous quality signal whose calibration target is itself ill-defined.

Third, in some rare instances where the base model is already well calibrated, our approach can slightly worsen the calibration error. Temperature scaling however suffers from this to a greater extent. Since one does not know a priori if a model will be well calibrated on a task, we believe our method to be a safe default choice. Further, greater data diversity during training and stronger data augmentation (Appendix ~\ref{sec:train-data-aug}) may alleviate this issue.

Finally, our method lacks theoretical guarantees. Although recent work has made substantial progress on understanding ICL for classification, most analyses assume an i.i.d.\ setting. Our calibrator operates on autoregressive demonstration predictions, where features at successive positions are correlated through the shared prompt context---a setting to which existing theory does not directly extend, and we leave this analysis to future work.

\bibliography{icl_calibration}

@String{AMS = "American Mathematical Society" }

@String{Chelsea = "Chelsea" }

@inproceedings{
xiong2024can,
title={Can {LLM}s Express Their Uncertainty? An Empirical Evaluation of Confidence Elicitation in {LLM}s},
author={Miao Xiong and Zhiyuan Hu and Xinyang Lu and YIFEI LI and Jie Fu and Junxian He and Bryan Hooi},
booktitle={The Twelfth International Conference on Learning Representations},
year={2024},
url={https://openreview.net/forum?id=gjeQKFxFpZ}
}

@inproceedings{tian-etal-2023-just,
    title = "Just Ask for Calibration: Strategies for Eliciting Calibrated Confidence Scores from Language Models Fine-Tuned with Human Feedback",
    author = "Tian, Katherine  and
      Mitchell, Eric  and
      Zhou, Allan  and
      Sharma, Archit  and
      Rafailov, Rafael  and
      Yao, Huaxiu  and
      Finn, Chelsea  and
      Manning, Christopher",
    editor = "Bouamor, Houda  and
      Pino, Juan  and
      Bali, Kalika",
    booktitle = "Proceedings of the 2023 Conference on Empirical Methods in Natural Language Processing",
    month = dec,
    year = "2023"
}

@article{Steyvers2024TheCG,
  title={The Calibration Gap between Model and Human Confidence in Large Language Models},
  author={Mark Steyvers and Heliodoro Tejeda Lemus and Aakriti Kumar and Catarina Belem and Sheer Karny and Xinyue Hu and Lukas Mayer and Padhraic Smyth},
  journal={ArXiv},
  year={2024},
  volume={abs/2401.13835},
  url={https://api.semanticscholar.org/CorpusID:267211649}
}

@inproceedings{WuWYK23,
  author       = {Zhiyong Wu and
                  Yaoxiang Wang and
                  Jiacheng Ye and
                  Lingpeng Kong},
  editor       = {Anna Rogers and
                  Jordan L. Boyd{-}Graber and
                  Naoaki Okazaki},
  title        = {Self-Adaptive In-Context Learning: An Information Compression Perspective
                  for In-Context Example Selection and Ordering},
  booktitle    = {Proceedings of the 61st Annual Meeting of the Association for Computational
                  Linguistics (Volume 1: Long Papers), {ACL} 2023, Toronto, Canada,
                  July 9-14, 2023},
  pages        = {1423--1436},
  publisher    = {Association for Computational Linguistics},
  year         = {2023}
}

@inproceedings{BrownMRSKDNSSAA20,
  author       = {Tom B. Brown and
                  Benjamin Mann and
                  Nick Ryder and
                  Melanie Subbiah and
                  Jared Kaplan and
                  Prafulla Dhariwal and
                  Arvind Neelakantan and
                  Pranav Shyam and
                  Girish Sastry and
                  Amanda Askell and
                  Sandhini Agarwal and
                  Ariel Herbert{-}Voss and
                  Gretchen Krueger and
                  Tom Henighan and
                  Rewon Child and
                  Aditya Ramesh and
                  Daniel M. Ziegler and
                  Jeffrey Wu and
                  Clemens Winter and
                  Christopher Hesse and
                  Mark Chen and
                  Eric Sigler and
                  Mateusz Litwin and
                  Scott Gray and
                  Benjamin Chess and
                  Jack Clark and
                  Christopher Berner and
                  Sam McCandlish and
                  Alec Radford and
                  Ilya Sutskever and
                  Dario Amodei},
  editor       = {Hugo Larochelle and
                  Marc'Aurelio Ranzato and
                  Raia Hadsell and
                  Maria{-}Florina Balcan and
                  Hsuan{-}Tien Lin},
  title        = {Language Models are Few-Shot Learners},
  booktitle    = {Advances in Neural Information Processing Systems 33: Annual Conference
                  on Neural Information Processing Systems 2020, NeurIPS 2020, December
                  6-12, 2020, virtual},
  year         = {2020}
}

@inproceedings{Alves2023SteeringLL,
  title={Steering Large Language Models for Machine Translation with Finetuning and In-Context Learning},
  author={Duarte M. Alves and Nuno M. Guerreiro and Joao Alves and Jos{\'e} P. Pombal and Ricardo Rei and Jos{\'e} Guilherme Camargo de Souza and Pierre Colombo and Andr{\'e} Martins},
  booktitle={Conference on Empirical Methods in Natural Language Processing},
  year={2023},
  url={https://api.semanticscholar.org/CorpusID:264405904}
}

@inproceedings{li-etal-2025-context-learning,
    title = "It{'}s All About In-Context Learning! Teaching Extremely Low-Resource Languages to {LLM}s",
    author = "Li, Yue  and
      Zhao, Zhixue  and
      Scarton, Carolina",
    editor = "Christodoulopoulos, Christos  and
      Chakraborty, Tanmoy  and
      Rose, Carolyn  and
      Peng, Violet",
    booktitle = "Proceedings of the 2025 Conference on Empirical Methods in Natural Language Processing",
    month = nov,
    year = "2025",
    address = "Suzhou, China",
    publisher = "Association for Computational Linguistics",
    url = "https://aclanthology.org/2025.emnlp-main.1502/",
    doi = "10.18653/v1/2025.emnlp-main.1502",
    pages = "29532--29547",
    ISBN = "979-8-89176-332-6"
}

@inproceedings{
jung2025trust,
title={Trust or Escalate: {LLM} Judges with Provable Guarantees for Human Agreement},
author={Jaehun Jung and Faeze Brahman and Yejin Choi},
booktitle={The Thirteenth International Conference on Learning Representations},
year={2025},
url={https://openreview.net/forum?id=UHPnqSTBPO}
}

@inproceedings{
zhang2025understanding,
title={Understanding the Generalization of In-Context Learning in Transformers: An Empirical Study},
author={Xingxuan Zhang and Haoran Wang and Jiansheng Li and Yuan Xue and Shikai Guan and Renzhe Xu and Hao Zou and Han Yu and Peng Cui},
booktitle={The Thirteenth International Conference on Learning Representations},
year={2025},
url={https://openreview.net/forum?id=yOhNLIqTEF}
}

@inproceedings{dang-etal-2025-improving,
    title = "Improving Large Language Models Function Calling and Interpretability via Guided-Structured Templates",
    author = "Dang, Hy  and
      Liu, Tianyi  and
      Wu, Zhuofeng  and
      Yang, Jingfeng  and
      Jiang, Haoming  and
      Yang, Tao  and
      Chen, Pei  and
      Wang, Zhengyang  and
      Wang, Helen  and
      Li, Huasheng  and
      Yin, Bing  and
      Jiang, Meng",
    editor = "Christodoulopoulos, Christos  and
      Chakraborty, Tanmoy  and
      Rose, Carolyn  and
      Peng, Violet",
    booktitle = "Proceedings of the 2025 Conference on Empirical Methods in Natural Language Processing",
    month = nov,
    year = "2025",
    address = "Suzhou, China",
    publisher = "Association for Computational Linguistics",
    url = "https://aclanthology.org/2025.emnlp-main.1242/",
    doi = "10.18653/v1/2025.emnlp-main.1242",
    pages = "24426--24442",
    ISBN = "979-8-89176-332-6"
}

@inproceedings{zhang2024study,
  title={A study on the calibration of in-context learning},
  author={Zhang, Hanlin and Zhang, YiFan and Yu, Yaodong and Madeka, Dhruv and Foster, Dean and Xing, Eric and Lakkaraju, Himabindu and Kakade, Sham},
  booktitle={Proceedings of the 2024 Conference of the North American Chapter of the Association for Computational Linguistics: Human Language Technologies (Volume 1: Long Papers)},
  pages={6118--6136},
  year={2024}
}

@misc{padayachy2026incontextlearning,
      title={In-Context Learning Enhanced Credibility Transformer}, 
      author={Kishan Padayachy and Ronald Richman and Salvatore Scognamiglio and Mario V. Wüthrich},
      year={2026},
      eprint={2509.08122},
      archivePrefix={arXiv},
      primaryClass={cs.LG},
      url={https://arxiv.org/abs/2509.08122}, 
}

@inproceedings{NEURIPS2024_b0a4b3e3,
 author = {Cruz, Andr\'{e} F. and Hardt, Moritz and Mendler-D\"{u}nner, Celestine},
 booktitle = {Advances in Neural Information Processing Systems},
 doi = {10.52202/079017-3089},
 editor = {A. Globerson and L. Mackey and D. Belgrave and A. Fan and U. Paquet and J. Tomczak and C. Zhang},
 pages = {97378--97407},
 publisher = {Curran Associates, Inc.},
 title = {Evaluating language models as risk scores},
 url = {https://proceedings.neurips.cc/paper_files/paper/2024/file/b0a4b3e384b4554e65a47ad1f6b0310a-Paper-Datasets_and_Benchmarks_Track.pdf},
 volume = {37},
 year = {2024}
}

@conference{smoothECE,
    author = "Jarosław Błasiok and Preetum Nakkiran",
    booktitle = "",
    title = "Smooth ECE: Principled Reliability Diagrams via Kernel Smoothing",
    year = "2024"
}

@article{brier,
    author = "Glenn W. Brier",
    title = "VERIFICATION OF FORECASTS EXPRESSED IN TERMS OF PROBABILITY",
    journal = "AMS American Metereology Society",
    year = "1950"
}

@inproceedings{tan-etal-2025-surprise,
    title = "Surprise Calibration for Better In-Context Learning",
    author = "Tan, Zhihang  and
      Hou, Jingrui  and
      Wang, Ping  and
      Hu, Qibiao  and
      Zhu, Peng",
    editor = "Christodoulopoulos, Christos  and
      Chakraborty, Tanmoy  and
      Rose, Carolyn  and
      Peng, Violet",
    booktitle = "Proceedings of the 2025 Conference on Empirical Methods in Natural Language Processing",
    month = nov,
    year = "2025",
    address = "Suzhou, China",
    publisher = "Association for Computational Linguistics",
    url = "https://aclanthology.org/2025.emnlp-main.1175/",
    doi = "10.18653/v1/2025.emnlp-main.1175",
    pages = "23056--23071",
    ISBN = "979-8-89176-332-6"
}

@inproceedings{zhou-etal-2024-batch,
    title = "Batch Calibration: Rethinking Calibration for In-Context Learning and Prompt Engineering",
    author = "Zhou, Han and Wan, Xingchen and Proleev, Lev and Mincu, Diana and Chen, Jilin and Heller, Katherine A. and Roy, Subhrajit",
    booktitle = "The Twelfth International Conference on Learning Representations",
    year = "2024",
    url = "https://arxiv.org/abs/2309.17249"
}

@inproceedings{Zhao2021CalibrateBU,
  title={Calibrate Before Use: Improving Few-Shot Performance of Language Models},
  author={Tony Zhao and Eric Wallace and Shi Feng and Dan Klein and Sameer Singh},
  booktitle={International Conference on Machine Learning},
  year={2021},
  url={https://api.semanticscholar.org/CorpusID:231979430}
}

@article{Jang2024RectifyingDS-ICC,
  title={Rectifying Demonstration Shortcut in In-Context Learning},
  author={Joonwon Jang and Sanghwan Jang and Wonbin Kweon and Minjin Jeon and Hwanjo Yu},
  journal={ArXiv},
  year={2024},
  volume={abs/2403.09488},
  url={https://api.semanticscholar.org/CorpusID:268385528}
}

@InProceedings{pmlr-shen-thermo,
  title = 	 {Thermometer: Towards Universal Calibration for Large Language Models},
  author =       {Shen, Maohao and Das, Subhro and Greenewald, Kristjan and Sattigeri, Prasanna and Wornell, Gregory W. and Ghosh, Soumya},
  booktitle = 	 {Proceedings of the 41st International Conference on Machine Learning},
  pages = 	 {44687--44711},
  year = 	 {2024},
  editor = 	 {Salakhutdinov, Ruslan and Kolter, Zico and Heller, Katherine and Weller, Adrian and Oliver, Nuria and Scarlett, Jonathan and Berkenkamp, Felix},
  volume = 	 {235},
  series = 	 {Proceedings of Machine Learning Research},
  month = 	 {21--27 Jul},
  publisher =    {PMLR},
  url = 	 {https://proceedings.mlr.press/v235/shen24c.html}
}

@inproceedings{apricot,
    title = "Calibrating Large Language Models Using Their Generations Only",
    author = "Ulmer, Dennis  and
      Gubri, Martin  and
      Lee, Hwaran  and
      Yun, Sangdoo  and
      Oh, Seong",
    editor = "Ku, Lun-Wei  and
      Martins, Andre  and
      Srikumar, Vivek",
    booktitle = "Proceedings of the 62nd Annual Meeting of the Association for Computational Linguistics (Volume 1: Long Papers)",
    month = aug,
    year = "2024",
    address = "Bangkok, Thailand",
    publisher = "Association for Computational Linguistics",
    url = "https://aclanthology.org/2024.acl-long.824/",
    doi = "10.18653/v1/2024.acl-long.824",
    pages = "15440--15459"
}

@article{singh-transient-icl-emergence,
  title={The Transient Nature of Emergent In-Context Learning in Transformers},
  author={Aaditya K. Singh and Stephanie C. Y. Chan and Ted Moskovitz and Erin Grant and Andrew M. Saxe and Felix Hill},
  journal={ArXiv},
  year={2023},
  volume={abs/2311.08360},
  url={https://api.semanticscholar.org/CorpusID:265157721}
}

@Article{bioengineering12060631,
AUTHOR = {Maity, Subhankar and Saikia, Manob Jyoti},
TITLE = {Large Language Models in Healthcare and Medical Applications: A Review},
JOURNAL = {Bioengineering},
VOLUME = {12},
YEAR = {2025},
NUMBER = {6},
ARTICLE-NUMBER = {631},
URL = {https://www.mdpi.com/2306-5354/12/6/631},
ISSN = {2306-5354},
DOI = {10.3390/bioengineering12060631}
}

@inproceedings{ross-etal-2025-when2call,
    title = "{W}hen2{C}all: When (not) to Call Tools",
    author = "Ross, Hayley  and
      Mahabaleshwarkar, Ameya Sunil  and
      Suhara, Yoshi",
    editor = "Chiruzzo, Luis  and
      Ritter, Alan  and
      Wang, Lu",
    booktitle = "Proceedings of the 2025 Conference of the Nations of the Americas Chapter of the Association for Computational Linguistics: Human Language Technologies (Volume 1: Long Papers)",
    month = apr,
    year = "2025",
    address = "Albuquerque, New Mexico",
    publisher = "Association for Computational Linguistics",
    url = "https://aclanthology.org/2025.naacl-long.174/",
    doi = "10.18653/v1/2025.naacl-long.174",
    pages = "3391--3409",
    ISBN = "979-8-89176-189-6"
}

@article{Huang2023MetaToolBF,
  title={MetaTool Benchmark for Large Language Models: Deciding Whether to Use Tools and Which to Use},
  author={Yue Huang and Jiawen Shi and Yuan Li and Chenrui Fan and Siyuan Wu and Qihui Zhang and Yixin Liu and Pan Zhou and Yao Wan and Neil Zhenqiang Gong and Lichao Sun},
  journal={ArXiv},
  year={2023},
  volume={abs/2310.03128},
  url={https://api.semanticscholar.org/CorpusID:263672025}
}

@unpublished{wikitoxic,
    author = {Thain, Nithum and Dixon Lucas and Wulczyn Ellery},
    title = {Wikipedia Talk Labels: Toxicity},
    year = {2017},
    url = {https://doi.org/10.6084/m9.figshare.4563973.v2}
}

@inproceedings{wildguard,
author = {Han, Seungju and Rao, Kavel and Ettinger, Allyson and Jiang, Liwei and Lin, Bill Yuchen and Lambert, Nathan and Choi, Yejin and Dziri, Nouha},
title = {WILDGUARD: open one-stop moderation tools for safety risks, jailbreaks, and refusals of LLMs},
year = {2024},
isbn = {9798331314385},
publisher = {Curran Associates Inc.},
address = {Red Hook, NY, USA},
booktitle = {Proceedings of the 38th International Conference on Neural Information Processing Systems},
articleno = {261},
numpages = {39},
location = {Vancouver, BC, Canada},
series = {NIPS '24}
}

@inproceedings{fitzgerald-etal-2023-massive,
    title = "{MASSIVE}: A 1{M}-Example Multilingual Natural Language Understanding Dataset with 51 Typologically-Diverse Languages",
    author = "FitzGerald, Jack  and
      Hench, Christopher  and
      Peris, Charith  and
      Mackie, Scott  and
      Rottmann, Kay  and
      Sanchez, Ana  and
      Nash, Aaron  and
      Urbach, Liam  and
      Kakarala, Vishesh  and
      Singh, Richa  and
      Ranganath, Swetha  and
      Crist, Laurie  and
      Britan, Misha  and
      Leeuwis, Wouter  and
      Tur, Gokhan  and
      Natarajan, Prem",
    editor = "Rogers, Anna  and
      Boyd-Graber, Jordan  and
      Okazaki, Naoaki",
    booktitle = "Proceedings of the 61st Annual Meeting of the Association for Computational Linguistics (Volume 1: Long Papers)",
    month = jul,
    year = "2023",
    address = "Toronto, Canada",
    publisher = "Association for Computational Linguistics",
    url = "https://aclanthology.org/2023.acl-long.235/",
    doi = "10.18653/v1/2023.acl-long.235",
    pages = "4277--4302"
}

@inproceedings{amazon-counterfactual,
  address = {Online and Punta Cana, Dominican Republic},
  author = {O{'}Neill, James  and
Rozenshtein, Polina  and
Kiryo, Ryuichi  and
Kubota, Motoko  and
Bollegala, Danushka},
  booktitle = {Proceedings of the 2021 Conference on Empirical Methods in Natural Language Processing},
  doi = {10.18653/v1/2021.emnlp-main.568},
  editor = {Moens, Marie-Francine  and
Huang, Xuanjing  and
Specia, Lucia  and
Yih, Scott Wen-tau},
  month = nov,
  pages = {7092--7108},
  publisher = {Association for Computational Linguistics},
  title = {{I} Wish {I} Would Have Loved This One, But {I} Didn{'}t {--} A Multilingual Dataset for Counterfactual Detection in Product Review},
  url = {https://aclanthology.org/2021.emnlp-main.568},
  year = {2021},
}

@inproceedings{talmor-etal-2019-commonsenseqa,
    title = "{C}ommonsense{QA}: A Question Answering Challenge Targeting Commonsense Knowledge",
    author = "Talmor, Alon  and
      Herzig, Jonathan  and
      Lourie, Nicholas  and
      Berant, Jonathan",
    booktitle = "Proceedings of the 2019 Conference of the North {A}merican Chapter of the Association for Computational Linguistics: Human Language Technologies, Volume 1 (Long and Short Papers)",
    month = jun,
    year = "2019",
    address = "Minneapolis, Minnesota",
    publisher = "Association for Computational Linguistics",
    url = "https://aclanthology.org/N19-1421",
    doi = "10.18653/v1/N19-1421",
    pages = "4149--4158",
    archivePrefix = "arXiv",
    eprint        = "1811.00937",
    primaryClass  = "cs",
}

@article{medqa,
  title={What Disease does this Patient Have? A Large-scale Open Domain Question Answering Dataset from Medical Exams},
  author={Jin, Di and Pan, Eileen and Oufattole, Nassim and Weng, Wei-Hung and Fang, Hanyi and Szolovits, Peter},
  journal={arXiv preprint arXiv:2009.13081},
  year={2020}
}

@article{qwen3,
    title={Qwen3 Technical Report}, 
    author={An Yang and Anfeng Li and Baosong Yang and Beichen Zhang and Binyuan Hui and Bo Zheng and Bowen Yu and Chang Gao and Chengen Huang and Chenxu Lv and Chujie Zheng and Dayiheng Liu and Fan Zhou and Fei Huang and Feng Hu and Hao Ge and Haoran Wei and Huan Lin and Jialong Tang and Jian Yang and Jianhong Tu and Jianwei Zhang and Jianxin Yang and Jiaxi Yang and Jing Zhou and Jingren Zhou and Junyang Lin and Kai Dang and Keqin Bao and Kexin Yang and Le Yu and Lianghao Deng and Mei Li and Mingfeng Xue and Mingze Li and Pei Zhang and Peng Wang and Qin Zhu and Rui Men and Ruize Gao and Shixuan Liu and Shuang Luo and Tianhao Li and Tianyi Tang and Wenbiao Yin and Xingzhang Ren and Xinyu Wang and Xinyu Zhang and Xuancheng Ren and Yang Fan and Yang Su and Yichang Zhang and Yinger Zhang and Yu Wan and Yuqiong Liu and Zekun Wang and Zeyu Cui and Zhenru Zhang and Zhipeng Zhou and Zihan Qiu},
    journal = {arXiv preprint arXiv:2505.09388},
    year={2025}
}

@misc{openai2025gptoss120bgptoss20bmodel,
      title={gpt-oss-120b and gpt-oss-20b Model Card}, 
      author={OpenAI},
      year={2025},
      eprint={2508.10925},
      archivePrefix={arXiv},
      primaryClass={cs.CL},
      url={https://arxiv.org/abs/2508.10925}, 
}

@article{grattafiori2024llama,
  title={The llama 3 herd of models},
  author={Grattafiori, Aaron and Dubey, Abhimanyu and Jauhri, Abhinav and Pandey, Abhinav and Kadian, Abhishek and Al-Dahle, Ahmad and Letman, Aiesha and Mathur, Akhil and Schelten, Alan and Vaughan, Alex and others},
  journal={arXiv preprint arXiv:2407.21783},
  year={2024}
}

@article{Abouelenin2025Phi4MiniTR,
  title={Phi-4-Mini Technical Report: Compact yet Powerful Multimodal Language Models via Mixture-of-LoRAs},
  author={Abdelrahman Abouelenin and Atabak Ashfaq and Adam Atkinson and Hany Hassan Awadalla and Nguyen Bach and Jianmin Bao and Alon Benhaim and Martin Cai and Vishrav Chaudhary and Congcong Chen and Dongdong Chen and Dongdong Chen and Junkun Chen and Weizhu Chen and Yen-Chun Chen and Yi-ling Chen and Qi Dai and Xiyang Dai and Ruchao Fan and Mei Gao and Mingcheng Gao and Amit Garg and Abhishek Goswami and Junheng Hao and Amr Hendy and Yuxuan Hu and Xin Jin and Mahmoud Khademi and Dongwoo Kim and Young Jin Kim and Gina Lee and Jinyu Li and Yunsheng Li and Chen Liang and Xihui Lin and Zeqi Lin and Meng-Jie Liu and Yang Liu and Gilsinia Lopez and Chong Luo and Piyush Madan and Vadim Mazalov and Ali Mousavi and Anh Nguyen and Jing Pan and Daniel Perez-Becker and Jacob Platin and Thomas Portet and Kai Qiu and Bo Ren and Liliang Ren and Sambuddha Roy and Ning Shang and Yelong Shen and Saksham Singhal and Subhojit Som and Xiaocheng Song and Tetyana Sych and Praneetha Vaddamanu and Shuohang Wang and Yiming Wang and Zhenghao Wang and Haibin Wu and Haoran Xu and Weijian Xu and Yifan Yang and Ziyi Yang and Donghan Yu and Ishmam Zabir and Jianwen Zhang and Li Lyna Zhang and Yunan Zhang and Xiren Zhou},
  journal={ArXiv},
  year={2025},
  volume={abs/2503.01743},
  url={https://api.semanticscholar.org/CorpusID:276747153}
}

@misc{gemmateam2026gemma4,
      title={Gemma 4 Technical Report}, 
      author={{Gemma Team}},
      year={2026},
      eprint={2607.02770},
      archivePrefix={arXiv},
      primaryClass={cs.CL},
      url={https://arxiv.org/abs/2607.02770}, 
}

@misc{gemini25,
      title={Gemini 2.5: Pushing the Frontier with Advanced Reasoning, Multimodality, Long Context, and Next Generation Agentic Capabilities}, 
      author={Gheorghe Comanici and Eric Bieber and Mike Schaekermann and Ice Pasupat and Noveen Sachdeva and Inderjit Dhillon and Marcel Blistein and Ori Ram and Dan Zhang and Evan Rosen and Luke Marris and Sam Petulla and Colin Gaffney and Asaf Aharoni and Nathan Lintz and Tiago Cardal Pais and Henrik Jacobsson and Idan Szpektor and Nan-Jiang Jiang and Krishna Haridasan and Ahmed Omran and Nikunj Saunshi and Dara Bahri and Gaurav Mishra and Eric Chu and Toby Boyd and Brad Hekman and Aaron Parisi and Chaoyi Zhang and Kornraphop Kawintiranon and Tania Bedrax-Weiss and Oliver Wang and Ya Xu and Ollie Purkiss and Uri Mendlovic and Ilaï Deutel and Nam Nguyen and Adam Langley and Flip Korn and Lucia Rossazza and Alexandre Ramé and Sagar Waghmare and Helen Miller and Nathan Byrd and Ashrith Sheshan and Raia Hadsell and Sangnie Bhardwaj and Pawel Janus and Tero Rissa and Dan Horgan and Alvin Abdagic and Lior Belenki and James Allingham and Anima Singh and Theo Guidroz and Srivatsan Srinivasan and Herman Schmit and Kristen Chiafullo and Andre Elisseeff and Nilpa Jha and Prateek Kolhar and Leonard Berrada and Frank Ding and Xiance Si and Shrestha Basu Mallick and Franz Och and Sofia Erell and Eric Ni and Tejasi Latkar and Sherry Yang and Petar Sirkovic and Ziqiang Feng and Robert Leland and Rachel Hornung and Gang Wu and Charles Blundell and Hamidreza Alvari and Po-Sen Huang and Cathy Yip and Sanja Deur and Li Liu and Gabriela Surita and Pablo Duque and Dima Damen and Johnson Jia and Arthur Guez and Markus Mircea and Animesh Sinha and Alberto Magni and Paweł Stradomski and Tal Marian and Vlado Galić and Wenhu Chen and Hisham Husain and Achintya Singhal and Dominik Grewe and François-Xavier Aubet and Shuang Song and Lorenzo Blanco and Leland Rechis and Lewis Ho and Rich Munoz and Kelvin Zheng and Jessica Hamrick and Kevin Mather and Hagai Taitelbaum and Eliza Rutherford and Yun Lei and Kuangyuan Chen and Anand Shukla and Erica Moreira and Eric Doi and Berivan Isik and Nir Shabat and Dominika Rogozińska and Kashyap Kolipaka and Jason Chang and Eugen Vušak and Srinivasan Venkatachary and Shadi Noghabi and Tarun Bharti and Younghoon Jun and Aleksandr Zaks and Simon Green and Jeshwanth Challagundla and William Wong and Muqthar Mohammad and Dean Hirsch and Yong Cheng and Iftekhar Naim and Lev Proleev and Damien Vincent and Aayush Singh and Maxim Krikun and Dilip Krishnan and Zoubin Ghahramani and Aviel Atias and Rajeev Aggarwal and Christo Kirov and Dimitrios Vytiniotis and Christy Koh and Alexandra Chronopoulou and Pawan Dogra and Vlad-Doru Ion and Gladys Tyen and Jason Lee and Felix Weissenberger and Trevor Strohman and Ashwin Balakrishna and Jack Rae and Marko Velic and Raoul de Liedekerke and Oded Elyada and Wentao Yuan and Canoee Liu and Lior Shani and Sergey Kishchenko and Bea Alessio and Yandong Li and Richard Song and Sam Kwei and Orion Jankowski and Aneesh Pappu and Youhei Namiki and Yenai Ma and Nilesh Tripuraneni and Colin Cherry and Marissa Ikonomidis and Yu-Cheng Ling and Colin Ji and Beka Westberg and Auriel Wright and Da Yu and David Parkinson and Swaroop Ramaswamy and Jerome Connor and Soheil Hassas Yeganeh and Snchit Grover and George Kenwright and Lubo Litchev and Chris Apps and Alex Tomala and Felix Halim and Alex Castro-Ros and Zefei Li and Anudhyan Boral and Pauline Sho and Michal Yarom and Eric Malmi and David Klinghoffer and Rebecca Lin and Alan Ansell and Pradeep Kumar S and Shubin Zhao and Siqi Zuo and Adam Santoro and Heng-Tze Cheng and Solomon Demmessie and Yuchi Liu and Nicole Brichtova and Allie Culp and Nathaniel Braun and Dan Graur and Will Ng and Nikhil Mehta and Aaron Phillips and Patrik Sundberg and Varun Godbole and Fangyu Liu and Yash Katariya and David Rim and Mojtaba Seyedhosseini and Sean Ammirati and Jonas Valfridsson and Mahan Malihi and Timothy Knight and Andeep Toor and Thomas Lampe and Abe Ittycheriah and Lewis Chiang and Chak Yeung and Alexandre Fréchette and Jinmeng Rao and Huisheng Wang and Himanshu Srivastava and Richard Zhang and Rocky Rhodes and Ariel Brand and Dean Weesner and Ilya Figotin and Felix Gimeno and Rachana Fellinger and Pierre Marcenac and José Leal and Eyal Marcus and Victor Cotruta and Rodrigo Cabrera and Sheryl Luo and Dan Garrette and Vera Axelrod and Sorin Baltateanu and David Barker and Dongkai Chen and Horia Toma and Ben Ingram and Jason Riesa and Chinmay Kulkarni and Yujing Zhang and Hongbin Liu and Chao Wang and Martin Polacek and Will Wu and Kai Hui and Adrian N Reyes and Yi Su and Megan Barnes and Ishaan Malhi and Anfal Siddiqui and Qixuan Feng and Mihai Damaschin and Daniele Pighin and Andreas Steiner and Samuel Yang and Ramya Sree Boppana and Simeon Ivanov and Arun Kandoor and Aditya Shah and Asier Mujika and Da Huang and Christopher A. Choquette-Choo and Mohak Patel and Tianhe Yu and Toni Creswell and Jerry and Liu and Catarina Barros and Yasaman Razeghi and Aurko Roy and Phil Culliton and Binbin Xiong and Jiaqi Pan and Thomas Strohmann and Tolly Powell and Babi Seal and Doug DeCarlo and Pranav Shyam and Kaan Katircioglu and Xuezhi Wang and Cassidy Hardin and Immanuel Odisho and Josef Broder and Oscar Chang and Arun Nair and Artem Shtefan and Maura O'Brien and Manu Agarwal and Sahitya Potluri and Siddharth Goyal and Amit Jhindal and Saksham Thakur and Yury Stuken and James Lyon and Kristina Toutanova and Fangxiaoyu Feng and Austin Wu and Ben Horn and Alek Wang and Alex Cullum and Gabe Taubman and Disha Shrivastava and Chongyang Shi and Hamish Tomlinson and Roma Patel and Tao Tu and Ada Maksutaj Oflazer and Francesco Pongetti and Mingyao Yang and Adrien Ali Taïga and Vincent Perot and Nuo Wang Pierse and Feng Han and Yoel Drori and Iñaki Iturrate and Ayan Chakrabarti and Legg Yeung and Dave Dopson and Yi-ting Chen and Apoorv Kulshreshtha and Tongfei Guo and Philip Pham and Tal Schuster and Junquan Chen and Alex Polozov and Jinwei Xing and Huanjie Zhou and Praneeth Kacham and Doron Kukliansky and Antoine Miech and Sergey Yaroshenko and Ed Chi and Sholto Douglas and Hongliang Fei and Mathieu Blondel and Preethi Myla and Lior Madmoni and Xing Wu and Daniel Keysers and Kristian Kjems and Isabela Albuquerque and Lijun Yu and Joel D'sa and Michelle Plantan and Vlad Ionescu and Jaume Sanchez Elias and Abhirut Gupta and Manish Reddy Vuyyuru and Fred Alcober and Tong Zhou and Kaiyang Ji and Florian Hartmann and Subha Puttagunta and Hugo Song and Ehsan Amid and Anca Stefanoiu and Andrew Lee and Paul Pucciarelli and Emma Wang and Amit Raul and Slav Petrov and Isaac Tian and Valentin Anklin and Nana Nti and Victor Gomes and Max Schumacher and Grace Vesom and Alex Panagopoulos and Konstantinos Bousmalis and Daniel Andor and Josh Jacob and Yuan Zhang and Bill Rosgen and Matija Kecman and Matthew Tung and Alexandra Belias and Noah Goodman and Paul Covington and Brian Wieder and Nikita Saxena and Elnaz Davoodi and Muhuan Huang and Sharath Maddineni and Vincent Roulet and Folawiyo Campbell-Ajala and Pier Giuseppe Sessa and Xintian and Wu and Guangda Lai and Paul Collins and Alex Haig and Vytenis Sakenas and Xiaowei Xu and Marissa Giustina and Laurent El Shafey and Pichi Charoenpanit and Shefali Garg and Joshua Ainslie and Boone Severson and Montse Gonzalez Arenas and Shreya Pathak and Sujee Rajayogam and Jie Feng and Michiel Bakker and Sheng Li and Nevan Wichers and Jamie Rogers and Xinyang Geng and Yeqing Li and Rolf Jagerman and Chao Jia and Nadav Olmert and David Sharon and Matthew Mauger and Sandeep Mariserla and Hongxu Ma and Megha Mohabey and Kyuyeun Kim and Alek Andreev and Scott Pollom and Juliette Love and Vihan Jain and Priyanka Agrawal and Yannick Schroecker and Alisa Fortin and Manfred Warmuth and Ji Liu and Andrew Leach and Irina Blok and Ganesh Poomal Girirajan and Roee Aharoni and Benigno Uria and Andrei Sozanschi and Dan Goldberg and Lucian Ionita and Marco Tulio Ribeiro and Martin Zlocha and Vighnesh Birodkar and Sami Lachgar and Liangzhe Yuan and Himadri Choudhury and Matt Ginsberg and Fei Zheng and Gregory Dibb and Emily Graves and Swachhand Lokhande and Gabriel Rasskin and George-Cristian Muraru and Corbin Quick and Sandeep Tata and Pierre Sermanet and Aditya Chawla and Itay Karo and Yan Wang and Susan Zhang and Orgad Keller and Anca Dragan and Guolong Su and Ian Chou and Xi Liu and Yiqing Tao and Shruthi Prabhakara and Marc Wilson and Ruibo Liu and Shibo Wang and Georgie Evans and David Du and Alfonso Castaño and Gautam Prasad and Mona El Mahdy and Sebastian Gerlach and Machel Reid and Jarrod Kahn and Amir Zait and Thanumalayan Sankaranarayana Pillai and Thatcher Ulrich and Guanyu Wang and Jan Wassenberg and Efrat Farkash and Kiran Yalasangi and Congchao Wang and Maria Bauza and Simon Bucher and Ting Liu and Jun Yan and Gary Leung and Vikas Sindhwani and Parker Barnes and Avi Singh and Ivan Jurin and Jichuan Chang and Niket Kumar Bhumihar and Sivan Eiger and Gui Citovsky and Ben Withbroe and Zhang Li and Siyang Xue and Niccolò Dal Santo and Georgi Stoyanov and Yves Raimond and Steven Zheng and Yilin Gao and Vít Listík and Sławek Kwasiborski and Rachel Saputro and Adnan Ozturel and Ganesh Mallya and Kushal Majmundar and Ross West and Paul Caron and Jinliang Wei and Lluis Castrejon and Sharad Vikram and Deepak Ramachandran and Nikhil Dhawan and Jiho Park and Sara Smoot and George van den Driessche and Yochai Blau and Chase Malik and Wei Liang and Roy Hirsch and Cicero Nogueira dos Santos and Eugene Weinstein and Aäron van den Oord and Sid Lall and Nicholas FitzGerald and Zixuan Jiang and Xuan Yang and Dale Webster and Ali Elqursh and Aedan Pope and Georges Rotival and David Raposo and Wanzheng Zhu and Jeff Dean and Sami Alabed and Dustin Tran and Arushi Gupta and Zach Gleicher and Jessica Austin and Edouard Rosseel and Megh Umekar and Dipanjan Das and Yinghao Sun and Kai Chen and Karolis Misiunas and Xiang Zhou and Yixian Di and Alyssa Loo and Josh Newlan and Bo Li and Vinay Ramasesh and Ying Xu and Alex Chen and Sudeep Gandhe and Radu Soricut and Nikita Gupta and Shuguang Hu and Seliem El-Sayed and Xavier Garcia and Idan Brusilovsky and Pu-Chin Chen and Andrew Bolt and Lu Huang and Alex Gurney and Zhiying Zhang and Alexander Pritzel and Jarek Wilkiewicz and Bryan Seybold and Bhargav Kanagal Shamanna and Felix Fischer and Josef Dean and Karan Gill and Ross Mcilroy and Abhishek Bhowmick and Jeremy Selier and Antoine Yang and Derek Cheng and Vladimir Magay and Jie Tan and Dhriti Varma and Christian Walder and Tomas Kocisky and Ryo Nakashima and Paul Natsev and Mike Kwong and Ionel Gog and Chiyuan Zhang and Sander Dieleman and Thomas Jimma and Andrey Ryabtsev and Siddhartha Brahma and David Steiner and Dayou Du and Ante Žužul and Mislav Žanić and Mukund Raghavachari and Willi Gierke and Zeyu Zheng and Dessie Petrova and Yann Dauphin and Yuchuan Liu and Ido Kessler and Steven Hand and Chris Duvarney and Seokhwan Kim and Hyo Lee and Léonard Hussenot and Jeffrey Hui and Josh Smith and Deepali Jain and Jiawei Xia and Gaurav Singh Tomar and Keyvan Amiri and Du Phan and Fabian Fuchs and Tobias Weyand and Nenad Tomasev and Alexandra Cordell and Xin Liu and Jonathan Mallinson and Pankaj Joshi and Andy Crawford and Arun Suggala and Steve Chien and Nick Fernando and Mariella Sanchez-Vargas and Duncan Williams and Phil Crone and Xiyang Luo and Igor Karpov and Jyn Shan and Terry Thurk and Robin Strudel and Paul Voigtlaender and Piyush Patil and Tim Dozat and Ali Khodaei and Sahil Singla and Piotr Ambroszczyk and Qiyin Wu and Yifan Chang and Brian Roark and Chaitra Hegde and Tianli Ding and Angelos Filos and Zhongru Wu and André Susano Pinto and Shuang Liu and Saarthak Khanna and Aditya Pandey and Siobhan Mcloughlin and Qiujia Li and Sam Haves and Allan Zhou and Elena Buchatskaya and Isabel Leal and Peter de Boursac and Nami Akazawa and Nina Anderson and Terry Chen and Krishna Somandepalli and Chen Liang and Sheela Goenka and Stephanie Winkler and Alexander Grushetsky and Yifan Ding and Jamie Smith and Fan Ye and Jordi Pont-Tuset and Eric Li and Ruichao Li and Tomer Golany and Dawid Wegner and Tao Jiang and Omer Barak and Yuan Shangguan and Eszter Vértes and Renee Wong and Jörg Bornschein and Alex Tudor and Michele Bevilacqua and Tom Schaul and Ankit Singh Rawat and Yang Zhao and Kyriakos Axiotis and Lei Meng and Cory McLean and Jonathan Lai and Jennifer Beattie and Nate Kushman and Yaxin Liu and Blair Kutzman and Fiona Lang and Jingchen Ye and Praneeth Netrapalli and Pushkar Mishra and Myriam Khan and Megha Goel and Rob Willoughby and David Tian and Honglei Zhuang and JD Chen and Zak Tsai and Tasos Kementsietsidis and Arjun Khare and James Keeling and Keyang Xu and Nathan Waters and Florent Altché and Ashok Popat and Bhavishya Mittal and David Saxton and Dalia El Badawy and Michael Mathieu and Zheng Zheng and Hao Zhou and Nishant Ranka and Richard Shin and Qingnan Duan and Tim Salimans and Ioana Mihailescu and Uri Shaham and Ming-Wei Chang and Yannis Assael and Nishanth Dikkala and Martin Izzard and Vincent Cohen-Addad and Cat Graves and Vlad Feinberg and Grace Chung and DJ Strouse and Danny Karmon and Sahand Sharifzadeh and Zoe Ashwood and Khiem Pham and Jon Blanton and Alex Vasiloff and Jarred Barber and Mark Geller and Aurick Zhou and Fedir Zubach and Tzu-Kuo Huang and Lei Zhang and Himanshu Gupta and Matt Young and Julia Proskurnia and Ronny Votel and Valentin Gabeur and Gabriel Barcik and Aditya Tripathi and Hongkun Yu and Geng Yan and Beer Changpinyo and Filip Pavetić and Amy Coyle and Yasuhisa Fujii and Jorge Gonzalez Mendez and Tianhao Zhou and Harish Rajamani and Blake Hechtman and Eddie Cao and Da-Cheng Juan and Yi-Xuan Tan and Valentin Dalibard and Yilun Du and Natalie Clay and Kaisheng Yao and Wenhao Jia and Dimple Vijaykumar and Yuxiang Zhou and Xinyi Bai and Wei-Chih Hung and Steven Pecht and Georgi Todorov and Nikhil Khadke and Pramod Gupta and Preethi Lahoti and Arnaud Autef and Karthik Duddu and James Lee-Thorp and Alexander Bykovsky and Tautvydas Misiunas and Sebastian Flennerhag and Santhosh Thangaraj and Jed McGiffin and Zack Nado and Markus Kunesch and Andreas Noever and Amir Hertz and Marco Liang and Victor Stone and Evan Palmer and Samira Daruki and Arijit Pramanik and Siim Põder and Austin Kyker and Mina Khan and Evgeny Sluzhaev and Marvin Ritter and Avraham Ruderman and Wenlei Zhou and Chirag Nagpal and Kiran Vodrahalli and George Necula and Paul Barham and Ellie Pavlick and Jay Hartford and Izhak Shafran and Long Zhao and Maciej Mikuła and Tom Eccles and Hidetoshi Shimokawa and Kanav Garg and Luke Vilnis and Hanwen Chen and Ilia Shumailov and Kuang-Huei Lee and Abdelrahman Abdelhamed and Meiyan Xie and Vered Cohen and Ester Hlavnova and Dan Malkin and Chawin Sitawarin and James Lottes and Pauline Coquinot and Tianli Yu and Sandeep Kumar and Jingwei Zhang and Aroma Mahendru and Zafarali Ahmed and James Martens and Tao Chen and Aviel Boag and Daiyi Peng and Coline Devin and Arseniy Klimovskiy and Mary Phuong and Danny Vainstein and Jin Xie and Bhuvana Ramabhadran and Nathan Howard and Xinxin Yu and Gitartha Goswami and Jingyu Cui and Sam Shleifer and Mario Pinto and Chih-Kuan Yeh and Ming-Hsuan Yang and Sara Javanmardi and Dan Ethier and Chace Lee and Jordi Orbay and Suyog Kotecha and Carla Bromberg and Pete Shaw and James Thornton and Adi Gerzi Rosenthal and Shane Gu and Matt Thomas and Ian Gemp and Aditya Ayyar and Asahi Ushio and Aarush Selvan and Joel Wee and Chenxi Liu and Maryam Majzoubi and Weiren Yu and Jake Abernethy and Tyler Liechty and Renke Pan and Hoang Nguyen and Qiong and Hu and Sarah Perrin and Abhinav Arora and Emily Pitler and Weiyi Wang and Kaushik Shivakumar and Flavien Prost and Ben Limonchik and Jing Wang and Yi Gao and Timothee Cour and Shyamal Buch and Huan Gui and Maria Ivanova and Philipp Neubeck and Kelvin Chan and Lucy Kim and Huizhong Chen and Naman Goyal and Da-Woon Chung and Lu Liu and Yao Su and Anastasia Petrushkina and Jiajun Shen and Armand Joulin and Yuanzhong Xu and Stein Xudong Lin and Yana Kulizhskaya and Ciprian Chelba and Shobha Vasudevan and Eli Collins and Vasilisa Bashlovkina and Tony Lu and Doug Fritz and Jongbin Park and Yanqi Zhou and Chen Su and Richard Tanburn and Mikhail Sushkov and Mitchelle Rasquinha and Jinning Li and Jennifer Prendki and Yiming Li and Pallavi LV and Shriya Sharma and Hen Fitoussi and Hui Huang and Andrew Dai and Phuong Dao and Mike Burrows and Henry Prior and Danfeng Qin and Golan Pundak and Lars Lowe Sjoesund and Art Khurshudov and Zhenkai Zhu and Albert Webson and Elizabeth Kemp and Tat Tan and Saurabh Agrawal and Susie Sargsyan and Liqun Cheng and Jim Stephan and Tom Kwiatkowski and David Reid and Arunkumar Byravan and Assaf Hurwitz Michaely and Nicolas Heess and Luowei Zhou and Sonam Goenka and Viral Carpenter and Anselm Levskaya and Bo Wang and Reed Roberts and Rémi Leblond and Sharat Chikkerur and Stav Ginzburg and Max Chang and Robert Riachi and Chuqiao and Xu and Zalán Borsos and Michael Pliskin and Julia Pawar and Morgane Lustman and Hannah Kirkwood and Ankit Anand and Aditi Chaudhary and Norbert Kalb and Kieran Milan and Sean Augenstein and Anna Goldie and Laurel Prince and Karthik Raman and Yanhua Sun and Vivian Xia and Aaron Cohen and Zhouyuan Huo and Josh Camp and Seher Ellis and Lukas Zilka and David Vilar Torres and Lisa Patel and Sho Arora and Betty Chan and Jonas Adler and Kareem Ayoub and Jacky Liang and Fayaz Jamil and Jiepu Jiang and Simon Baumgartner and Haitian Sun and Yael Karov and Yaroslav Akulov and Hui Zheng and Irene Cai and Claudio Fantacci and James Rubin and Alex Rav Acha and Mengchao Wang and Nina D'Souza and Rohit Sathyanarayana and Shengyang Dai and Simon Rowe and Andrey Simanovsky and Omer Goldman and Yuheng Kuang and Xiaoyue Pan and Andrew Rosenberg and Tania Rojas-Esponda and Praneet Dutta and Amy Zeng and Irina Jurenka and Greg Farquhar and Yamini Bansal and Shariq Iqbal and Becca Roelofs and Ga-Young Joung and Parker Beak and Changwan Ryu and Ryan Poplin and Yan Wu and Jean-Baptiste Alayrac and Senaka Buthpitiya and Olaf Ronneberger and Caleb Habtegebriel and Wei Li and Paul Cavallaro and Aurora Wei and Guy Bensky and Timo Denk and Harish Ganapathy and Jeff Stanway and Pratik Joshi and Francesco Bertolini and Jessica Lo and Olivia Ma and Zachary Charles and Geta Sampemane and Himanshu Sahni and Xu Chen and Harry Askham and David Gaddy and Peter Young and Jiewen Tan and Matan Eyal and Arthur Bražinskas and Li Zhong and Zhichun Wu and Mark Epstein and Kai Bailey and Andrew Hard and Kamyu Lee and Sasha Goldshtein and Alex Ruiz and Mohammed Badawi and Matthias Lochbrunner and JK Kearns and Ashley Brown and Fabio Pardo and Theophane Weber and Haichuan Yang and Pan-Pan Jiang and Berkin Akin and Zhao Fu and Marcus Wainwright and Chi Zou and Meenu Gaba and Pierre-Antoine Manzagol and Wendy Kan and Yang Song and Karina Zainullina and Rui Lin and Jeongwoo Ko and Salil Deshmukh and Apoorv Jindal and James Svensson and Divya Tyam and Heri Zhao and Christine Kaeser-Chen and Scott Baird and Pooya Moradi and Jamie Hall and Qiuchen Guo and Vincent Tsang and Bowen Liang and Fernando Pereira and Suhas Ganesh and Ivan Korotkov and Jakub Adamek and Sridhar Thiagarajan and Vinh Tran and Charles Chen and Chris Tar and Sanil Jain and Ishita Dasgupta and Taylan Bilal and David Reitter and Kai Zhao and Giulia Vezzani and Yasmin Gehman and Pulkit Mehta and Lauren Beltrone and Xerxes Dotiwalla and Sergio Guadarrama and Zaheer Abbas and Stefani Karp and Petko Georgiev and Chun-Sung Ferng and Marc Brockschmidt and Liqian Peng and Christoph Hirnschall and Vikas Verma and Yingying Bi and Ying Xiao and Avigail Dabush and Kelvin Xu and Phil Wallis and Randall Parker and Qifei Wang and Yang Xu and Ilkin Safarli and Dinesh Tewari and Yin Zhang and Seungyeon Kim and Andrea Gesmundo and Mackenzie Thomas and Sergey Levi and Ahmed Chowdhury and Kanishka Rao and Peter Garst and Sam Conway-Rahman and Helen Ran and Kay McKinney and Zhisheng Xiao and Wenhao Yu and Rohan Agrawal and Axel Stjerngren and Catalin Ionescu and Jingjing Chen and Vivek Sharma and Justin Chiu and Fei Liu and Ken Franko and Clayton Sanford and Xingyu Cai and Paul Michel and Sanjay Ganapathy and Jane Labanowski and Zachary Garrett and Ben Vargas and Sean Sun and Bryan Gale and Thomas Buschmann and Guillaume Desjardins and Nimesh Ghelani and Palak Jain and Mudit Verma and Chulayuth Asawaroengchai and Julian Eisenschlos and Jitendra Harlalka and Hideto Kazawa and Don Metzler and Joshua Howland and Ying Jian and Jake Ades and Viral Shah and Tynan Gangwani and Seungji Lee and Roman Ring and Steven M. Hernandez and Dean Reich and Amer Sinha and Ashutosh Sathe and Joe Kovac and Ashleah Gill and Ajay Kannan and Andrea D'olimpio and Martin Sevenich and Jay Whang and Been Kim and Khe Chai Sim and Jilin Chen and Jiageng Zhang and Shuba Lall and Yossi Matias and Bill Jia and Abe Friesen and Sara Nasso and Ashish Thapliyal and Bryan Perozzi and Ting Yu and Anna Shekhawat and Safeen Huda and Peter Grabowski and Eric Wang and Ashwin Sreevatsa and Hilal Dib and Mehadi Hassen and Parker Schuh and Vedrana Milutinovic and Chris Welty and Michael Quinn and Ali Shah and Bangju Wang and Gabe Barth-Maron and Justin Frye and Natalie Axelsson and Tao Zhu and Yukun Ma and Irene Giannoumis and Hanie Sedghi and Chang Ye and Yi Luan and Kevin Aydin and Bilva Chandra and Vivek Sampathkumar and Ronny Huang and Victor Lavrenko and Ahmed Eleryan and Zhi Hong and Steven Hansen and Sara Mc Carthy and Bidisha Samanta and Domagoj Ćevid and Xin Wang and Fangtao Li and Michael Voznesensky and Matt Hoffman and Andreas Terzis and Vikash Sehwag and Gil Fidel and Luheng He and Mu Cai and Yanzhang He and Alex Feng and Martin Nikoltchev and Samrat Phatale and Jason Chase and Rory Lawton and Ming Zhang and Tom Ouyang and Manuel Tragut and Mehdi Hafezi Manshadi and Arjun Narayanan and Jiaming Shen and Xu Gao and Tolga Bolukbasi and Nick Roy and Xin Li and Daniel Golovin and Liviu Panait and Zhen Qin and Guangxing Han and Thomas Anthony and Sneha Kudugunta and Viorica Patraucean and Aniket Ray and Xinyun Chen and Xiaochen Yang and Tanuj Bhatia and Pranav Talluri and Alex Morris and Andrija Ražnatović and Bethanie Brownfield and James An and Sheng Peng and Patrick Kane and Ce Zheng and Nico Duduta and Joshua Kessinger and James Noraky and Siqi Liu and Keran Rong and Petar Veličković and Keith Rush and Alex Goldin and Fanny Wei and Shiva Mohan Reddy Garlapati and Caroline Pantofaru and Okwan Kwon and Jianmo Ni and Eric Noland and Julia Di Trapani and Françoise Beaufays and Abhijit Guha Roy and Yinlam Chow and Aybuke Turker and Geoffrey Cideron and Lantao Mei and Jon Clark and Qingyun Dou and Matko Bošnjak and Ralph Leith and Yuqing Du and Amir Yazdanbakhsh and Milad Nasr and Chester Kwak and Suraj Satishkumar Sheth and Alex Kaskasoli and Ankesh Anand and Balaji Lakshminarayanan and Sammy Jerome and David Bieber and Chun-Te Chu and Alexandre Senges and Tianxiao Shen and Mukund Sridhar and Ndaba Ndebele and Benjamin Beyret and Shakir Mohamed and Mia Chen and Markus Freitag and Jiaxian Guo and Luyang Liu and Paul Roit and Heng Chen and Shen Yan and Tom Stone and JD Co-Reyes and Jeremy Cole and Salvatore Scellato and Shekoofeh Azizi and Hadi Hashemi and Alicia Jin and Anand Iyer and Marcella Valentine and András György and Arun Ahuja and Daniel Hernandez Diaz and Chen-Yu Lee and Nathan Clement and Weize Kong and Drew Garmon and Ishaan Watts and Kush Bhatia and Khyatti Gupta and Matt Miecnikowski and Hugo Vallet and Ankur Taly and Edward Loper and Saket Joshi and James Atwood and Jo Chick and Mark Collier and Fotis Iliopoulos and Ryan Trostle and Beliz Gunel and Ramiro Leal-Cavazos and Arnar Mar Hrafnkelsson and Michael Guzman and Xiaoen Ju and Andy Forbes and Jesse Emond and Kushal Chauhan and Ben Caine and Li Xiao and Wenjun Zeng and Alexandre Moufarek and Daniel Murphy and Maya Meng and Nitish Gupta and Felix Riedel and Anil Das and Elijah Lawal and Shashi Narayan and Tiberiu Sosea and James Swirhun and Linda Friso and Behnam Neyshabur and Jing Lu and Sertan Girgin and Michael Wunder and Edouard Yvinec and Aroonalok Pyne and Victor Carbune and Shruti Rijhwani and Yang Guo and Tulsee Doshi and Anton Briukhov and Max Bain and Ayal Hitron and Xuanhui Wang and Ashish Gupta and Ke Chen and Cosmo Du and Weiyang Zhang and Dhruv Shah and Arjun Akula and Max Dylla and Ashyana Kachra and Weicheng Kuo and Tingting Zou and Lily Wang and Luyao Xu and Jifan Zhu and Justin Snyder and Sachit Menon and Orhan Firat and Igor Mordatch and Yuan Yuan and Natalia Ponomareva and Rory Blevins and Lawrence Moore and Weijun Wang and Phil Chen and Martin Scholz and Artur Dwornik and Jason Lin and Sicheng Li and Diego Antognini and Te I and Xiaodan Song and Matt Miller and Uday Kalra and Adam Raveret and Oscar Akerlund and Felix Wu and Andrew Nystrom and Namrata Godbole and Tianqi Liu and Hannah DeBalsi and Jewel Zhao and Buhuang Liu and Avi Caciularu and Lauren Lax and Urvashi Khandelwal and Victoria Langston and Eric Bailey and Silvio Lattanzi and Yufei Wang and Neel Kovelamudi and Sneha Mondal and Guru Guruganesh and Nan Hua and Ofir Roval and Paweł Wesołowski and Rishikesh Ingale and Jonathan Halcrow and Tim Sohn and Christof Angermueller and Bahram Raad and Eli Stickgold and Eva Lu and Alec Kosik and Jing Xie and Timothy Lillicrap and Austin Huang and Lydia Lihui Zhang and Dominik Paulus and Clement Farabet and Alex Wertheim and Bing Wang and Rishabh Joshi and Chu-ling Ko and Yonghui Wu and Shubham Agrawal and Lily Lin and XiangHai Sheng and Peter Sung and Tyler Breland-King and Christina Butterfield and Swapnil Gawde and Sumeet Singh and Qiao Zhang and Raj Apte and Shilpa Shetty and Adrian Hutter and Tao Li and Elizabeth Salesky and Federico Lebron and Jonni Kanerva and Michela Paganini and Arthur Nguyen and Rohith Vallu and Jan-Thorsten Peter and Sarmishta Velury and David Kao and Jay Hoover and Anna Bortsova and Colton Bishop and Shoshana Jakobovits and Alessandro Agostini and Alekh Agarwal and Chang Liu and Charles Kwong and Sasan Tavakkol and Ioana Bica and Alex Greve and Anirudh GP and Jake Marcus and Le Hou and Tom Duerig and Rivka Moroshko and Dave Lacey and Andy Davis and Julien Amelot and Guohui Wang and Frank Kim and Theofilos Strinopoulos and Hui Wan and Charline Le Lan and Shankar Krishnan and Haotian Tang and Peter Humphreys and Junwen Bai and Idan Heimlich Shtacher and Diego Machado and Chenxi Pang and Ken Burke and Dangyi Liu and Renga Aravamudhan and Yue Song and Ed Hirst and Abhimanyu Singh and Brendan Jou and Liang Bai and Francesco Piccinno and Chuyuan Kelly Fu and Robin Alazard and Barak Meiri and Daniel Winter and Charlie Chen and Mingda Zhang and Jens Heitkaemper and John Lambert and Jinhyuk Lee and Alexander Frömmgen and Sergey Rogulenko and Pranav Nair and Paul Niemczyk and Anton Bulyenov and Bibo Xu and Hadar Shemtov and Morteza Zadimoghaddam and Serge Toropov and Mateo Wirth and Hanjun Dai and Sreenivas Gollapudi and Daniel Zheng and Alex Kurakin and Chansoo Lee and Kalesha Bullard and Nicolas Serrano and Ivana Balazevic and Yang Li and Johan Schalkwyk and Mark Murphy and Mingyang Zhang and Kevin Sequeira and Romina Datta and Nishant Agrawal and Charles Sutton and Nithya Attaluri and Mencher Chiang and Wael Farhan and Gregory Thornton and Kate Lin and Travis Choma and Hung Nguyen and Kingshuk Dasgupta and Dirk Robinson and Iulia Comşa and Michael Riley and Arjun Pillai and Basil Mustafa and Ben Golan and Amir Zandieh and Jean-Baptiste Lespiau and Billy Porter and David Ross and Sujeevan Rajayogam and Mohit Agarwal and Subhashini Venugopalan and Bobak Shahriari and Qiqi Yan and Hao Xu and Taylor Tobin and Pavel Dubov and Hongzhi Shi and Adrià Recasens and Anton Kovsharov and Sebastian Borgeaud and Lucio Dery and Shanthal Vasanth and Elena Gribovskaya and Linhai Qiu and Mahdis Mahdieh and Wojtek Skut and Elizabeth Nielsen and CJ Zheng and Adams Yu and Carrie Grimes Bostock and Shaleen Gupta and Aaron Archer and Chris Rawles and Elinor Davies and Alexey Svyatkovskiy and Tomy Tsai and Yoni Halpern and Christian Reisswig and Bartek Wydrowski and Bo Chang and Joan Puigcerver and Mor Hazan Taege and Jian Li and Eva Schnider and Xinjian Li and Dragos Dena and Yunhan Xu and Umesh Telang and Tianze Shi and Heiga Zen and Kyle Kastner and Yeongil Ko and Neesha Subramaniam and Aviral Kumar and Pete Blois and Zhuyun Dai and John Wieting and Yifeng Lu and Yoel Zeldes and Tian Xie and Anja Hauth and Alexandru Ţifrea and Yuqi Li and Sam El-Husseini and Dan Abolafia and Howard Zhou and Wen Ding and Sahra Ghalebikesabi and Carlos Guía and Andrii Maksai and Ágoston Weisz and Sercan Arik and Nick Sukhanov and Aga Świetlik and Xuhui Jia and Luo Yu and Weiyue Wang and Mark Brand and Dawn Bloxwich and Sean Kirmani and Zhe Chen and Alec Go and Pablo Sprechmann and Nithish Kannen and Alen Carin and Paramjit Sandhu and Isabel Edkins and Leslie Nooteboom and Jai Gupta and Loren Maggiore and Javad Azizi and Yael Pritch and Pengcheng Yin and Mansi Gupta and Danny Tarlow and Duncan Smith and Desi Ivanov and Mohammad Babaeizadeh and Ankita Goel and Satish Kambala and Grace Chu and Matej Kastelic and Michelle Liu and Hagen Soltau and Austin Stone and Shivani Agrawal and Min Kim and Kedar Soparkar and Srinivas Tadepalli and Oskar Bunyan and Rachel Soh and Arvind Kannan and DY Kim and Blake JianHang Chen and Afief Halumi and Sudeshna Roy and Yulong Wang and Olcan Sercinoglu and Gena Gibson and Sijal Bhatnagar and Motoki Sano and Daniel von Dincklage and Qingchun Ren and Blagoj Mitrevski and Mirek Olšák and Jennifer She and Carl Doersch and Jilei and Wang and Bingyuan Liu and Qijun Tan and Tamar Yakar and Tris Warkentin and Alex Ramirez and Carl Lebsack and Josh Dillon and Rajiv Mathews and Tom Cobley and Zelin Wu and Zhuoyuan Chen and Jon Simon and Swaroop Nath and Tara Sainath and Alexei Bendebury and Ryan Julian and Bharath Mankalale and Daria Ćurko and Paulo Zacchello and Adam R. Brown and Kiranbir Sodhia and Heidi Howard and Sergi Caelles and Abhinav Gupta and Gareth Evans and Anna Bulanova and Lesley Katzen and Roman Goldenberg and Anton Tsitsulin and Joe Stanton and Benoit Schillings and Vitaly Kovalev and Corey Fry and Rushin Shah and Kuo Lin and Shyam Upadhyay and Cheng Li and Soroush Radpour and Marcello Maggioni and Jing Xiong and Lukas Haas and Jenny Brennan and Aishwarya Kamath and Nikolay Savinov and Arsha Nagrani and Trevor Yacovone and Ryan Kappedal and Kostas Andriopoulos and Li Lao and YaGuang Li and Grigory Rozhdestvenskiy and Kazuma Hashimoto and Andrew Audibert and Sophia Austin and Daniel Rodriguez and Anian Ruoss and Garrett Honke and Deep Karkhanis and Xi Xiong and Qing Wei and James Huang and Zhaoqi Leng and Vittal Premachandran and Stan Bileschi and Georgios Evangelopoulos and Thomas Mensink and Jay Pavagadhi and Denis Teplyashin and Paul Chang and Linting Xue and Garrett Tanzer and Sally Goldman and Kaushal Patel and Shixin Li and Jeremy Wiesner and Ivy Zheng and Ian Stewart-Binks and Jie Han and Zhi Li and Liangchen Luo and Karel Lenc and Mario Lučić and Fuzhao Xue and Ryan Mullins and Alexey Guseynov and Chung-Ching Chang and Isaac Galatzer-Levy and Adam Zhang and Garrett Bingham and Grace Hu and Ale Hartman and Yue Ma and Jordan Griffith and Alex Irpan and Carey Radebaugh and Summer Yue and Lijie Fan and Victor Ungureanu and Christina Sorokin and Hannah Teufel and Peiran Li and Rohan Anil and Dimitris Paparas and Todd Wang and Chu-Cheng Lin and Hui Peng and Megan Shum and Goran Petrovic and Demetra Brady and Richard Nguyen and Klaus Macherey and Zhihao Li and Harman Singh and Madhavi Yenugula and Mariko Iinuma and Xinyi Chen and Kavya Kopparapu and Alexey Stern and Shachi Dave and Chandu Thekkath and Florence Perot and Anurag Kumar and Fangda Li and Yang Xiao and Matthew Bilotti and Mohammad Hossein Bateni and Isaac Noble and Lisa Lee and Amelio Vázquez-Reina and Julian Salazar and Xiaomeng Yang and Boyu Wang and Ela Gruzewska and Anand Rao and Sindhu Raghuram and Zheng Xu and Eyal Ben-David and Jieru Mei and Sid Dalmia and Zhaoyi Zhang and Yuchen Liu and Gagan Bansal and Helena Pankov and Steven Schwarcz and Andrea Burns and Christine Chan and Sumit Sanghai and Ricky Liang and Ethan Liang and Antoine He and Amy Stuart and Arun Narayanan and Yukun Zhu and Christian Frank and Bahar Fatemi and Amit Sabne and Oran Lang and Indro Bhattacharya and Shane Settle and Maria Wang and Brendan McMahan and Andrea Tacchetti and Livio Baldini Soares and Majid Hadian and Serkan Cabi and Timothy Chung and Nikita Putikhin and Gang Li and Jeremy Chen and Austin Tarango and Henryk Michalewski and Mehran Kazemi and Hussain Masoom and Hila Sheftel and Rakesh Shivanna and Archita Vadali and Ramona Comanescu and Doug Reid and Joss Moore and Arvind Neelakantan and Michaël Sander and Jonathan Herzig and Aviv Rosenberg and Mostafa Dehghani and JD Choi and Michael Fink and Reid Hayes and Eric Ge and Shitao Weng and Chia-Hua Ho and John Karro and Kalpesh Krishna and Lam Nguyen Thiet and Amy Skerry-Ryan and Daniel Eppens and Marco Andreetto and Navin Sarma and Silvano Bonacina and Burcu Karagol Ayan and Megha Nawhal and Zhihao Shan and Mike Dusenberry and Shantanu Thakoor and Sagar Gubbi and Duc Dung Nguyen and Reut Tsarfaty and Samuel Albanie and Jovana Mitrović and Meet Gandhi and Bo-Juen Chen and Alessandro Epasto and Georgi Stephanov and Ye Jin and Samuel Gehman and Aida Amini and Jack Weber and Feryal Behbahani and Shawn Xu and Miltos Allamanis and Xi Chen and Myle Ott and Claire Sha and Michal Jastrzebski and Hang Qi and David Greene and Xinyi Wu and Abodunrinwa Toki and Daniel Vlasic and Jane Shapiro and Ragha Kotikalapudi and Zhe Shen and Takaaki Saeki and Sirui Xie and Albin Cassirer and Shikhar Bharadwaj and Tatsuya Kiyono and Srinadh Bhojanapalli and Elan Rosenfeld and Sam Ritter and Jieming Mao and João Gabriel Oliveira and Zoltan Egyed and Bernd Bandemer and Emilio Parisotto and Keisuke Kinoshita and Juliette Pluto and Petros Maniatis and Steve Li and Yaohui Guo and Golnaz Ghiasi and Jean Tarbouriech and Srimon Chatterjee and Julie Jin and Katrina and Xu and Jennimaria Palomaki and Séb Arnold and Madhavi Sewak and Federico Piccinini and Mohit Sharma and Ben Albrecht and Sean Purser-haskell and Ashwin Vaswani and Chongyan Chen and Matheus Wisniewski and Qin Cao and John Aslanides and Nguyet Minh Phu and Maximilian Sieb and Lauren Agubuzu and Anne Zheng and Daniel Sohn and Marco Selvi and Anders Andreassen and Krishan Subudhi and Prem Eruvbetine and Oliver Woodman and Tomas Mery and Sebastian Krause and Xiaoqi Ren and Xiao Ma and Jincheng Luo and Dawn Chen and Wei Fan and Henry Griffiths and Christian Schuler and Alice Li and Shujian Zhang and Jean-Michel Sarr and Shixin Luo and Riccardo Patana and Matthew Watson and Dani Naboulsi and Michael Collins and Sailesh Sidhwani and Emiel Hoogeboom and Sharon Silver and Emily Caveness and Xiaokai Zhao and Mikel Rodriguez and Maxine Deines and Libin Bai and Patrick Griffin and Marco Tagliasacchi and Emily Xue and Spandana Raj Babbula and Bo Pang and Nan Ding and Gloria Shen and Elijah Peake and Remi Crocker and Shubha Srinivas Raghvendra and Danny Swisher and Woohyun Han and Richa Singh and Ling Wu and Vladimir Pchelin and Tsendsuren Munkhdalai and Dana Alon and Geoff Bacon and Efren Robles and Jannis Bulian and Melvin Johnson and George Powell and Felipe Tiengo Ferreira and Yaoyiran Li and Frederik Benzing and Mihajlo Velimirović and Hubert Soyer and William Kong and Tony and Nguyên and Zhen Yang and Jeremiah Liu and Joost van Amersfoort and Daniel Gillick and Baochen Sun and Nathalie Rauschmayr and Katie Zhang and Serena Zhan and Tao Zhou and Alexey Frolov and Chengrun Yang and Denis Vnukov and Louis Rouillard and Hongji Li and Amol Mandhane and Nova Fallen and Rajesh Venkataraman and Clara Huiyi Hu and Jennifer Brennan and Jenny Lee and Jerry Chang and Martin Sundermeyer and Zhufeng Pan and Rosemary Ke and Simon Tong and Alex Fabrikant and William Bono and Jindong Gu and Ryan Foley and Yiran Mao and Manolis Delakis and Dhruva Bhaswar and Roy Frostig and Nick Li and Avital Zipori and Cath Hope and Olga Kozlova and Swaroop Mishra and Josip Djolonga and Craig Schiff and Majd Al Merey and Eleftheria Briakou and Peter Morgan and Andy Wan and Avinatan Hassidim and RJ Skerry-Ryan and Kuntal Sengupta and Mary Jasarevic and Praveen Kallakuri and Paige Kunkle and Hannah Brennan and Tom Lieber and Hassan Mansoor and Julian Walker and Bing Zhang and Annie Xie and Goran Žužić and Adaeze Chukwuka and Alex Druinsky and Donghyun Cho and Rui Yao and Ferjad Naeem and Shiraz Butt and Eunyoung Kim and Zhipeng Jia and Mandy Jordan and Adam Lelkes and Mark Kurzeja and Sophie Wang and James Zhao and Andrew Over and Abhishek Chakladar and Marcel Prasetya and Neha Jha and Sriram Ganapathy and Yale Cong and Prakash Shroff and Carl Saroufim and Sobhan Miryoosefi and Mohamed Hammad and Tajwar Nasir and Weijuan Xi and Yang Gao and Young Maeng and Ben Hora and Chin-Yi Cheng and Parisa Haghani and Yoad Lewenberg and Caden Lu and Martin Matysiak and Naina Raisinghani and Huiyu Wang and Lexi Baugher and Rahul Sukthankar and Minh Giang and John Schultz and Noah Fiedel and Minmin Chen and Cheng-Chun Lee and Tapomay Dey and Hao Zheng and Shachi Paul and Celine Smith and Andy Ly and Yicheng Wang and Rishabh Bansal and Bartek Perz and Susanna Ricco and Stasha Blank and Vaishakh Keshava and Deepak Sharma and Marvin Chow and Kunal Lad and Komal Jalan and Simon Osindero and Craig Swanson and Jacob Scott and Anastasija Ilić and Xiaowei Li and Siddhartha Reddy Jonnalagadda and Afzal Shama Soudagar and Yan Xiong and Bat-Orgil Batsaikhan and Daniel Jarrett and Naveen Kumar and Maulik Shah and Matt Lawlor and Austin Waters and Mark Graham and Rhys May and Sabela Ramos and Sandra Lefdal and Zeynep Cankara and Nacho Cano and Brendan O'Donoghue and Jed Borovik and Frederick Liu and Jordan Grimstad and Mahmoud Alnahlawi and Katerina Tsihlas and Tom Hudson and Nikolai Grigorev and Yiling Jia and Terry Huang and Tobenna Peter Igwe and Sergei Lebedev and Xiaodan Tang and Igor Krivokon and Frankie Garcia and Melissa Tan and Eric Jia and Peter Stys and Shikhar Vashishth and Yu Liang and Balaji Venkatraman and Chenjie Gu and Anastasios Kementsietsidis and Chen Zhu and Junehyuk Jung and Yunfei Bai and Mohammad Javad Hosseini and Faruk Ahmed and Aditya Gupta and Xin Yuan and Shereen Ashraf and Shitij Nigam and Gautam Vasudevan and Pranjal Awasthi and Adi Mayrav Gilady and Zelda Mariet and Ramy Eskander and Haiguang Li and Hexiang Hu and Guillermo Garrido and Philippe Schlattner and George Zhang and Rohun Saxena and Petar Dević and Kritika Muralidharan and Ashwin Murthy and Yiqian Zhou and Min Choi and Arissa Wongpanich and Zhengdong Wang and Premal Shah and Yuntao Xu and Yiling Huang and Stephen Spencer and Alice Chen and James Cohan and Junjie Wang and Jonathan Tompson and Junru Wu and Ruba Haroun and Haiqiong Li and Blanca Huergo and Fan Yang and Tongxin Yin and James Wendt and Michael Bendersky and Rahma Chaabouni and Javier Snaider and Johan Ferret and Abhishek Jindal and Tara Thompson and Andrew Xue and Will Bishop and Shubham Milind Phal and Archit Sharma and Yunhsuan Sung and Prabakar Radhakrishnan and Mo Shomrat and Reeve Ingle and Roopali Vij and Justin Gilmer and Mihai Dorin Istin and Sam Sobell and Yang Lu and Emily Nottage and Dorsa Sadigh and Jeremiah Willcock and Tingnan Zhang and Steve Xu and Sasha Brown and Katherine Lee and Gary Wang and Yun Zhu and Yi Tay and Cheolmin Kim and Audrey Gutierrez and Abhanshu Sharma and Yongqin Xian and Sungyong Seo and Claire Cui and Elena Pochernina and Cip Baetu and Krzysztof Jastrzębski and Mimi Ly and Mohamed Elhawaty and Dan Suh and Eren Sezener and Pidong Wang and Nancy Yuen and George Tucker and Jiahao Cai and Zuguang Yang and Cindy Wang and Alex Muzio and Hai Qian and Jae Yoo and Derek Lockhart and Kevin R. McKee and Mandy Guo and Malika Mehrotra and Artur Mendonça and Sanket Vaibhav Mehta and Sherry Ben and Chetan Tekur and Jiaqi Mu and Muye Zhu and Victoria Krakovna and Hongrae Lee and AJ Maschinot and Sébastien Cevey and HyunJeong Choe and Aijun Bai and Hansa Srinivasan and Derek Gasaway and Nick Young and Patrick Siegler and Dan Holtmann-Rice and Vihari Piratla and Kate Baumli and Roey Yogev and Alex Hofer and Hado van Hasselt and Svetlana Grant and Yuri Chervonyi and David Silver and Andrew Hogue and Ayushi Agarwal and Kathie Wang and Preeti Singh and Four Flynn and Josh Lipschultz and Robert David and Lizzetth Bellot and Yao-Yuan Yang and Long Le and Filippo Graziano and Kate Olszewska and Kevin Hui and Akanksha Maurya and Nikos Parotsidis and Weijie Chen and Tayo Oguntebi and Joe Kelley and Anirudh Baddepudi and Johannes Mauerer and Gregory Shaw and Alex Siegman and Lin Yang and Shravya Shetty and Subhrajit Roy and Yunting Song and Wojciech Stokowiec and Ryan Burnell and Omkar Savant and Robert Busa-Fekete and Jin Miao and Samrat Ghosh and Liam MacDermed and Phillip Lippe and Mikhail Dektiarev and Zach Behrman and Fabian Mentzer and Kelvin Nguyen and Meng Wei and Siddharth Verma and Chris Knutsen and Sudeep Dasari and Zhipeng Yan and Petr Mitrichev and Xingyu Wang and Virat Shejwalkar and Jacob Austin and Srinivas Sunkara and Navneet Potti and Yan Virin and Christian Wright and Gaël Liu and Oriana Riva and Etienne Pot and Greg Kochanski and Quoc Le and Gargi Balasubramaniam and Arka Dhar and Yuguo Liao and Adam Bloniarz and Divyansh Shukla and Elizabeth Cole and Jong Lee and Sheng Zhang and Sushant Kafle and Siddharth Vashishtha and Parsa Mahmoudieh and Grace Chen and Raphael Hoffmann and Pranesh Srinivasan and Agustin Dal Lago and Yoav Ben Shalom and Zi Wang and Michael Elabd and Anuj Sharma and Junhyuk Oh and Suraj Kothawade and Maigo Le and Marianne Monteiro and Shentao Yang and Kaiz Alarakyia and Robert Geirhos and Diana Mincu and Håvard Garnes and Hayato Kobayashi and Soroosh Mariooryad and Kacper Krasowiak and Zhixin and Lai and Shibl Mourad and Mingqiu Wang and Fan Bu and Ophir Aharoni and Guanjie Chen and Abhimanyu Goyal and Vadim Zubov and Ankur Bapna and Elahe Dabir and Nisarg Kothari and Kay Lamerigts and Nicola De Cao and Jeremy Shar and Christopher Yew and Nitish Kulkarni and Dre Mahaarachchi and Mandar Joshi and Zhenhai Zhu and Jared Lichtarge and Yichao Zhou and Hannah Muckenhirn and Vittorio Selo and Oriol Vinyals and Peter Chen and Anthony Brohan and Vaibhav Mehta and Sarah Cogan and Ruth Wang and Ty Geri and Wei-Jen Ko and Wei Chen and Fabio Viola and Keshav Shivam and Lisa Wang and Madeleine Clare Elish and Raluca Ada Popa and Sébastien Pereira and Jianqiao Liu and Raphael Koster and Donnie Kim and Gufeng Zhang and Sayna Ebrahimi and Partha Talukdar and Yanyan Zheng and Petra Poklukar and Ales Mikhalap and Dale Johnson and Anitha Vijayakumar and Mark Omernick and Matt Dibb and Ayush Dubey and Qiong Hu and Apurv Suman and Vaibhav Aggarwal and Ilya Kornakov and Fei Xia and Wing Lowe and Alexey Kolganov and Ted Xiao and Vitaly Nikolaev and Steven Hemingray and Bonnie Li and Joana Iljazi and Mikołaj Rybiński and Ballie Sandhu and Peggy Lu and Thang Luong and Rodolphe Jenatton and Vineetha Govindaraj and Hui and Li and Gabriel Dulac-Arnold and Wonpyo Park and Henry Wang and Abhinit Modi and Jean Pouget-Abadie and Kristina Greller and Rahul Gupta and Robert Berry and Prajit Ramachandran and Jinyu Xie and Liam McCafferty and Jianling Wang and Kilol Gupta and Hyeontaek Lim and Blaž Bratanič and Andy Brock and Ilia Akolzin and Jim Sproch and Dan Karliner and Duhyeon Kim and Adrian Goedeckemeyer and Noam Shazeer and Cordelia Schmid and Daniele Calandriello and Parul Bhatia and Krzysztof Choromanski and Ceslee Montgomery and Dheeru Dua and Ana Ramalho and Helen King and Yue Gao and Lynn Nguyen and David Lindner and Divya Pitta and Oleaser Johnson and Khalid Salama and Diego Ardila and Michael Han and Erin Farnese and Seth Odoom and Ziyue Wang and Xiangzhuo Ding and Norman Rink and Ray Smith and Harshal Tushar Lehri and Eden Cohen and Neera Vats and Tong He and Parthasarathy Gopavarapu and Adam Paszke and Miteyan Patel and Wouter Van Gansbeke and Lucia Loher and Luis Castro and Maria Voitovich and Tamara von Glehn and Nelson George and Simon Niklaus and Zach Eaton-Rosen and Nemanja Rakićević and Erik Jue and Sagi Perel and Carrie Zhang and Yuval Bahat and Angéline Pouget and Zhi Xing and Fantine Huot and Ashish Shenoy and Taylor Bos and Vincent Coriou and Bryan Richter and Natasha Noy and Yaqing Wang and Santiago Ontanon and Siyang Qin and Gleb Makarchuk and Demis Hassabis and Zhuowan Li and Mandar Sharma and Kumaran Venkatesan and Iurii Kemaev and Roxanne Daniel and Shiyu Huang and Saloni Shah and Octavio Ponce and Warren and Chen and Manaal Faruqui and Jialin Wu and Slavica Andačić and Szabolcs Payrits and Daniel McDuff and Tom Hume and Yuan Cao and MH Tessler and Qingze Wang and Yinan Wang and Ivor Rendulic and Eirikur Agustsson and Matthew Johnson and Tanya Lando and Andrew Howard and Sri Gayatri Sundara Padmanabhan and Mayank Daswani and Andrea Banino and Michael Kilgore and Jonathan Heek and Ziwei Ji and Alvaro Caceres and Conglong Li and Nora Kassner and Alexey Vlaskin and Zeyu Liu and Alex Grills and Yanhan Hou and Roykrong Sukkerd and Gowoon Cheon and Nishita Shetty and Larisa Markeeva and Piotr Stanczyk and Tejas Iyer and Yuan Gong and Shawn Gao and Keerthana Gopalakrishnan and Tim Blyth and Malcolm Reynolds and Avishkar Bhoopchand and Misha Bilenko and Dero Gharibian and Vicky Zayats and Aleksandra Faust and Abhinav Singh and Min Ma and Hongyang Jiao and Sudheendra Vijayanarasimhan and Lora Aroyo and Vikas Yadav and Sarah Chakera and Ashwin Kakarla and Vilobh Meshram and Karol Gregor and Gabriela Botea and Evan Senter and Dawei Jia and Geza Kovacs and Neha Sharma and Sebastien Baur and Kai Kang and Yifan He and Lin Zhuo and Marija Kostelac and Itay Laish and Songyou Peng and Louis O'Bryan and Daniel Kasenberg and Girish Ramchandra Rao and Edouard Leurent and Biao Zhang and Sage Stevens and Ana Salazar and Ye Zhang and Ivan Lobov and Jake Walker and Allen Porter and Morgan Redshaw and Han Ke and Abhishek Rao and Alex Lee and Hoi Lam and Michael Moffitt and Jaeyoun Kim and Siyuan Qiao and Terry Koo and Robert Dadashi and Xinying Song and Mukund Sundararajan and Peng Xu and Chizu Kawamoto and Yan Zhong and Clara Barbu and Apoorv Reddy and Mauro Verzetti and Leon Li and George Papamakarios and Hanna Klimczak-Plucińska and Mary Cassin and Koray Kavukcuoglu and Rigel Swavely and Alain Vaucher and Jeffrey Zhao and Ross Hemsley and Michael Tschannen and Heming Ge and Gaurav Menghani and Yang Yu and Natalie Ha and Wei He and Xiao Wu and Maggie Song and Rachel Sterneck and Stefan Zinke and Dan A. Calian and Annie Marsden and Alejandro Cruzado Ruiz and Matteo Hessel and Almog Gueta and Benjamin Lee and Brian Farris and Manish Gupta and Yunjie Li and Mohammad Saleh and Vedant Misra and Kefan Xiao and Piermaria Mendolicchio and Gavin Buttimore and Varvara Krayvanova and Nigamaa Nayakanti and Matthew Wiethoff and Yash Pande and Azalia Mirhoseini and Ni Lao and Jasmine Liu and Yiqing Hua and Angie Chen and Yury Malkov and Dmitry Kalashnikov and Shubham Gupta and Kartik Audhkhasi and Yuexiang Zhai and Sudhindra Kopalle and Prateek Jain and Eran Ofek and Clemens Meyer and Khuslen Baatarsukh and Hana Strejček and Jun Qian and James Freedman and Ricardo Figueira and Michal Sokolik and Olivier Bachem and Raymond Lin and Dia Kharrat and Chris Hidey and Pingmei Xu and Dennis Duan and Yin Li and Muge Ersoy and Richard Everett and Kevin Cen and Rebeca Santamaria-Fernandez and Amir Taubenfeld and Ian Mackinnon and Linda Deng and Polina Zablotskaia and Shashank Viswanadha and Shivanker Goel and Damion Yates and Yunxiao Deng and Peter Choy and Mingqing Chen and Abhishek Sinha and Alex Mossin and Yiming Wang and Arthur Szlam and Susan Hao and Paul Kishan Rubenstein and Metin Toksoz-Exley and Miranda Aperghis and Yin Zhong and Junwhan Ahn and Michael Isard and Olivier Lacombe and Florian Luisier and Chrysovalantis Anastasiou and Yogesh Kalley and Utsav Prabhu and Emma Dunleavy and Shaan Bijwadia and Justin Mao-Jones and Kelly Chen and Rama Pasumarthi and Emily Wood and Adil Dostmohamed and Nate Hurley and Jiri Simsa and Alicia Parrish and Mantas Pajarskas and Matt Harvey and Ondrej Skopek and Yony Kochinski and Javier Rey and Verena Rieser and Denny Zhou and Sun Jae Lee and Trilok Acharya and Guowang Li and Joe Jiang and Xiaofan Zhang and Bryant Gipson and Ethan Mahintorabi and Marco Gelmi and Nima Khajehnouri and Angel Yeh and Kayi Lee and Loic Matthey and Leslie Baker and Trang Pham and Han Fu and Alex Pak and Prakhar Gupta and Cristina Vasconcelos and Adam Sadovsky and Brian Walker and Sissie Hsiao and Patrik Zochbauer and Andreea Marzoca and Noam Velan and Junhao Zeng and Gilles Baechler and Danny Driess and Divya Jain and Yanping Huang and Lizzie Tao and John Maggs and Nir Levine and Jon Schneider and Erika Gemzer and Samuel Petit and Shan Han and Zach Fisher and Dustin Zelle and Courtney Biles and Eugene Ie and Asya Fadeeva and Casper Liu and Juliana Vicente Franco and Adrian Collister and Hao Zhang and Renshen Wang and Ruizhe Zhao and Leandro Kieliger and Kurt Shuster and Rui Zhu and Boqing Gong and Lawrence Chan and Ruoxi Sun and Sujoy Basu and Roland Zimmermann and Jamie Hayes and Abhishek Bapna and Jasper Snoek and Weel Yang and Puranjay Datta and Jad Al Abdallah and Kevin Kilgour and Lu Li and SQ Mah and Yennie Jun and Morgane Rivière and Abhijit Karmarkar and Tammo Spalink and Tao Huang and Lucas Gonzalez and Duc-Hieu Tran and Averi Nowak and John Palowitch and Martin Chadwick and Ellie Talius and Harsh Mehta and Thibault Sellam and Philipp Fränken and Massimo Nicosia and Kyle He and Aditya Kini and David Amos and Sugato Basu and Harrison Jobe and Eleni Shaw and Qiantong Xu and Colin Evans and Daisuke Ikeda and Chaochao Yan and Larry Jin and Lun Wang and Sachin Yadav and Ilia Labzovsky and Ramesh Sampath and Ada Ma and Candice Schumann and Aditya Siddhant and Rohin Shah and John Youssef and Rishabh Agarwal and Natalie Dabney and Alessio Tonioni and Moran Ambar and Jing Li and Isabelle Guyon and Benny Li and David Soergel and Boya Fang and Georgi Karadzhov and Cristian Udrescu and Trieu Trinh and Vikas Raunak and Seb Noury and Dee Guo and Sonal Gupta and Mara Finkelstein and Denis Petek and Lihao Liang and Greg Billock and Pei Sun and David Wood and Yiwen Song and Xiaobin Yu and Tatiana Matejovicova and Regev Cohen and Kalyan Andra and David D'Ambrosio and Zhiwei Deng and Vincent Nallatamby and Ebrahim Songhori and Rumen Dangovski and Andrew Lampinen and Pankil Botadra and Adam Hillier and Jiawei Cao and Nagabhushan Baddi and Adhi Kuncoro and Toshihiro Yoshino and Ankit Bhagatwala and Marcáurelio Ranzato and Rylan Schaeffer and Tianlin Liu and Shuai Ye and Obaid Sarvana and John Nham and Chenkai Kuang and Isabel Gao and Jinoo Baek and Shubham Mittal and Ayzaan Wahid and Anita Gergely and Bin Ni and Josh Feldman and Carrie Muir and Pascal Lamblin and Wolfgang Macherey and Ethan Dyer and Logan Kilpatrick and Víctor Campos and Mukul Bhutani and Stanislav Fort and Yanif Ahmad and Aliaksei Severyn and Kleopatra Chatziprimou and Oleksandr Ferludin and Mason Dimarco and Aditya Kusupati and Joe Heyward and Dan Bahir and Kevin Villela and Katie Millican and Dror Marcus and Sanaz Bahargam and Caglar Unlu and Nicholas Roth and Zichuan Wei and Siddharth Gopal and Deepanway Ghoshal and Edward Lee and Sharon Lin and Jennie Lees and Dayeong Lee and Anahita Hosseini and Connie Fan and Seth Neel and Marcus Wu and Yasemin Altun and Honglong Cai and Enrique Piqueras and Josh Woodward and Alessandro Bissacco and Salem Haykal and Mahyar Bordbar and Prasha Sundaram and Sarah Hodkinson and Daniel Toyama and George Polovets and Austin Myers and Anu Sinha and Tomer Levinboim and Kashyap Krishnakumar and Rachita Chhaparia and Tatiana Sholokhova and Nitesh Bharadwaj Gundavarapu and Ganesh Jawahar and Haroon Qureshi and Jieru Hu and Nikola Momchev and Matthew Rahtz and Renjie Wu and Aishwarya P S and Kedar Dhamdhere and Meiqi Guo and Umang Gupta and Ali Eslami and Mariano Schain and Michiel Blokzijl and David Welling and Dave Orr and Levent Bolelli and Nicolas Perez-Nieves and Mikhail Sirotenko and Aman Prasad and Arjun Kar and Borja De Balle Pigem and Tayfun Terzi and Gellért Weisz and Dipankar Ghosh and Aditi Mavalankar and Dhruv Madeka and Kaspar Daugaard and Hartwig Adam and Viraj Shah and Dana Berman and Maggie Tran and Steven Baker and Ewa Andrejczuk and Grishma Chole and Ganna Raboshchuk and Mahdi Mirzazadeh and Thais Kagohara and Shimu Wu and Christian Schallhart and Bernett Orlando and Chen Wang and Alban Rrustemi and Hao Xiong and Hao Liu and Arpi Vezer and Nolan Ramsden and Shuo-yiin Chang and Sidharth Mudgal and Yan Li and Nino Vieillard and Yedid Hoshen and Farooq Ahmad and Ambrose Slone and Amy Hua and Natan Potikha and Mirko Rossini and Jon Stritar and Sushant Prakash and Zifeng Wang and Xuanyi Dong and Alireza Nazari and Efrat Nehoran and Kaan Tekelioglu and Yinxiao Li and Kartikeya Badola and Tom Funkhouser and Yuanzhen Li and Varun Yerram and Ramya Ganeshan and Daniel Formoso and Karol Langner and Tian Shi and Huijian Li and Yumeya Yamamori and Amayika Panda and Alaa Saade and Angelo Scorza Scarpati and Chris Breaux and CJ Carey and Zongwei Zhou and Cho-Jui Hsieh and Sophie Bridgers and Alena Butryna and Nishesh Gupta and Vaibhav Tulsyan and Sanghyun Woo and Evgenii Eltyshev and Will Grathwohl and Chanel Parks and Seth Benjamin and Rina Panigrahy and Shenil Dodhia and Daniel De Freitas and Chris Sauer and Will Song and Ferran Alet and Jackson Tolins and Cosmin Paduraru and Xingyi Zhou and Brian Albert and Zizhao Zhang and Lei Shu and Mudit Bansal and Sarah Nguyen and Amir Globerson and Owen Xiao and James Manyika and Tom Hennigan and Rong Rong and Josip Matak and Anton Bakalov and Ankur Sharma and Danila Sinopalnikov and Andrew Pierson and Stephen Roller and Geoff Brown and Mingcen Gao and Toshiyuki Fukuzawa and Amin Ghafouri and Kenny Vassigh and Iain Barr and Zhicheng Wang and Anna Korsun and Rajesh Jayaram and Lijie Ren and Tim Zaman and Samira Khan and Yana Lunts and Dan Deutsch and Dave Uthus and Nitzan Katz and Masha Samsikova and Amr Khalifa and Nikhil Sethi and Jiao Sun and Luming Tang and Uri Alon and Xianghong Luo and Dian Yu and Abhishek Nayyar and Bryce Petrini and Will Truong and Vincent Hellendoorn and Nikolai Chinaev and Chris Alberti and Wei Wang and Jingcao Hu and Vahab Mirrokni and Ananth Balashankar and Avia Aharon and Aahil Mehta and Ahmet Iscen and Joseph Kready and Lucas Manning and Anhad Mohananey and Yuankai Chen and Anshuman Tripathi and Allen Wu and Igor Petrovski and Dawsen Hwang and Martin Baeuml and Shreyas Chandrakaladharan and Yuan Liu and Rey Coaguila and Maxwell Chen and Sally Ma and Pouya Tafti and Susheel Tatineni and Terry Spitz and Jiayu Ye and Paul Vicol and Mihaela Rosca and Adrià Puigdomènech and Zohar Yahav and Sanjay Ghemawat and Hanzhao Lin and Phoebe Kirk and Zaid Nabulsi and Sergey Brin and Bernd Bohnet and Ken Caluwaerts and Aditya Srikanth Veerubhotla and Dan Zheng and Zihang Dai and Petre Petrov and Yichong Xu and Ramin Mehran and Zhuo Xu and Luisa Zintgraf and Jiho Choi and Spurthi Amba Hombaiah and Romal Thoppilan and Sashank Reddi and Lukasz Lew and Li Li and Kellie Webster and KP Sawhney and Lampros Lamprou and Siamak Shakeri and Mayank Lunayach and Jianmin Chen and Sumit Bagri and Alex Salcianu and Ying Chen and Yani Donchev and Charlotte Magister and Signe Nørly and Vitor Rodrigues and Tomas Izo and Hila Noga and Joe Zou and Thomas Köppe and Wenxuan Zhou and Kenton Lee and Xiangzhu Long and Danielle Eisenbud and Anthony Chen and Connor Schenck and Chi Ming To and Peilin Zhong and Emanuel Taropa and Minh Truong and Omer Levy and Danilo Martins and Zhiyuan Zhang and Christopher Semturs and Kelvin Zhang and Alex Yakubovich and Pol Moreno and Lara McConnaughey and Di Lu and Sam Redmond and Lotte Weerts and Yonatan Bitton and Tiziana Refice and Nicolas Lacasse and Arthur Conmy and Corentin Tallec and Julian Odell and Hannah Forbes-Pollard and Arkadiusz Socala and Jonathan Hoech and Pushmeet Kohli and Alanna Walton and Rui Wang and Mikita Sazanovich and Kexin Zhu and Andrei Kapishnikov and Rich Galt and Matthew Denton and Ben Murdoch and Caitlin Sikora and Kareem Mohamed and Wei Wei and Uri First and Tim McConnell and Luis C. Cobo and James Qin and Thi Avrahami and Daniel Balle and Yu Watanabe and Annie Louis and Adam Kraft and Setareh Ariafar and Yiming Gu and Eugénie Rives and Charles Yoon and Andrei Rusu and James Cobon-Kerr and Chris Hahn and Jiaming Luo and Yuvein and Zhu and Niharika Ahuja and Rodrigo Benenson and Raphaël Lopez Kaufman and Honglin Yu and Lloyd Hightower and Junlin Zhang and Darren Ni and Lisa Anne Hendricks and Gabby Wang and Gal Yona and Lalit Jain and Pablo Barrio and Surya Bhupatiraju and Siva Velusamy and Allan Dafoe and Sebastian Riedel and Tara Thomas and Zhe Yuan and Mathias Bellaiche and Sheena Panthaplackel and Klemen Kloboves and Sarthak Jauhari and Canfer Akbulut and Todor Davchev and Evgeny Gladchenko and David Madras and Aleksandr Chuklin and Tyrone Hill and Quan Yuan and Mukundan Madhavan and Luke Leonhard and Dylan Scandinaro and Qihang Chen and Ning Niu and Arthur Douillard and Bogdan Damoc and Yasumasa Onoe and Fabian Pedregosa and Fred Bertsch and Chas Leichner and Joseph Pagadora and Jonathan Malmaud and Sameera Ponda and Andy Twigg and Oleksii Duzhyi and Jingwei Shen and Miaosen Wang and Roopal Garg and Jing Chen and Utku Evci and Jonathan Lee and Leon Liu and Koji Kojima and Masa Yamaguchi and Arunkumar Rajendran and AJ Piergiovanni and Vinodh Kumar Rajendran and Marco Fornoni and Gabriel Ibagon and Harry Ragan and Sadh MNM Khan and John Blitzer and Andrew Bunner and Guan Sun and Takahiro Kosakai and Scott Lundberg and Ndidi Elue and Kelvin Guu and SK Park and Jane Park and Arunachalam Narayanaswamy and Chengda Wu and Jayaram Mudigonda and Trevor Cohn and Hairong Mu and Ravi Kumar and Laura Graesser and Yichi Zhang and Richard Killam and Vincent Zhuang and Mai Giménez and Wael Al Jishi and Ruy Ley-Wild and Alex Zhai and Kazuki Osawa and Diego Cedillo and Jialu Liu and Mayank Upadhyay and Marcin Sieniek and Roshan Sharma and Tom Paine and Anelia Angelova and Sravanti Addepalli and Carolina Parada and Kingshuk Majumder and Avery Lamp and Sanjiv Kumar and Xiang Deng and Artiom Myaskovsky and Tea Sabolić and Jeffrey Dudek and Sarah York and Félix de Chaumont Quitry and Jiazhong Nie and Dee Cattle and Alok Gunjan and Bilal Piot and Waleed Khawaja and Seojin Bang and Simon Wang and Siavash Khodadadeh and Raghavender R and Praynaa Rawlani and Richard Powell and Kevin Lee and Johannes Griesser and GS Oh and Cesar Magalhaes and Yujia Li and Simon Tokumine and Hadas Natalie Vogel and Dennis Hsu and Arturo BC and Disha Jindal and Matan Cohen and Zi Yang and Junwei Yuan and Dario de Cesare and Tony Bruguier and Jun Xu and Monica Roy and Alon Jacovi and Dan Belov and Rahul Arya and Phoenix Meadowlark and Shlomi Cohen-Ganor and Wenting Ye and Patrick Morris-Suzuki and Praseem Banzal and Gan Song and Pranavaraj Ponnuramu and Fred Zhang and George Scrivener and Salah Zaiem and Alif Raditya Rochman and Kehang Han and Badih Ghazi and Kate Lee and Shahar Drath and Daniel Suo and Antonious Girgis and Pradeep Shenoy and Duy Nguyen and Douglas Eck and Somit Gupta and Le Yan and Joao Carreira and Anmol Gulati and Ruoxin Sang and Daniil Mirylenka and Emma Cooney and Edward Chou and Mingyang Ling and Cindy Fan and Ben Coleman and Guilherme Tubone and Ravin Kumar and Jason Baldridge and Felix Hernandez-Campos and Angeliki Lazaridou and James Besley and Itay Yona and Neslihan Bulut and Quentin Wellens and AJ Pierigiovanni and Jasmine George and Richard Green and Pu Han and Connie Tao and Geoff Clark and Chong You and Abbas Abdolmaleki and Justin Fu and Tongzhou Chen and Ashwin Chaugule and Angad Chandorkar and Altaf Rahman and Will Thompson and Penporn Koanantakool and Mike Bernico and Jie Ren and Andrey Vlasov and Sergei Vassilvitskii and Maciej Kula and Yizhong Liang and Dahun Kim and Yangsibo Huang and Chengxi Ye and Dmitry Lepikhin and Wesley Helmholz},
      year={2025},
      eprint={2507.06261},
      archivePrefix={arXiv},
      primaryClass={cs.CL},
      url={https://arxiv.org/abs/2507.06261}, 
}

@inproceedings{vllm,
  title={Efficient Memory Management for Large Language Model Serving with PagedAttention},
  author={Woosuk Kwon and Zhuohan Li and Siyuan Zhuang and Ying Sheng and Lianmin Zheng and Cody Hao Yu and Joseph E. Gonzalez and Hao Zhang and Ion Stoica},
  booktitle={Proceedings of the ACM SIGOPS 29th Symposium on Operating Systems Principles},
  year={2023}
}

@inproceedings{seetharaman2020autoclip,
  title={AutoClip: Adaptive Gradient Clipping for Source Separation Networks},
  author={Seetharaman Prem and Wichern Gordon and Pardo Bryan and Le Roux Jonathan},
  booktitle={2020 IEEE 30th International Workshop on Machine Learning for Signal Processing (MLSP)},
  year={2020},
  organization={IEEE}
}

\appendix

\section{Dataset Details}
\label{sec:datasetdetails}

\begin{table}[h]
\centering
\vskip 0.15in
\begin{small}
\begin{tabularx}{\columnwidth}{l r X}
\toprule
\textbf{Dataset} & \textbf{NC} & \textbf{HF Repo / Source} \\
\midrule
\rowcolor[gray]{.95} \multicolumn{3}{l}{\textit{Training Datasets}} \\
snli & 3 & stanfordnlp/snli \\
sst5 & 5 & SetFit/sst5 \\
rte & 2 & SetFit/rte \\
Trec-QC & 6 & SetFit/TREC-QC \\
agnews & 4 & SetFit/ag\_news \\
goemotions & 28 & mrm8488/goemotions \\
dbpedia\_70 & 70 & DeveloperOats/DBPedia\_Classes \\
toxic chat & 2 & lmsys/toxic-chat \\
newsgroups20 & 20 & SetFit/20\_newsgroups \\
yahoo answers & 10 & community-datasets/yahoo\_answers\_topics \\
\midrule
\rowcolor[gray]{.95} \multicolumn{3}{l}{\textit{Validation Datasets}} \\
qqp & 2 & SetFit/qqp \\
banking77 & 77 & mteb/banking77 \\
dbpedia\_9 & 9 & DeveloperOats/DBPedia\_Classes \\
Yelp reviews & 5 & Yelp/yelp\_review\_full \\
\midrule
\rowcolor[gray]{.95} \multicolumn{3}{l}{\textit{Test (Unseen) Datasets}} \\
\amazonc & 2 & mteb/amazon\_counterfactual \\
\commons & 5 & tau/commonsense\_qa \\
massive intent & 60 & mteb/amazon\_massive\_intent \\
medqa & 4 & bigbio/med\_qa \\
metatool & 47 & \href{https://github.com/HowieHwong/MetaTool/tree/master/dataset}{MetaTool GitHub} \\
when2call & 3 & nvidia/When2Call \\
wikitoxic & 2 & OxAISH-AL-LLM/wiki\_toxic \\
wildguard & 15 & allenai/wildguardmix \\
\bottomrule
\end{tabularx}
\end{small}
\caption{Consolidated Datasets: Training, Validation, and Test with their number of classes (NC) and source.}
\label{tab:all_datasets}
\end{table}

\section{Calibrator Overview}
\sysname\ is a causal transformer encoder that is a stack of the following layers:
\begin{enumerate}
    \item A linear embedding layer to which sinusoidal position embeddings are added
    \item 12 standard transformer encoder layers with hidden dimension $h$=64 and 8 attention heads. All layers use a causal mask for attention. 
    \item A final linear layer that maps $R^h$ hidden features to $R$
    \item A sigmoid layer that maps the last layer's output logits to likelihood of prediction correctness, which serves as the calibrated confidence
\end{enumerate}

\section{Related Work}
\label{sec:related}
\xhdr{Calibration of LLM Predictions}
Modern LLMs are known to be overconfident, assigning high probabilities to predictions even when the true answer involves substantial aleatoric uncertainty~\cite{NEURIPS2024_b0a4b3e3,Steyvers2024TheCG,xiong2024can,tian-etal-2023-just}. This miscalibration is exacerbated under in-context learning: \citet{zhang2024study} show that calibration error does not follow a predictable pattern as the number of demonstrations increases, and that standard post-hoc methods such as temperature scaling do not reliably improve it. These findings motivate the development of calibration techniques designed specifically for the ICL setting.

\xhdr{Prompt-Bias Correction Methods}
A prominent family of approaches frames ICL miscalibration as a prompt-induced label bias and seeks to subtract it from the predicted logits. Contextual Calibration \cite{Zhao2021CalibrateBU} estimates this bias by replacing the test input with content-free tokens such as ``N/A'' or ``[MASK]''. Batch Calibration \cite{zhou-etal-2024-batch} instead averages prediction logits over the test batch to estimate the label prior. In-Context Calibration \cite{Jang2024RectifyingDS-ICC} uses only the in-context examples to estimate semantic priors, avoiding the need for additional test-time data.
These methods primarily target accuracy improvement rather than statistical calibration, require task-specific labeled data and are designed for randomly sampled demonstrations. When demonstrations are selected by similarity to the test input, subtracting the estimated ``bias'' can reduce confidence in the correct class and degrade both accuracy and calibration.

\xhdr{Learned Calibration from In-Context Signals}
A more recent direction learns to calibrate by modeling the sequential structure of in-context predictions. Surprise Calibration \cite{tan-etal-2025-surprise} interprets ICL as sequential Bayesian inference and extracts from the negative log-probability of the true label at each demonstration position a per-class logit adjustment. While conceptually related to our use of prediction--label discrepancies across the demonstration sequence, Surprise Calibration differs in several important respects: it trains a separate model per LLM--task pair, it targets accuracy rather than calibration error, and it has been evaluated only on tasks with 2--3 label spaces.

Separately, the Simulated Annotators approach \cite{jung2025trust} improves calibration by averaging predictions across $N$ random permutations of the in-context examples. While effective, this strategy requires $N$ forward passes per test input, making it substantially more expensive than methods that calibrate a single prediction.

\xhdr{Auxiliary Calibration Models}
An alternative line of work trains a dedicated auxiliary model to predict an LLM's confidence. APRICOT \cite{apricot} trains a DeBERTa model on the textual input and the LLM's generated answer, using cluster-level accuracy computed by grouping questions with similar embeddings as the calibration target. 
The auxiliary model must be retrained for each target LLM, and the approach does not leverage in-context examples or their sequential structure.
Thermometer \cite{pmlr-shen-thermo} trains an MLP on the LLM's hidden-state representations to predict a per-task calibration temperature. While some transfer is observed within model families and across related tasks, the reliance on internal representations of fixed dimensionality ties the calibrator to a specific LLM architecture.

Both approaches lack the cross-model and cross-task universality that \sysname\ targets. In contrast, our calibrator operates on task-agnostic, prediction-derived features and exploits the sequential structure of in-context predictions, enabling a single trained model to generalize to unseen LLMs and tasks.

\section{Baseline Details}
\label{sec:baselines}
We provide the formal details for our baselines here.
For all methods below, we treat returned log-probabilities as logits (vLLM does not expose logits directly; softmax-normalized probabilities are identical either way).

\xhdr{In-Context Calibration (\ICC)} \cite{Jang2024RectifyingDS-ICC}
\ICC\ estimates a prompt-induced ``semantic prior'' by predicting on each demonstration while holding it out from the context.
Let \(\mathcal{C} = \{(x_1,y_1),\dots,(x_k,y_k)\}\) be the ordered demonstrations. We compute
\begin{equation}
\begin{aligned}
P_i &= P\!\left(y \mid x_i, \{(x_j,y_j)\}_{j \in [k]\setminus\{i\}}\right),\space \bar{P} = \frac{1}{k} \sum_{i=1}^{k} P_i, \\
\tilde{\mathbf{z}} &= \mathbf{z} - \log \bar{P}.
\end{aligned}
\end{equation}
where \(\mathbf{z}\) and \(\tilde{\mathbf{z}}\) denote the original and calibrated logits, respectively.

\xhdr{Temperature scaling from autoregressive demonstration predictions (\ICTemp)}
\ICTemp\ fits a single temperature per prompt using autoregressive predictions on the demonstrations:
\begin{equation}
\begin{aligned}
\mathbf{z}_i &= f_{\mathrm{LM}}\!\left(x_i \mid \{(x_j,y_j)\}_{j=1}^{i-1}\right), \\
T &= \arg\min_T \sum_{i=1}^{k} \mathrm{NLL}\!\left(\sigma(\mathbf{z}_i / T), y_i\right) \,,
\end{aligned}
\end{equation}
where $\mathrm{NLL}$ refers to negative log-likelihood.

\xhdr{Temperature scaling from fixed $t$-shot demonstration predictions (\FSICTemp)}
\FSICTemp\ fits a temperature using $t$-shot predictions for each demonstration (with $t=(k-1)/2$):
\begin{equation}
\begin{aligned}
\mathcal{C}_i^{(t)} &\subset \mathcal{C} \setminus \{(x_i, y_i)\}, \quad |\mathcal{C}_i^{(t)}| = t, \\
\mathbf{z}_i^{(t)} &= f_{\mathrm{LM}}\!\left(x_i \mid \mathcal{C}_i^{(t)}\right), \\
T^{*} &= \arg\min_T \sum_{i=1}^{k} \mathrm{NLL}\!\left(\sigma(\mathbf{z}_i^{(t)} / T),\, y_i\right).
\end{aligned}
\end{equation}

\xhdr{Permutation averaging (\Permute)} \cite{jung2025trust}
To reduce sensitivity to demonstration order, In-Context Permutation (\Permute) averages predictions across $N=5$ random permutations of \(\mathcal{C}\):
\begin{equation}
\bar{P}(y \mid x_{test}) =
\frac{1}{N} \sum_{n=1}^{N} p\!\left(y \mid x_{test}, \pi_n(\mathcal{C})\right).
\end{equation}

\paragraph{Implementation Details for Temperature Search.}
\phantomsection\label{sec:temp-search}
All temperature scaling baselines optimize a per-sequence temperature---a scalar $T_i \in \mathbb{R}_{>0}$ for each test instance $i$---by minimizing the negative log-likelihood over the demonstration predictions associated with that instance. The optimization is performed via up to 500 iterations of ternary search over the interval $[0.01, 100.0]$, since the cross-entropy loss is convex (and hence unimodal) in $\log T$. Search terminates when the interval width falls below $\tau = 10^{-4}$, yielding temperature estimates accurate to within $10^{-4}$ in absolute value. For \sysname\, the equivalent temperature is recovered via binary search over the same interval $[0.01, 100.0]$ with $500$ iterations: given the calibrated confidence $\hat{c}_{k+1} \in (0,1)$ predicted by the transformer, we find $T^* = \arg\min_{T} \left|\max_{j} \operatorname{softmax}(T \cdot \mathbf{z}_{k+1})_j - \hat{c}_{k+1}\right|$ by bisection. In the rare case that $\hat{c}_{k+1}$ is outside the confidence interval attainable given the scaling temperature interval, the temperature is clamped to the corresponding interval bound.

\section{Extracting Label Distribution with a Label Token Trie}
\label{sec:trie}
As mentioned in \S\ref{sec:llmpd}, we calculate label probabilities by organizing the label set into a \emph{label token trie}, where each path from the root to a leaf corresponds to the token sequence of one label.

At each internal node, we perform a forward pass with the partial label appended to the prompt and extract logits only over the valid next tokens—those that correspond to children of the current node. Applying a softmax over these restricted logits yields a properly normalized conditional distribution at each branching point. The probability of a complete label is then the product of conditional probabilities along its path:
\begin{equation}
P_\cM(y \mid x, \cC) = \prod_{i=1}^{|y|} P_\cM\!\left(y^{(i)} \mid y^{(1:i-1)}, x, \cC\right),
\end{equation}
where $y^{(i)}$ denotes the $i$-th token of label $y$ and $y^{(1:i-1)}$ the preceding tokens. 
Since the softmax at each branching point is taken over the valid continuations only, the resulting label probabilities are guaranteed to sum to 1 over $\cY_\tau$ regardless of the number of classes or the depth of the trie.

A toy illustration of our label token trie application is given below for the following labels: \begin{itemize}
    \item \texttt{card\_arrived}
    \item \texttt{card\_declined}
    \item \texttt{card\_in\_transit}
    \item \texttt{cash\_transferred}
\end{itemize}

\begin{verbatim}
ROOT
|-- card
|   |-- _arrived    -> card_arrived
|   |-- _declined   -> card_declined
|   `-- _in
|       `-- _trans
|           `-- it  -> card_in_transit
`-- cash
    `-- _trans
        `-- ferred  -> cash_transferred
\end{verbatim}

We first obtain the logits for tokens \texttt{card} and \texttt{cash} by prompting the model, then append, for example, \texttt{card} to the prompt and perform inference to get logits for the next token candidates: \texttt{\_arrived}, \texttt{\_declined}, and \texttt{\_in}. Since \texttt{\_transit} is the only valid completion after \texttt{card\_in}, we assign a dummy logit to it, as the softmax over the only candidate next token will yield a probability of 1 regardless of the true logit value.

\begin{flalign*}
&P(\texttt{card\_in\_transit} \mid x, C) &&\\
&= P(\texttt{card} \mid x, C) &&\\
&\quad \times P(\texttt{\_in} \mid \texttt{card}, x, C) &&\\
&\quad \times P(\texttt{\_transit} \mid \texttt{card\_in}, x, C) &&
\end{flalign*}

\section{Training Details}

\subsection{Data Augmentation}
\label{sec:train-data-aug}
This section provides the precise temperature schedules for the two data augmentation strategies described in \S\ref{sec:data-aug}. In both cases, let $\bar{c}$ denote the mean prediction confidence across all positions in the original (unaugmented) trajectory.

\subsubsection{Prediction Confidence Augmentation}

We define a sharpness parameter
\begin{equation}
\beta =
\begin{cases}
4, & \text{if } \bar{c} < 0.95, \\
5, & \text{otherwise},
\end{cases}
\end{equation}
and compute the upper bound of the temperature range as
\begin{equation}
T_{\max} = 1 + \max\!\left(0,\; \tanh\!\left(\beta\,(\bar{c} - 0.5)\right)\right).
\end{equation}
A temperature $T \sim \mathcal{U}[1, T_{\max}]$ is sampled and applied to the logits $\mathbf{z}_t$ at every position: $\mathbf{z}_t \leftarrow \mathbf{z}_t / T$. Features from \S\ref{sec:feat}, barring the correctness indicator, are then recomputed from the rescaled logits. The higher value of $\beta$ for very confident trajectories ($\bar{c} \geq 0.95$) ensures that near-certain predictions receive meaningful perturbation, while the $\tanh$ saturation prevents excessively large temperatures that would flatten the distribution beyond any realistic miscalibration regime.

\subsubsection{Label Augmentation}

For a given trajectory, we simulate a near-calibrated regime with probability $\bar{c}/2$ and a severely miscalibrated regime otherwise. Each regime defines its own temperature interval from which $T$ is sampled.

\xhdr{Near-calibrated regime}
We set
\begin{align*}
\beta &=
\begin{cases}
1, & \text{if } \bar{c} < 0.95, \\
2, & \text{otherwise},
\end{cases}
\\
T_{\max} &= 1 + \max\!\left(0,\; \tanh\!\left(\beta\,(\bar{c} - 0.5)\right)\right),
\end{align*}
and sample $T \sim \mathcal{U}[1, T_{\max}]$. The mild scaling keeps the sampling distribution close to the original predictions, producing synthetic labels that are largely consistent with the model's confidence.

\xhdr{Miscalibrated regime}
We set $\beta$ as above and compute a wider interval:
\begin{equation}
T_{\max} = 2 + 3\,\max\!\left(0.5,\; \tanh\!\left(\beta\,(\bar{c} - 0.4)\right)\right),
\end{equation}
with $T \sim \mathcal{U}[2, T_{\max}]$. The substantially larger temperatures flatten the predictive distribution, generating labels that frequently disagree with the model's stated confidence.

In both regimes, the augmented labels are obtained by sampling from the temperature-scaled softmax at each position:
\begin{equation}
\tilde{y}_t \sim \mathrm{Categorical}\!\left(\mathrm{softmax}\!\left(\mathbf{z}_t / T\right)\right),
\end{equation}
and features 3–5 from \S\ref{sec:feat} are recomputed using $\tilde{y}_t$ in place of the original ground-truth labels.

\subsection{Training with embeddings}
For the ablations that test beneficiality of embeddings, we use \texttt{google/embeddinggemma-300m} with an embedding dimension of 128. In order to promote the model relying on ICL instead of IWL (In Weights Learning) to predict likelihood of correctness of prediction, during training, we permute each embedding within each sequence of inputs by a random but shared order, and this order is different for each sequence. We do this only during the data loading phase. At test time, we used the embeddings as is, without permutation. The random permutations during training ensure that:

\begin{enumerate}
    \item The transformer does not have access to a consistent representation of the sentences to memorize and use for IWL.
    \item The transformer can only learn to calculate similarity at the current position with its in context examples.
    \item One may potentially use any arbitrary embedding model to generate the embeddings, since we only need to use them for calculating similarities - so long as the embedding dimension is the same as the one used during training(in our case, 128). This better fits our theme of a universal calibrator that uses general features. 
\end{enumerate}

When trained without permuting the embeddings this way, the trained calibrator overfits to the training dataset and generalized poorly to unseen datasets, as seen in Table~\ref{tab:ablation}. Nevertheless, even with the permutation trick, the calibration performance is slightly worse than not using embeddings. We suspect it is because even high cosine similarity often signals superficial similarity that isn't predictive enough of likelihood of correctness of the sentences being similar.

\subsection{Additional Training Details}
For generating the training datasets for \sysname\ for each training dataset-LLM combination in table~\ref{tab:all_datasets}, we generate 5000 sequences of autoregressive predictions over the in-context examples, each of length 21 in order to learn up to 20 shot calibration, comprising at each step of features described in Section~\ref{sec:feat}.

For the checkpoint selection and ablation evaluations, we generate 1000 sequences for each validation/unseen dataset respectively in table~\ref{tab:all_datasets}. During training, a checkpoint is saved whenever \textit{both} calibrated ECE and Brier scores improve for the validation datasets over the last saved checkpoint.

In all the generated calibration training, validation and test (unseen) datasets, the sequences are equally divided between both the demonstration sampling strategies. 

During training, at each iteration, we sample a fraction of the training LLM-dataset-sampling strategy combinations based on the \texttt{dataset dropout} hyperparameter reported in table ~\ref{tab:calibrator-hyperparams}. This prevents certain combinations from dominating the loss and helps regularise the training of the calibrator for better generalisation 

\subsection{Calibrator Training Hyperparameters}
We train the $0.6M$ parameter \sysname\ on a single Nvidia B200 with the hyperparameters listed in Table~\ref{tab:calibrator-hyperparams}. Training a single checkpoint takes less than 30 mins on a single B200 with our data volume.

For gradient clipping, we adopt the AutoClip strategy ~\cite{seetharaman2020autoclip},
which sets the clipping threshold adaptively as the $p$-th percentile of the gradient
norm history observed during training. We use $p=20$ with a history window of $40k$ iterations. 

The choices of weight decay parameters and architectural specifications were inspired by recent work \cite{singh-transient-icl-emergence}, which suggests that a weight decay on the MLP layers encourages ICL emergence over In-Weights Learning/memorization.

\begin{table}[H]
\centering
\footnotesize
\begin{tabular}{lll}
\toprule
\textbf{Category} & \textbf{Hyperparameter} & \textbf{Value} \\
\midrule
\textit{Architecture} & Embedding dim & 64 \\
 & Attention heads & 8 \\
 & Layers & 12 \\
 & Sequence length & 21 \\
 & Dropout & 0.25 \\
\addlinespace
\textit{Optimization} & Optimizer & AdamW \\
 & Learning rate & $1 \times 10^{-5}$ \\
 & Weight decay (attention) & 0.00 \\
 & Weight decay (MLP) & 0.05 \\
 & Weight decay (other) & 0.02 \\
 & Gradient clipping & 20th \\
 & gradnorm history percentile & \\
\addlinespace
\textit{Training} & Iterations & 30{,}000 \\
 & Batch size & 16 \\
 & Evaluation interval & 800 \\
 & Dataset dropout & 0.8 \\
\addlinespace
\textit{Augmentation} & Temperature augmentation & True \\
 & Label augmentation & True \\
\bottomrule
\end{tabular}
\caption{Training hyperparameters for \sysname.}
\label{tab:calibrator-hyperparams}
\end{table}

\section{Evaluation Metrics}
\label{sec:metrics}
To assess calibration quality, we compute Expected Calibration Error (ECE) on the evaluation datasets using smooth ECE \cite{smoothECE} which produces more reliable estimates by performing kernel smoothing over the prediction-label pairs. This provides a smoothed estimate instead of the following standard binned approach:
\[
\text{ECE} = \sum_{m=1}^{M} \frac{|B_m|}{n} \left| \text{acc}(B_m) - \text{conf}(B_m) \right|,
\]
where $B_m$ are confidence bins, $\text{acc}(B_m)$ is the empirical accuracy, and $\text{conf}(B_m)$ is the mean predicted confidence in each bin. Additionally, we calculate brier score ~\cite{brier} defined as

\[
\text{BS} = \frac{1}{N} \sum_{i=1}^{N} \sum_{j=1}^{C} \left(\hat{p}_{ij} - y_{ij}\right)^2
\]
where $N$ is the number of samples, $C$ is the number of classes, $\hat{p}_{ij}$ is the predicted probability for class $j$ on sample $i$, and $y_{ij} \in \{0,1\}$ is the one-hot encoded true label.

\section{Additional Experimental Results}
\label{sec:add-exp-res}

\subsection{Tables}
In addition to ECE, we report the Brier score for all the LLMs in tables~\ref{tab:open-llms-brier} and~\ref{tab:gemini-brier}. 
The accuracy across datasets for Phi-4-mini-instruct for both the demonstration sampling settings can be found in Tables \ref{tab:accuracy_similarity_phi} and \ref{tab:accuracy_random_phi}. The variances of the ECEs and Brier scores for all open weights LLMs are in tables 11-16.

\begin{table*}[h]
\centering
\footnotesize
\setlength\tabcolsep{2.0pt}
\resizebox{\textwidth}{!}{
\begin{tabular}{l | cccccc | cccc | cccc}
\toprule
& \multicolumn{6}{c|}{\texttt{Phi-4-mini-instruct}} & \multicolumn{4}{c|}{\texttt{Qwen-3.5-9B}} & \multicolumn{4}{c}{\texttt{Gemma-4-E4B}} \\
\cmidrule(lr){2-7} \cmidrule(lr){8-11} \cmidrule(lr){12-15}
Dataset & Uncalib & \ICC & \Permute & \ICTemp & \FSICTemp & Ours & Uncalib & \ICTemp & \FSICTemp & Ours & Uncalib & \ICTemp & \FSICTemp & Ours \\
\midrule
\multicolumn{15}{c}{\textbf{Random Sampling}} \\
\midrule
\amazonc & .366 & .481 & \textbf{.303} & .348 & .350 & \underline{.319} & \textbf{.149} & .174 & .175 & \underline{.158} & .253 & .250 & \underline{.247} & \textbf{.219} \\
\commons & .492 & .531 & .471 & .458 & .456 & \textbf{.439} & \textbf{.234} & .253 & .254 & \textbf{.234} & .393 & .360 & \underline{.359} & \textbf{.344} \\
massive\_intent & .637 & .766 & \textbf{.587} & .600 & .601 & \underline{.592} & \underline{.320} & .327 & .326 & \textbf{.306} & .366 & .353 & \underline{.350} & \textbf{.333} \\
medqa & .685 & .704 & .675 & \underline{.642} & .643 & \textbf{.621} & \textbf{.326} & .353 & .352 & \underline{.327} & .624 & \underline{.532} & .533 & \textbf{.525} \\
metatool & .586 & .677 & .545 & .547 & \underline{.542} & \textbf{.532} & .284 & \underline{.283} & .285 & \textbf{.264} & .345 & \underline{.315} & .316 & \textbf{.307} \\
when2call & .814 & .817 & .788 & \underline{.657} & \textbf{.655} & .661 & \underline{.423} & \underline{.423} & .427 & \textbf{.398} & .490 & .440 & \underline{.438} & \textbf{.421} \\
wikitoxic & .487 & .474 & \underline{.429} & .458 & .451 & \textbf{.421} & \underline{.222} & .244 & .244 & \textbf{.203} & .261 & \underline{.243} & .245 & \textbf{.228} \\
wildguard & 1.1 & 1.1 & 1.1 & \textbf{.887} & \underline{.891} & .892 & \underline{.505} & .510 & .511 & \textbf{.496} & .519 & \underline{.481} & .482 & \textbf{.469} \\
\midrule
Average & .646 & .694 & .612 & .575 & \underline{.574} & \textbf{.560} & \underline{.308} & .321 & .322 & \textbf{.298} & .406 & .372 & \underline{.371} & \textbf{.356} \\
\midrule
\multicolumn{15}{c}{\textbf{Similarity Sampling}} \\
\midrule
\amazonc & .401 & - & \underline{.347} & .364 & .365 & \textbf{.343} & \textbf{.155} & .183 & .183 & \underline{.165} & .366 & .333 & \underline{.321} & \textbf{.300} \\
\commons & .475 & - & \textbf{.428} & .464 & .466 & \underline{.437} & \underline{.248} & .265 & .266 & \textbf{.247} & .363 & .335 & \underline{.334} & \textbf{.317} \\
massive\_intent & .271 & - & \textbf{.242} & .268 & .270 & \underline{.256} & \underline{.200} & .205 & .208 & \textbf{.196} & .226 & \underline{.223} & \underline{.223} & \textbf{.206} \\
medqa & .610 & - & \underline{.594} & .596 & .596 & \textbf{.571} & \textbf{.282} & .313 & .314 & \underline{.289} & .539 & .484 & \underline{.482} & \textbf{.468} \\
metatool & .244 & - & \textbf{.224} & .238 & .236 & \underline{.230} & .170 & \underline{.165} & .167 & \textbf{.157} & .194 & .177 & \underline{.176} & \textbf{.169} \\
when2call & .763 & - & .717 & \textbf{.599} & \underline{.601} & .610 & .263 & \underline{.259} & \underline{.259} & \textbf{.250} & .319 & \underline{.277} & \underline{.277} & \textbf{.260} \\
wikitoxic & .373 & - & \underline{.337} & .354 & .346 & \textbf{.325} & \underline{.213} & .214 & .215 & \textbf{.191} & .259 & \underline{.231} & .232 & \textbf{.222} \\
wildguard & .668 & - & .623 & \underline{.615} & .631 & \textbf{.602} & .437 & \underline{.423} & .428 & \textbf{.402} & .461 & \underline{.416} & .421 & \textbf{.404} \\
\midrule
Average & .476 & - & .439 & \underline{.437} & .439 & \textbf{.422} & \underline{.246} & .253 & .255 & \textbf{.237} & .341 & .309 & \underline{.308} & \textbf{.293} \\
\bottomrule
\end{tabular}
}
\caption{Brier Scores for calibration methods across the open weights LLMs}
\label{tab:open-llms-brier}
\end{table*}

\begin{table}[h]
\centering
\footnotesize
\setlength\tabcolsep{4pt}
\begin{tabular}{l c c c c}
\toprule
& Uncalib & \ICTemp & \FSICTemp & Ours \\
Datasets & R / S & R / S & R / S & R / S \\
\midrule
\amazonc               & \underline{.170}/.194 & .181/\underline{.189} & .188/.193 & \textbf{.157}/\textbf{.175} \\
\commons               & \underline{.241}/\underline{.235} & .262/.251 & .260/.249 & \textbf{.224}/\textbf{.220} \\
medqa                  & \underline{.226}/\underline{.196} & .267/.248 & .235/.213 & \textbf{.207}/\textbf{.191} \\
when2call              & .349/.207 & \underline{.333}/\underline{.198} & \underline{.333}/.199 & \textbf{.305}/\textbf{.181} \\
\midrule
Average                & \underline{.246}/\underline{.208} & .261/.221 & .254/.214 & \textbf{.223}/\textbf{.192} \\
\bottomrule
\end{tabular}
\caption{Comparison of Brier scores on the \texttt{Gemini-2.5-flash} LLM. Values are reported as Random (R) / Similarity (S) sampled contexts.}
\label{tab:gemini-brier}
\end{table}

\begin{table}[h]
\footnotesize
\setlength\tabcolsep{2.0pt}
\begin{tabular}{lccccc}
\toprule
                       & \multicolumn{5}{c}{Calibration Methods}                     \\
\cmidrule{2-6}
Datasets               & Uncalib  & \Permute & \ICTemp & \FSICTemp & Ours \\
\midrule
\amazonc               & 76.6          & 77.1          & 76.6  & 76.6  & 76.6        \\
\commons               & 70.6          & 71.5          & 70.6  & 70.6  & 70.6        \\
massive\_intent        & 83.0          & 84.0          & 83.0  & 83.0  & 83.0        \\
medqa                  & 57.1          & 57.8          & 57.1  & 57.1  & 57.1        \\
metatool               & 84.4          & 85.5          & 84.4  & 84.4  & 84.4        \\
when2call              & 53.3          & 54.8          & 53.3  & 53.3  & 53.3        \\
wikitoxic              & 78.4          & 79.0          & 78.4  & 78.4  & 78.4        \\
wildguard              & 55.2          & 56.2          & 55.2  & 55.2  & 55.2        \\
\midrule
Average                & 69.8          & 70.7          & 69.8  & 69.8  & 69.8        \\
\bottomrule
\end{tabular}
\caption{Accuracy comparison for calibration methods | Phi-4-mini-instruct | Similarity Sampling}
\label{tab:accuracy_similarity_phi}
\end{table}

\begin{table}[h]
\footnotesize
\setlength\tabcolsep{2.0pt}
\begin{tabular}{lcccccc}
\toprule
                       & \multicolumn{6}{c}{Calibration Methods}                           \\
\cmidrule{2-7}
Datasets               & Uncalib & \ICC  & \Permute & \ICTemp & \FSICTemp & Ours \\
\midrule
\amazonc               & 78.0         & 69.6  & 79.7          & 78.0  & 78.0  & 78.0        \\
\commons               & 69.7         & 67.7  & 70.1          & 69.7  & 69.7  & 69.7        \\
massive\_intent        & 58.5         & 51.1  & 60.2          & 58.5  & 58.5  & 58.5        \\
medqa                  & 52.5         & 51.2  & 52.4          & 52.5  & 52.5  & 52.5        \\
metatool               & 61.8         & 55.8  & 63.0          & 61.8  & 61.8  & 61.8        \\
when2call              & 52.6         & 50.1  & 53.5          & 52.6  & 52.6  & 52.6        \\
wikitoxic              & 72.6         & 72.7  & 73.7          & 72.6  & 72.6  & 72.6        \\
wildguard              & 26.4         & 31.1  & 26.2          & 26.4  & 26.4  & 26.4        \\
\midrule
Average                & 59.0         & 56.2  & 59.8          & 59.0  & 59.0  & 59.0        \\
\bottomrule
\end{tabular}
\caption{Accuracy comparison for calibration methods | Phi-4-mini-instruct | Random Sampling}
\label{tab:accuracy_random_phi}
\end{table}

\begin{table*}[h]
\centering
\small
\begin{tabular}{lcccccc}
\toprule
Dataset & Uncalibrated & \ICC & \Permute & \ICTemp & \FSICTemp & \sysname \\
\midrule
\multicolumn{7}{c}{Random Sampling} \\
\midrule
\amazonc & 15.3 (1.8) & 20.6 (2.2) & 8.4 (1.7) & 7.8 (1.6) & \underline{7.6 (1.4)} & \textbf{4.7 (1.0)} \\
\commons & 21.8 (1.8) & 23.7 (2.3) & 19.3 (1.6) & 7.9 (1.5) & \underline{7.4 (1.5)} & \textbf{5.0 (1.0)} \\
massive\_intent & 21.2 (2.0) & 28.9 (2.5) & 15.3 (1.6) & \underline{8.8 (2.0)} & 8.9 (2.0) & \textbf{4.9 (0.7)} \\
medqa & 21.8 (1.4) & 23.9 (1.9) & 21.2 (1.3) & \underline{8.6 (1.6)} & 8.8 (1.7) & \textbf{6.6 (1.2)} \\
metatool & 23.0 (2.2) & 27.4 (2.1) & 17.9 (2.3) & 7.1 (1.4) & \underline{6.8 (1.4)} & \textbf{5.1 (1.0)} \\
when2call & 36.4 (2.1) & 34.8 (3.1) & 33.5 (2.1) & \underline{11.6 (1.9)} & \textbf{11.4 (1.9)} & 13.1 (2.0) \\
wikitoxic & 19.9 (2.7) & 20.0 (2.6) & 14.6 (2.6) & 12.6 (1.7) & \underline{12.3 (1.7)} & \textbf{9.0 (1.2)} \\
wildguard & 46.5 (2.5) & 43.0 (2.6) & 41.0 (2.6) & \textbf{8.1 (1.7)} & 9.5 (1.8) & \underline{9.3 (1.7)} \\
\midrule
Average & 25.7 (2.1) & 27.8 (2.4) & 21.4 (2.0) & \underline{9.1 (1.7)} & \underline{9.1 (1.7)} & \textbf{7.2 (1.3)} \\
\midrule
\multicolumn{7}{c}{Similarity Sampling} \\
\midrule
\amazonc & 17.1 (1.7) & - & 11.2 (1.8) & 8.6 (1.4) & \underline{8.2 (1.4)} & \textbf{6.0 (1.4)} \\
\commons & 18.3 (1.8) & - & 13.6 (1.6) & \underline{9.7 (1.5)} & 9.9 (1.4) & \textbf{4.8 (1.0)} \\
massive\_intent & 10.7 (1.6) & - & \underline{6.5 (1.1)} & 7.7 (1.3) & 8.1 (1.3) & \textbf{5.2 (0.9)} \\
medqa & 17.3 (2.1) & - & 14.8 (1.6) & \underline{8.8 (1.5)} & 8.9 (1.5) & \textbf{5.0 (1.0)} \\
metatool & 9.4 (1.6) & - & 5.9 (1.3) & 6.0 (1.2) & \underline{5.7 (1.3)} & \textbf{5.6 (1.0)} \\
when2call & 33.1 (2.8) & - & 28.3 (2.3) & \textbf{8.0 (1.5)} & \underline{8.5 (1.4)} & 10.5 (2.2) \\
wikitoxic & 15.9 (3.2) & - & 12.6 (2.8) & 10.2 (1.6) & \underline{9.3 (1.8)} & \textbf{6.5 (1.3)} \\
wildguard & 24.9 (2.0) & - & 19.9 (1.5) & \underline{8.0 (1.8)} & 9.9 (2.1) & \textbf{4.9 (0.7)} \\
\midrule
Average & 18.3 (2.2) & - & 14.1 (1.8) & \underline{8.4 (1.5)} & 8.6 (1.5) & \textbf{6.1 (1.3)} \\
\bottomrule
\end{tabular}
\caption{Calibration Error (ECE) for Phi-4-mini-instruct across calibration methods. Standard deviations reported in parentheses.}
\label{tab:ece_results}
\end{table*}

\begin{table*}
\centering
\small
\begin{tabular}{lcccccc}
\toprule
Dataset & Uncalibrated & \ICC & \Permute & \ICTemp & \FSICTemp & \sysname \\
\midrule
\multicolumn{7}{c}{Random Sampling} \\
\midrule
\amazonc & .366 (.033) & .481 (.034) & \textbf{.303 (.032)} & .348 (.029) & .350 (.027) & \underline{.319 (.026)} \\
\commons & .492 (.028) & .531 (.034) & .471 (.025) & .458 (.021) & \underline{.456 (.021)} & \textbf{.439 (.020)} \\
massive\_intent & .637 (.019) & .766 (.039) & \textbf{.587 (.016)} & .600 (.016) & .601 (.016) & \underline{.592 (.015)} \\
medqa & .685 (.024) & .704 (.027) & .675 (.024) & \underline{.642 (.023)} & .643 (.022) & \textbf{.621 (.022)} \\
metatool & .586 (.031) & .677 (.033) & .545 (.028) & .547 (.026) & \underline{.542 (.023)} & \textbf{.532 (.022)} \\
when2call & .814 (.034) & .817 (.043) & .788 (.034) & \underline{.657 (.018)} & \textbf{.655 (.017)} & .661 (.025) \\
wikitoxic & .487 (.044) & .474 (.041) & \underline{.429 (.037)} & .458 (.026) & .451 (.027) & \textbf{.421 (.027)} \\
wildguard & 1.138 (.038) & 1.069 (.040) & 1.067 (.032) & \textbf{.887 (.013)} & \underline{.891 (.011)} & .892 (.017) \\
\midrule
Average & .646 (.031) & .694 (.036) & .612 (.029) & .575 (.022) & \underline{.574 (.021)} & \textbf{.560 (.022)} \\
\midrule
\multicolumn{7}{c}{Similarity Sampling} \\
\midrule
\amazonc & .401 (.033) & - & \underline{.347 (.029)} & .364 (.030) & .365 (.026) & \textbf{.343 (.026)} \\
\commons & .475 (.034) & - & \textbf{.428 (.028)} & .464 (.027) & .466 (.025) & \underline{.437 (.025)} \\
massive\_intent & .271 (.020) & - & \textbf{.242 (.016)} & .268 (.015) & .270 (.016) & \underline{.256 (.015)} \\
medqa & .610 (.027) & - & \underline{.594 (.022)} & .596 (.020) & .596 (.017) & \textbf{.571 (.019)} \\
metatool & .244 (.030) & - & \textbf{.224 (.025)} & .238 (.026) & .236 (.026) & \underline{.230 (.024)} \\
when2call & .763 (.044) & - & .717 (.036) & \textbf{.599 (.018)} & \underline{.601 (.020)} & .610 (.025) \\
wikitoxic & .373 (.052) & - & \underline{.337 (.045)} & .354 (.036) & .346 (.038) & \textbf{.325 (.035)} \\
wildguard & .668 (.028) & - & .623 (.026) & \underline{.615 (.020)} & .631 (.024) & \textbf{.602 (.021)} \\
\midrule
Average & .476 (.035) & - & .439 (.030) & \underline{.437 (.025)} & .439 (.025) & \textbf{.422 (.024)} \\
\bottomrule
\end{tabular}
\caption{Brier Scores for Phi-4-mini-instruct across  calibration methods. Standard deviations reported in parentheses.}
\label{tab:brier_results}
\end{table*}

\begin{table*}
\centering
\small
\begin{tabular}{lcccc}
\toprule
Dataset & Uncalibrated & \ICTemp & \FSICTemp & \sysname \\
\midrule
\multicolumn{5}{c}{Similarity Sampling} \\
\midrule
\amazonc & \textbf{4.3 (0.8)} & 7.3 (0.9) & 7.3 (0.8) & \underline{6.3 (1.1)} \\
\commons & 6.9 (1.3) & \underline{6.7 (0.9)} & \underline{6.7 (0.9)} & \textbf{4.9 (1.3)} \\
massive\_intent & 7.1 (1.7) & \underline{6.8 (1.1)} & 7.2 (1.1) & \textbf{4.8 (1.2)} \\
medqa & \textbf{4.4 (0.7)} & 6.3 (1.3) & 6.5 (1.4) & \underline{5.6 (0.9)} \\
metatool & 7.6 (1.2) & \underline{4.4 (0.9)} & 4.5 (1.0) & \textbf{3.8 (0.6)} \\
when2call & 7.2 (1.6) & \underline{5.1 (1.1)} & 5.5 (1.1) & \textbf{4.6 (0.9)} \\
wikitoxic & 10.4 (2.0) & \underline{7.2 (1.2)} & 7.4 (1.2) & \textbf{3.4 (0.7)} \\
wildguard & 18.4 (1.8) & \underline{9.0 (1.9)} & 9.7 (1.7) & \textbf{5.9 (1.4)} \\
\midrule
Average & 8.3 (1.5) & \underline{6.6 (1.2)} & 6.8 (1.2) & \textbf{4.9 (1.0)} \\
\midrule
\multicolumn{5}{c}{Random Sampling} \\
\midrule
\amazonc & \textbf{4.9 (1.0)} & 7.0 (1.0) & 7.2 (0.9) & \underline{5.4 (0.9)} \\
\commons & 6.6 (1.4) & \underline{6.1 (1.3)} & 6.3 (1.4) & \textbf{5.0 (1.1)} \\
massive\_intent & 10.7 (1.4) & \underline{7.0 (1.1)} & 7.1 (1.3) & \textbf{4.5 (0.6)} \\
medqa & \underline{6.0 (1.3)} & 6.7 (1.5) & 6.6 (1.3) & \textbf{5.0 (1.0)} \\
metatool & 12.8 (1.5) & \underline{6.2 (1.6)} & 6.4 (1.5) & \textbf{3.9 (0.7)} \\
when2call & 14.2 (1.6) & \underline{7.6 (1.5)} & 8.0 (1.4) & \textbf{4.7 (0.8)} \\
wikitoxic & 10.4 (1.9) & 7.0 (1.2) & \underline{6.9 (1.4)} & \textbf{3.4 (0.6)} \\
wildguard & 16.3 (2.0) & \underline{11.1 (1.7)} & 11.5 (1.4) & \textbf{7.7 (1.3)} \\
\midrule
Average & 10.2 (1.5) & \underline{7.3 (1.4)} & 7.5 (1.3) & \textbf{4.9 (0.9)} \\
\bottomrule
\end{tabular}
\caption{Calibration Error (ECE) for Qwen3.5-9B across calibration methods. Standard deviations reported in parentheses.}
\label{tab:qwen_ece_results}
\end{table*}

\begin{table*}
\centering
\small
\begin{tabular}{lcccc}
\toprule
Dataset & Uncalibrated & \ICTemp & \FSICTemp & \sysname \\
\midrule
\multicolumn{5}{c}{Random Sampling} \\
\midrule
\amazonc & \textbf{.149 (.025)} & .174 (.025) & .175 (.026) & \underline{.158 (.021)} \\
\commons & \underline{.234 (.022)} & .253 (.029) & .254 (.028) & \textbf{.234 (.018)} \\
massive\_intent & \underline{.320 (.024)} & .327 (.031) & .326 (.031) & \textbf{.306 (.025)} \\
medqa & \textbf{.326 (.018)} & .353 (.023) & .352 (.024) & \underline{.327 (.015)} \\
metatool & .284 (.022) & \underline{.283 (.022)} & .285 (.022) & \textbf{.264 (.019)} \\
when2call & \underline{.423 (.022)} & \underline{.423 (.024)} & .427 (.021) & \textbf{.398 (.019)} \\
wikitoxic & \underline{.222 (.028)} & .244 (.030) & .244 (.032) & \textbf{.203 (.021)} \\
wildguard & \underline{.505 (.031)} & .510 (.022) & .511 (.024) & \textbf{.496 (.022)} \\
\midrule
Average & \underline{.308 (.024)} & .321 (.026) & .322 (.026) & \textbf{.298 (.020)} \\
\midrule
\multicolumn{5}{c}{Similarity Sampling} \\
\midrule
\amazonc & \textbf{.155 (.016)} & .183 (.018) & .183 (.018) & \underline{.165 (.013)} \\
\commons & \underline{.248 (.021)} & .265 (.021) & .266 (.022) & \textbf{.247 (.016)} \\
massive\_intent & \underline{.200 (.031)} & .205 (.031) & .208 (.032) & \textbf{.196 (.028)} \\
medqa & \textbf{.282 (.017)} & .313 (.022) & .314 (.022) & \underline{.289 (.015)} \\
metatool & .170 (.016) & \underline{.165 (.015)} & .167 (.015) & \textbf{.157 (.013)} \\
when2call & .263 (.027) & \underline{.259 (.023)} & \underline{.259 (.023)} & \textbf{.250 (.020)} \\
wikitoxic & \underline{.213 (.029)} & .214 (.028) & .215 (.027) & \textbf{.191 (.022)} \\
wildguard & .437 (.037) & \underline{.423 (.035)} & .428 (.034) & \textbf{.402 (.033)} \\
\midrule
Average & \underline{.246 (.025)} & .253 (.025) & .255 (.025) & \textbf{.237 (.021)} \\
\bottomrule
\end{tabular}
\caption{Brier Scores for Qwen3.5-9B across calibration methods. Standard deviations reported in parentheses.}
\label{tab:qwen_brier_results}
\end{table*}

\begin{table*}
\centering
\small
\begin{tabular}{lcccc}
\toprule
Dataset & Uncalibrated & \ICTemp & \FSICTemp & \sysname \\
\midrule
\multicolumn{5}{c}{Random Sampling} \\
\midrule
\amazonc & 14.2 (1.8) & 5.8 (0.8) & \underline{5.7 (1.0)} & \textbf{3.4 (0.7)} \\
\commons & 20.7 (2.5) & 7.3 (1.6) & \underline{7.1 (1.4)} & \textbf{4.9 (1.3)} \\
massive\_intent & 17.2 (2.0) & 8.0 (1.3) & \underline{7.4 (0.8)} & \textbf{4.1 (0.5)} \\
medqa & 28.8 (3.0) & \textbf{7.2 (1.4)} & \textbf{7.2 (1.4)} & \underline{7.6 (2.4)} \\
metatool & 20.0 (1.8) & \underline{6.9 (1.4)} & \underline{6.9 (1.6)} & \textbf{4.9 (0.9)} \\
when2call & 25.6 (1.5) & \underline{9.0 (1.6)} & \underline{9.0 (1.4)} & \textbf{7.9 (1.1)} \\
wikitoxic & 15.2 (2.2) & 6.6 (1.3) & \underline{6.5 (1.0)} & \textbf{5.9 (1.5)} \\
wildguard & 25.3 (2.9) & 10.4 (1.9) & \underline{10.2 (1.9)} & \textbf{5.4 (1.1)} \\
\midrule
Average & 20.9 (2.3) & 7.6 (1.4) & \underline{7.5 (1.4)} & \textbf{5.5 (1.3)} \\
\midrule
\multicolumn{5}{c}{Similarity Sampling} \\
\midrule
\amazonc & 18.9 (1.8) & \underline{6.8 (1.2)} & \textbf{5.9 (1.1)} & \textbf{5.9 (1.3)} \\
\commons & 19.3 (1.5) & 7.0 (1.5) & \underline{6.8 (1.6)} & \textbf{4.1 (0.7)} \\
massive\_intent & 11.3 (2.3) & 8.0 (1.1) & \underline{7.7 (1.2)} & \textbf{3.9 (0.6)} \\
medqa & 24.7 (2.2) & 7.7 (1.4) & \underline{7.5 (1.5)} & \textbf{6.5 (1.3)} \\
metatool & 11.6 (1.9) & 5.6 (1.6) & \underline{5.3 (1.5)} & \textbf{3.4 (0.8)} \\
when2call & 17.9 (2.2) & \underline{6.3 (1.3)} & 6.6 (1.2) & \textbf{4.4 (0.9)} \\
wikitoxic & 14.8 (2.1) & 6.4 (1.5) & \underline{6.3 (1.4)} & \textbf{5.8 (1.6)} \\
wildguard & 25.1 (1.8) & \underline{8.0 (1.4)} & 8.6 (1.5) & \textbf{7.3 (1.4)} \\
\midrule
Average & 18.0 (2.0) & 7.0 (1.4) & \underline{6.8 (1.4)} & \textbf{5.2 (1.1)} \\
\bottomrule
\end{tabular}
\caption{Calibration Error (ECE) for gemma-4-E4B-it across calibration methods. Standard deviations reported in parentheses.}
\label{tab:gemma_ece_results}
\end{table*}

\begin{table*}
\centering
\small
\begin{tabular}{lcccc}
\toprule
Dataset & Uncalibrated & \ICTemp & \FSICTemp & \sysname \\
\midrule
\multicolumn{5}{c}{Random Sampling} \\
\midrule
\amazonc & .253 (.024) & .250 (.022) & \underline{.247 (.023)} & \textbf{.219 (.019)} \\
\commons & .393 (.040) & .360 (.030) & \underline{.359 (.031)} & \textbf{.344 (.032)} \\
massive\_intent & .366 (.036) & .353 (.032) & \underline{.350 (.033)} & \textbf{.333 (.028)} \\
medqa & .624 (.051) & \underline{.532 (.036)} & .533 (.034) & \textbf{.525 (.039)} \\
metatool & .345 (.026) & \underline{.315 (.020)} & .316 (.022) & \textbf{.307 (.019)} \\
when2call & .490 (.023) & .440 (.018) & \underline{.438 (.018)} & \textbf{.421 (.017)} \\
wikitoxic & .261 (.035) & \textbf{.243 (.031)} & \underline{.245 (.032)} & \textbf{.228 (.030)} \\
wildguard & .519 (.041) & \underline{.481 (.024)} & .482 (.023) & \textbf{.469 (.027)} \\
\midrule
Average & .406 (.036) & .372 (.027) & \underline{.371 (.028)} & \textbf{.356 (.027)} \\
\midrule
\multicolumn{5}{c}{Similarity Sampling} \\
\midrule
\amazonc & .366 (.031) & .333 (.028) & \underline{.321 (.023)} & \textbf{.300 (.024)} \\
\commons & .363 (.020) & .335 (.016) & \underline{.334 (.016)} & \textbf{.317 (.015)} \\
massive\_intent & .226 (.029) & \underline{.223 (.027)} & \underline{.223 (.026)} & \textbf{.206 (.022)} \\
medqa & .539 (.036) & .484 (.024) & \underline{.482 (.025)} & \textbf{.468 (.024)} \\
metatool & .194 (.026) & .177 (.014) & \underline{.176 (.014)} & \textbf{.169 (.015)} \\
when2call & .319 (.031) & \underline{.277 (.020)} & \underline{.277 (.019)} & \textbf{.260 (.019)} \\
wikitoxic & .259 (.038) & \textbf{.231 (.036)} & \underline{.232 (.036)} & \textbf{.222 (.032)} \\
wildguard & .461 (.037) & \underline{.416 (.031)} & .421 (.029) & \textbf{.404 (.031)} \\
\midrule
Average & .341 (.032) & .309 (.025) & \underline{.308 (.024)} & \textbf{.293 (.024)} \\
\bottomrule
\end{tabular}
\caption{Brier Scores for gemma-4-E4B-it across calibration methods. Standard deviations reported in parentheses.}
\label{tab:gemma_brier_results}
\end{table*}

\subsection{Plots}
The \texttt{Phi-4-mini-instruct} calibration performance of different methods across shots for all datasets can be found in Figures 4-11. 

\begin{figure*}[h]
    \begin{subfigure}{\textwidth}
        \includegraphics[width=\linewidth]{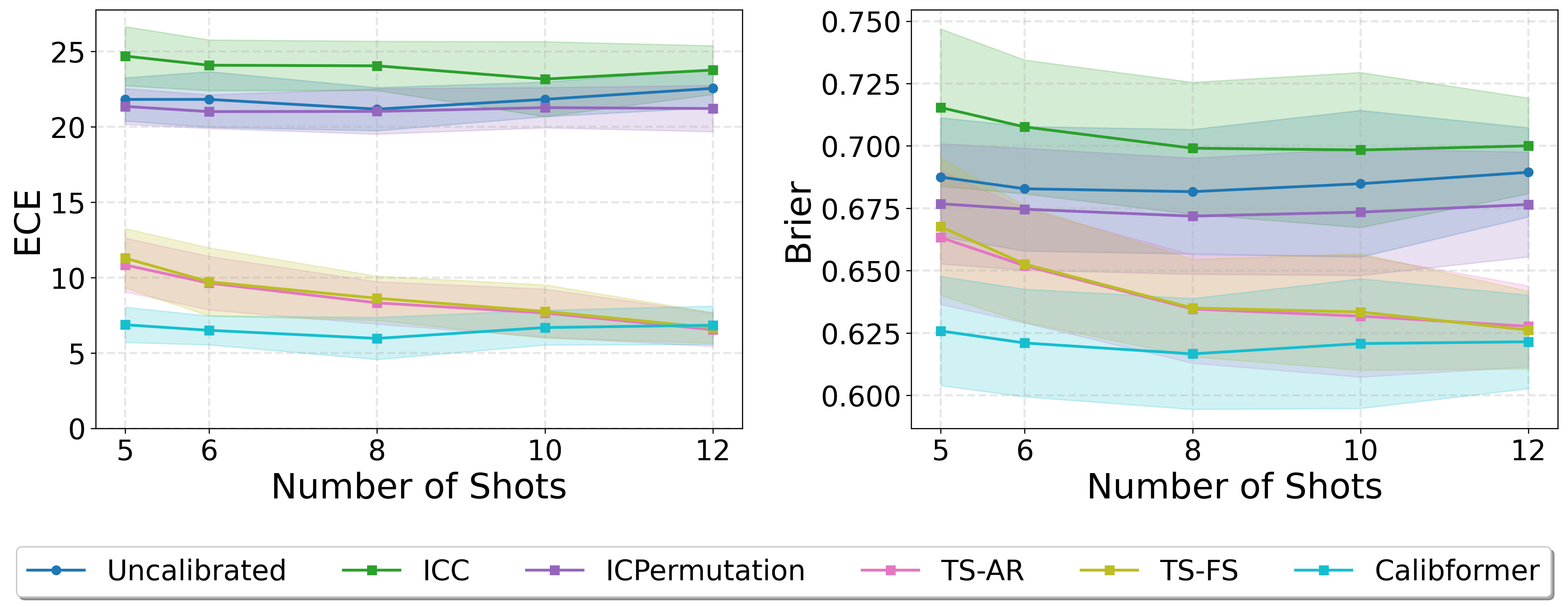}
        \caption{Random sampling}
    \end{subfigure}
    \vspace{0.5em}
    \begin{subfigure}{\textwidth}
        \includegraphics[width=\linewidth]{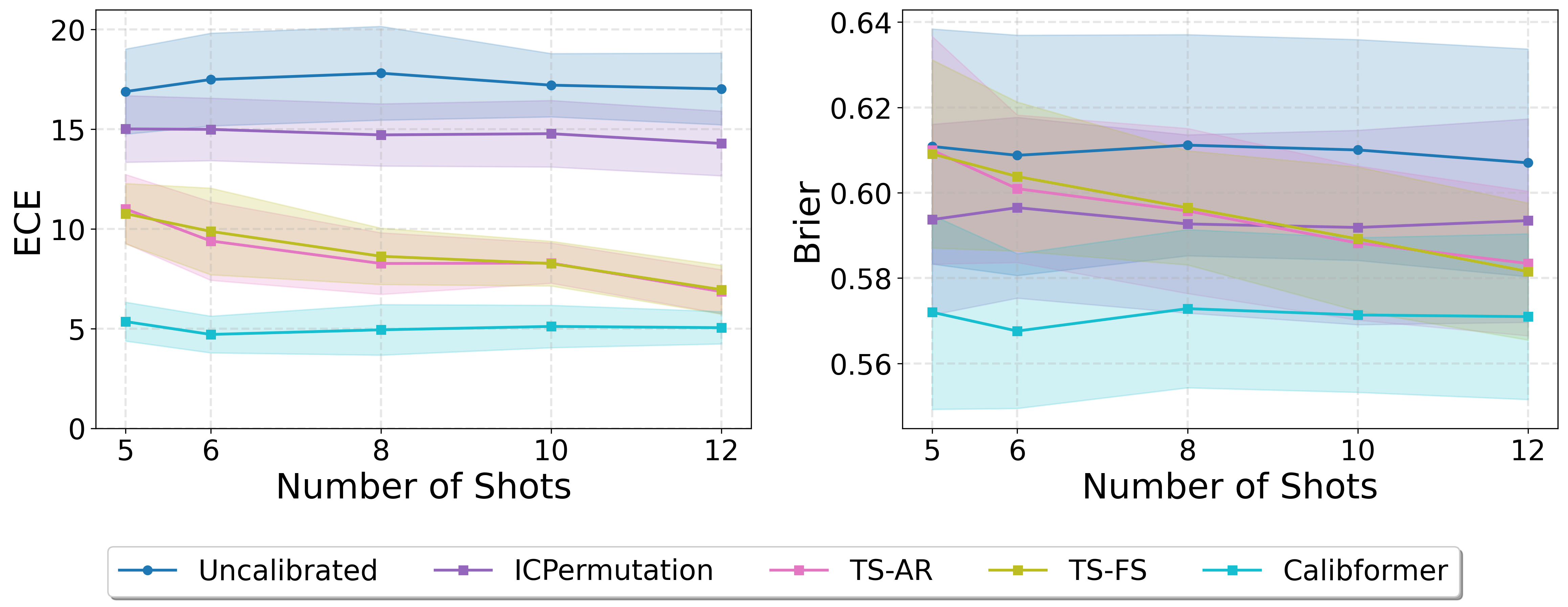}
        \caption{Similarity sampling}
    \end{subfigure}
    \caption{Calibration performance | Phi 4 mini instruct | MedQA}
    \label{fig:medqa}
\end{figure*}

\begin{figure*}[t]
    \begin{subfigure}{\textwidth}
        \includegraphics[width=\linewidth]{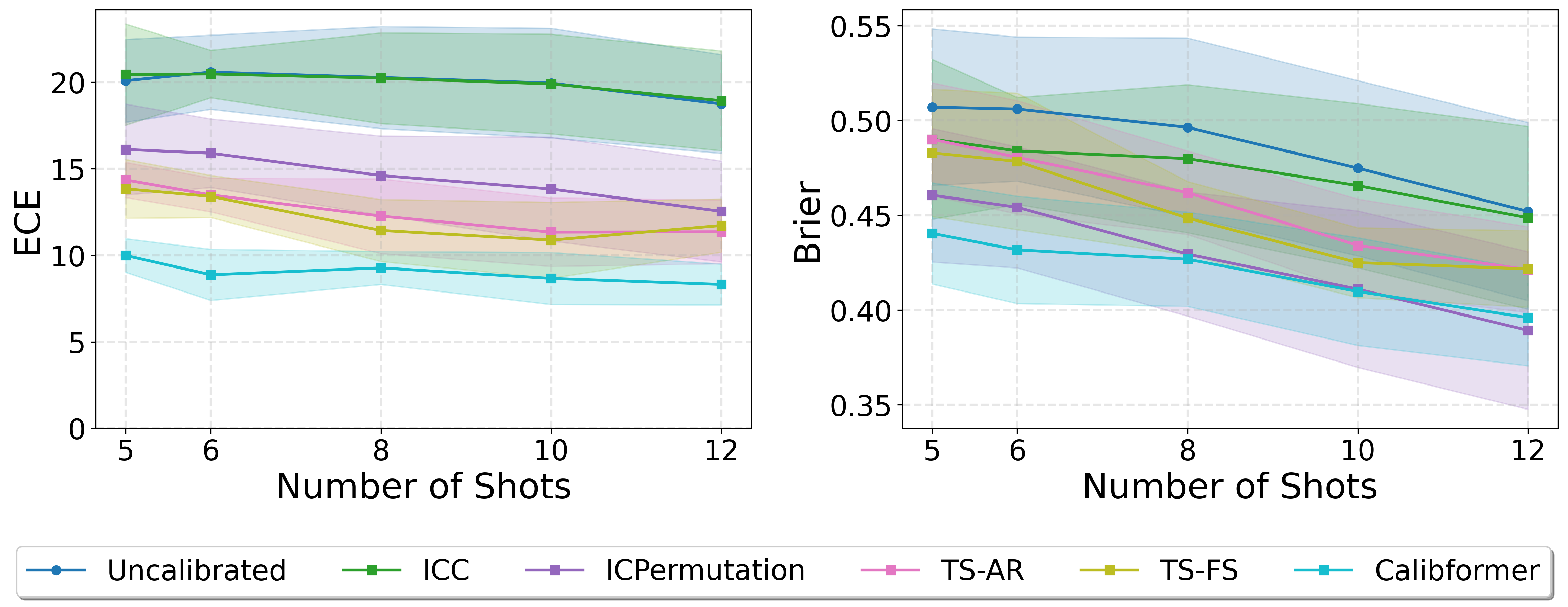}
        \caption{Random sampling}
    \end{subfigure}
    \vspace{0.5em}
    \begin{subfigure}{\textwidth}
        \includegraphics[width=\linewidth]{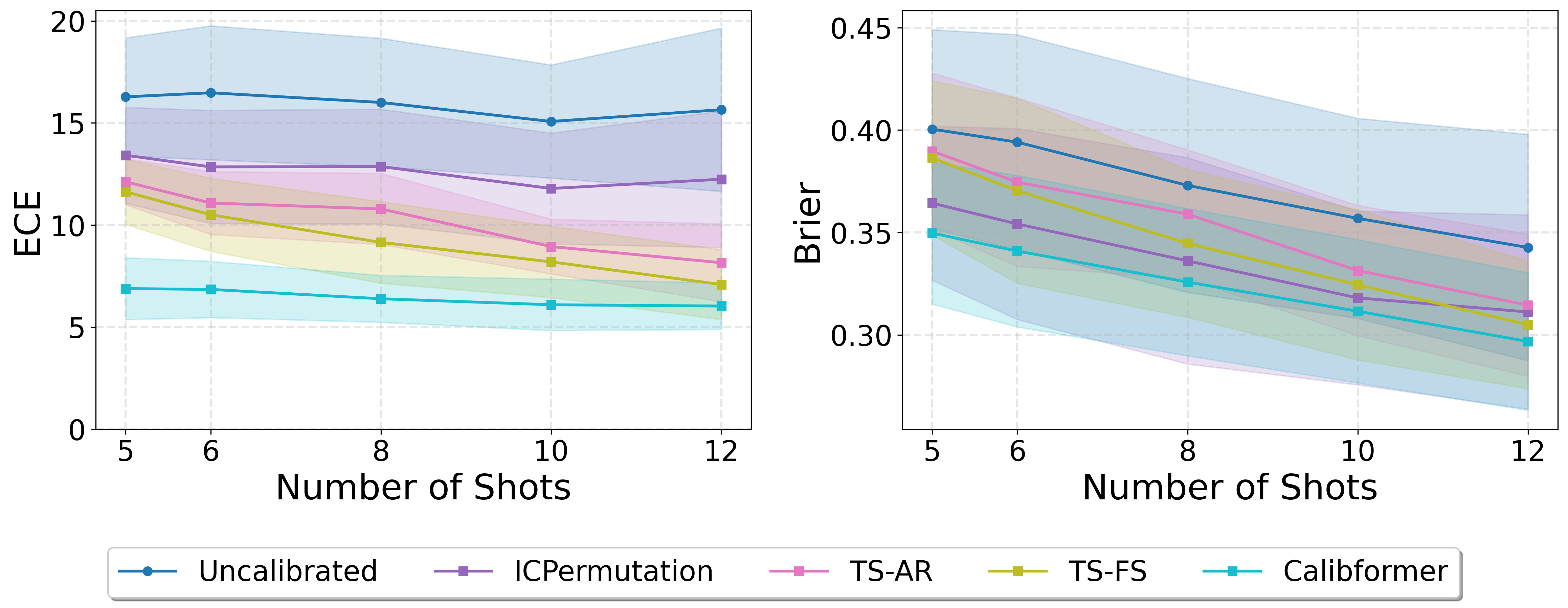}
        \caption{Similarity sampling}
    \end{subfigure}
    \caption{Calibration performance | Phi 4 mini instruct | Wikitoxic}
    \label{fig:wikitoxic}
\end{figure*}

\begin{figure*}[t]
    \begin{subfigure}{\textwidth}
        \includegraphics[width=\linewidth]{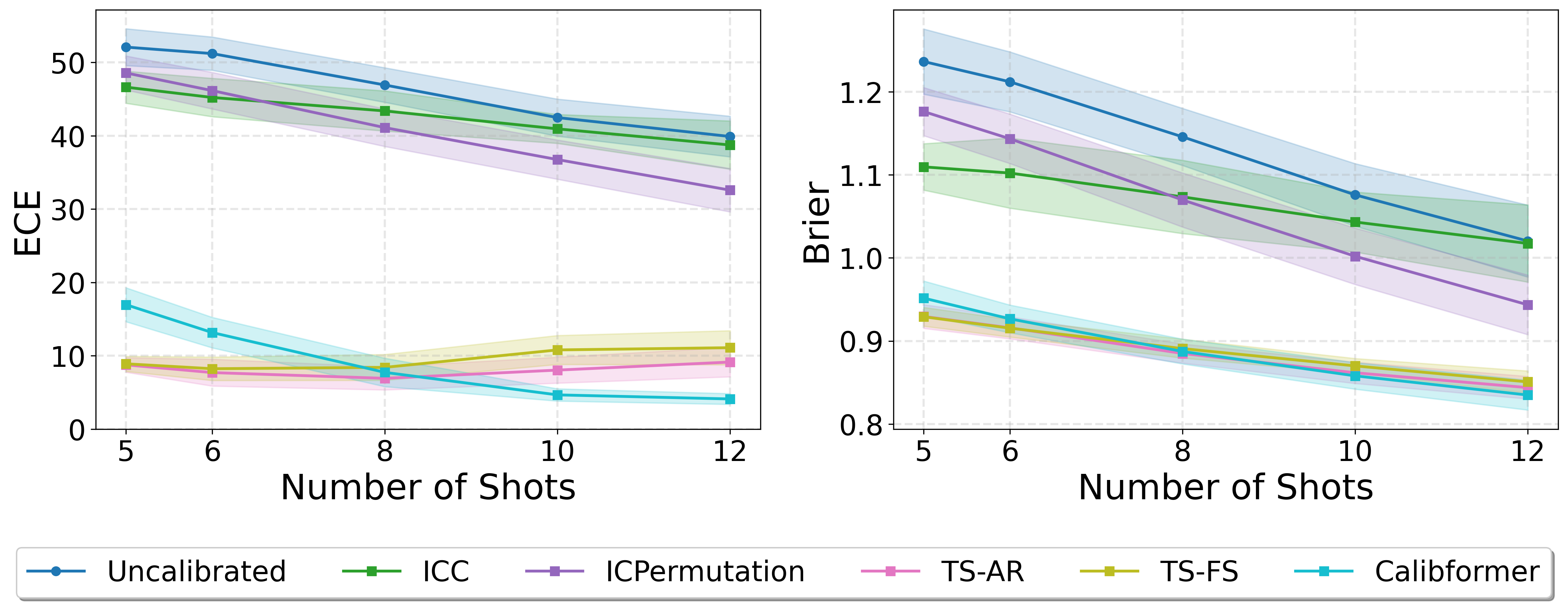}
        \caption{Random sampling}
    \end{subfigure}
    \vspace{0.5em}
    \begin{subfigure}{\textwidth}
        \includegraphics[width=\linewidth]{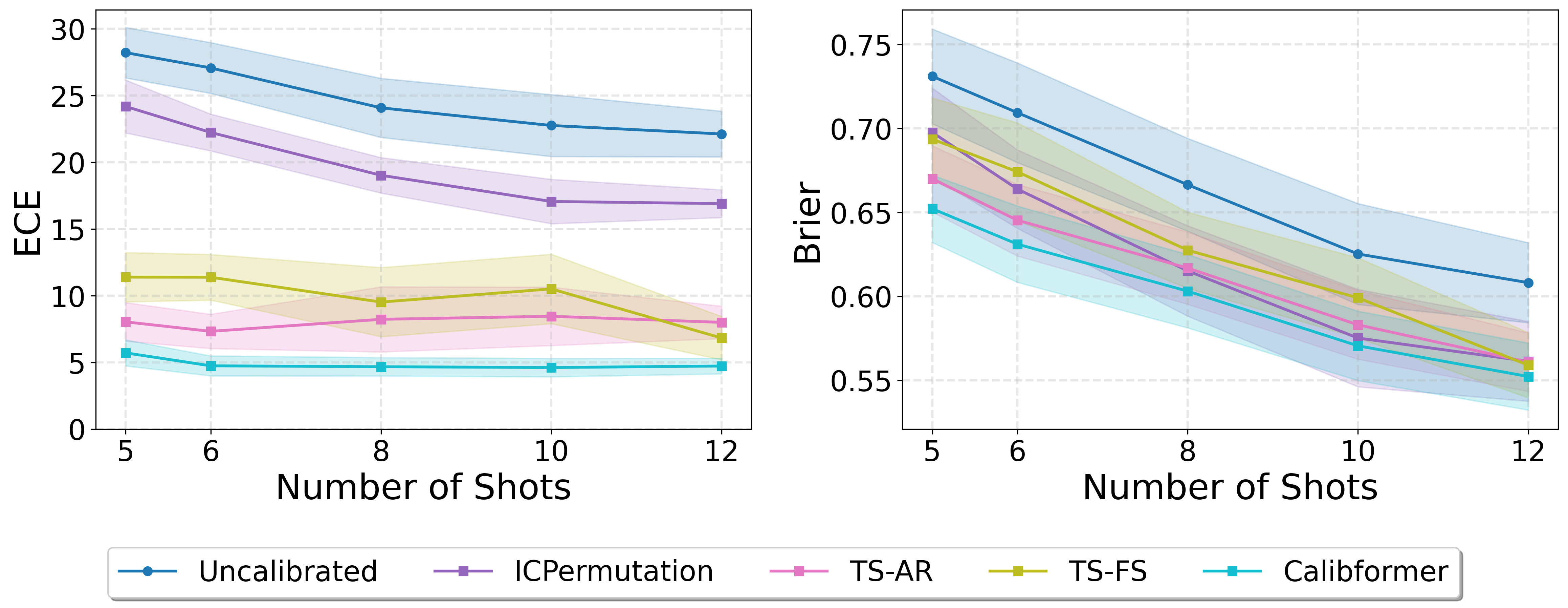}
        \caption{Similarity sampling}
    \end{subfigure}
    \caption{Calibration performance | Phi 4 mini instruct | Wildguard}
    \label{fig:wildguard}
\end{figure*}

\begin{figure*}[t]
    \begin{subfigure}{\textwidth}
        \includegraphics[width=\linewidth]{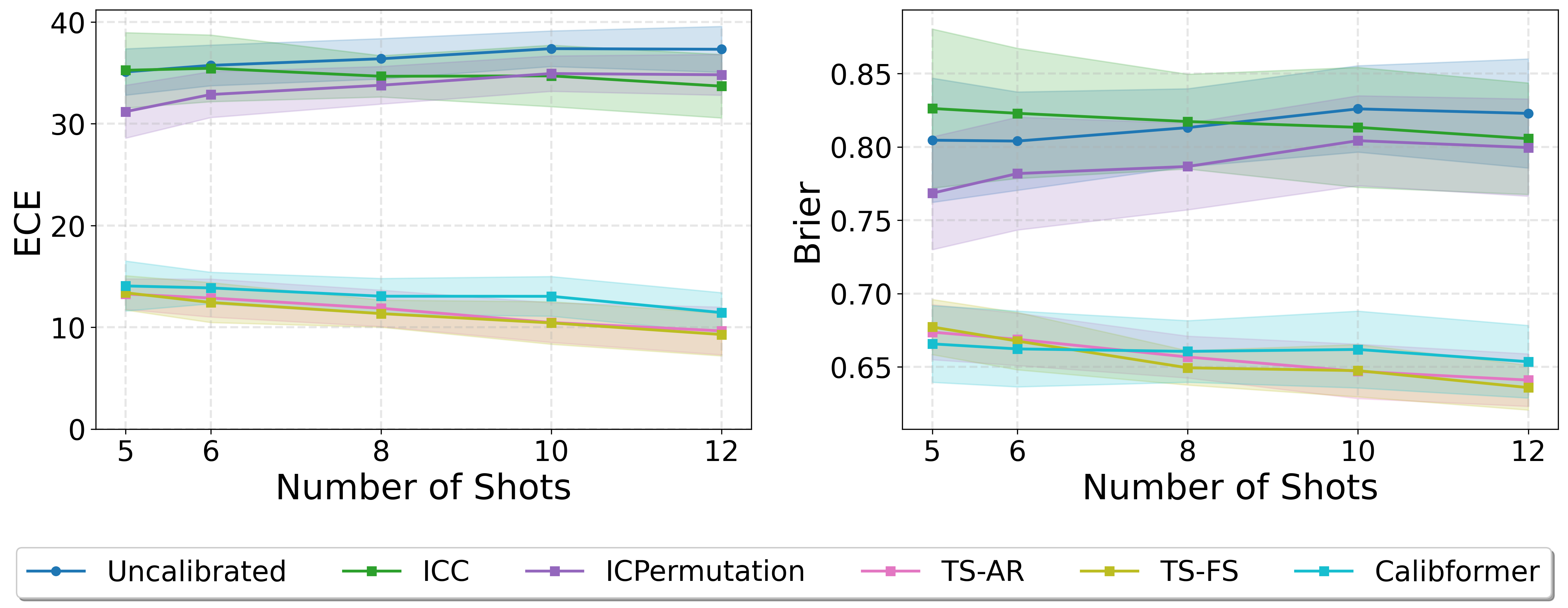}
        \caption{Random sampling}
    \end{subfigure}
    \vspace{0.5em}
    \begin{subfigure}{\textwidth}
        \includegraphics[width=\linewidth]{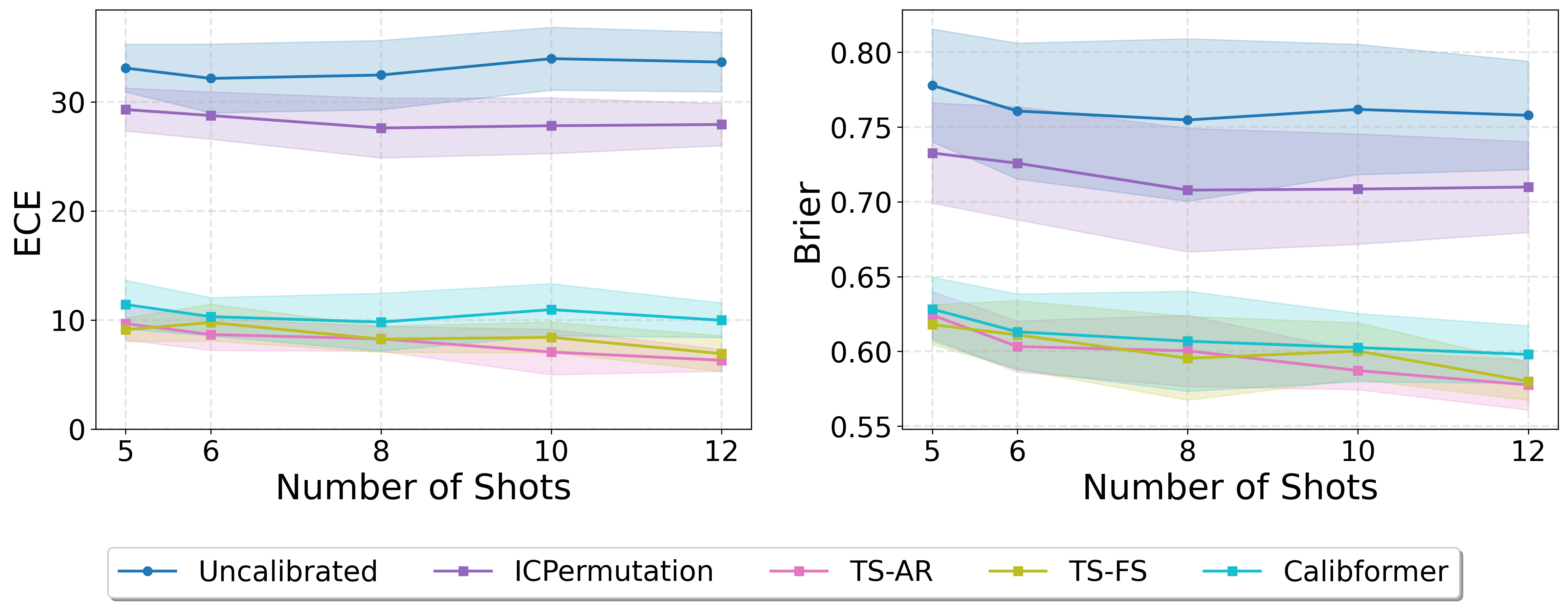}
        \caption{Similarity sampling}
    \end{subfigure}
    \caption{Calibration performance | Phi 4 mini instruct | When2call}
    \label{fig:when2call}
\end{figure*}

\begin{figure*}[t]
    \begin{subfigure}{\textwidth}
        \includegraphics[width=\linewidth]{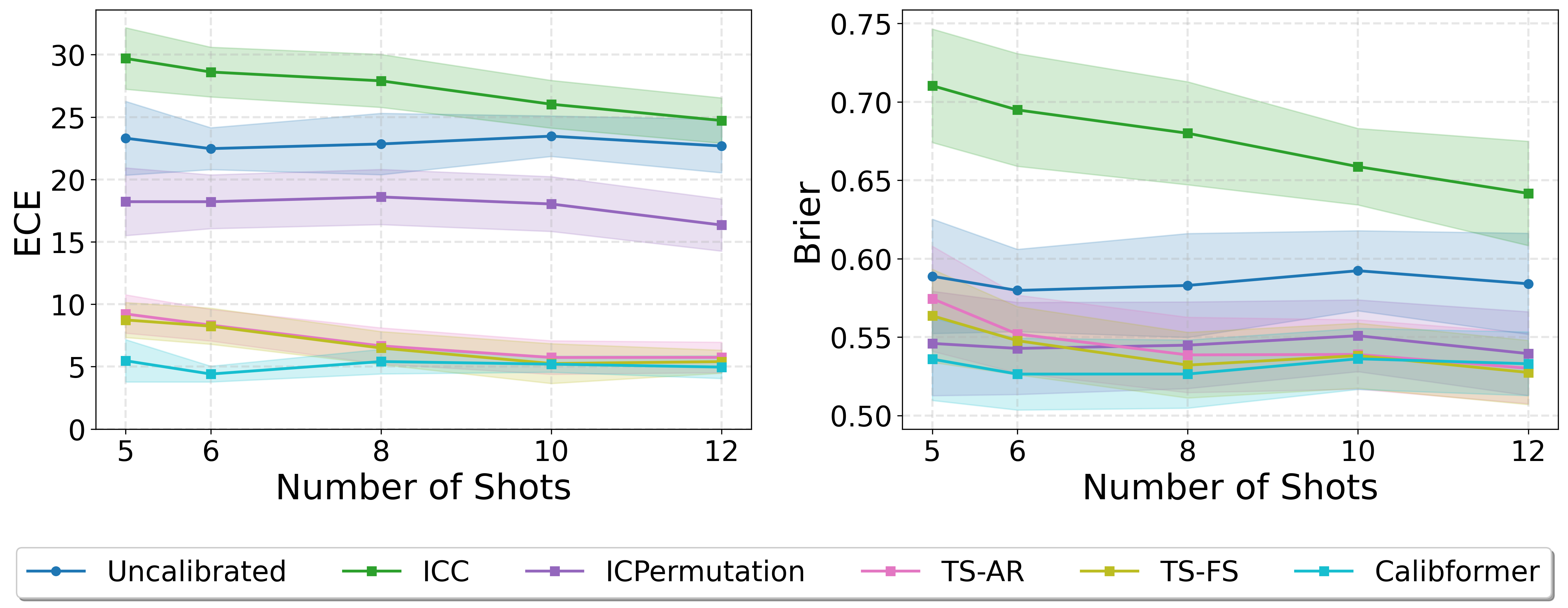}
        \caption{Random sampling}
    \end{subfigure}
    \vspace{0.5em}
    \begin{subfigure}{\textwidth}
        \includegraphics[width=\linewidth]{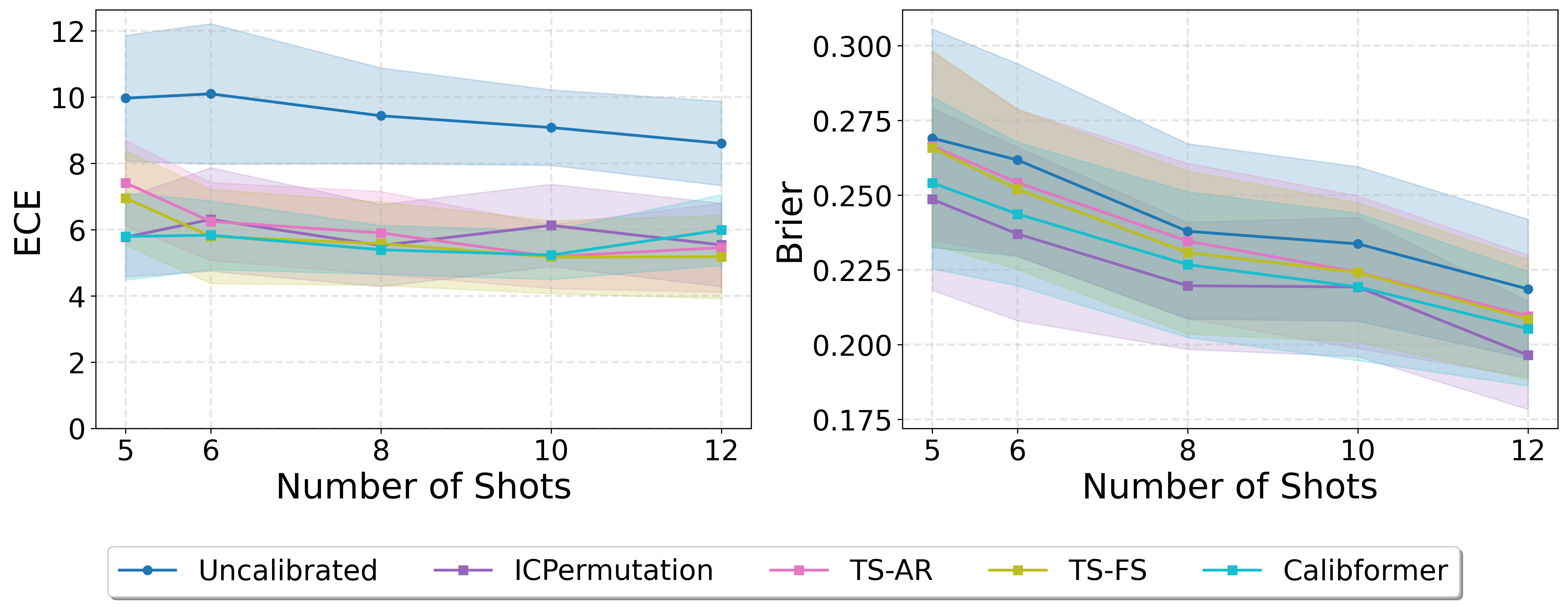}
        \caption{Similarity sampling}
    \end{subfigure}
    \caption{Calibration performance | Phi 4 mini instruct | MetaTool}
    \label{fig:metatool}
\end{figure*}

\begin{figure*}[t]
    \begin{subfigure}{\textwidth}
        \includegraphics[width=\linewidth]{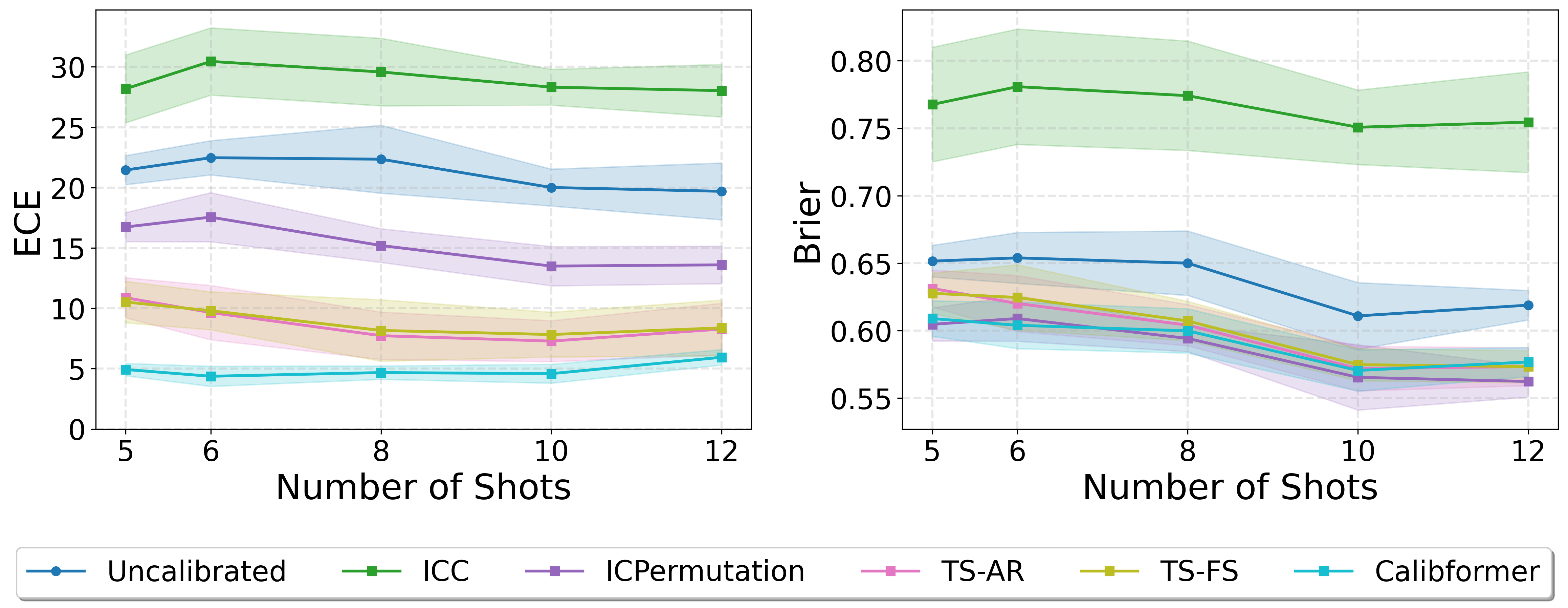}
        \caption{Random sampling}
    \end{subfigure}
    \vspace{0.5em}
    \begin{subfigure}{\textwidth}
        \includegraphics[width=\linewidth]{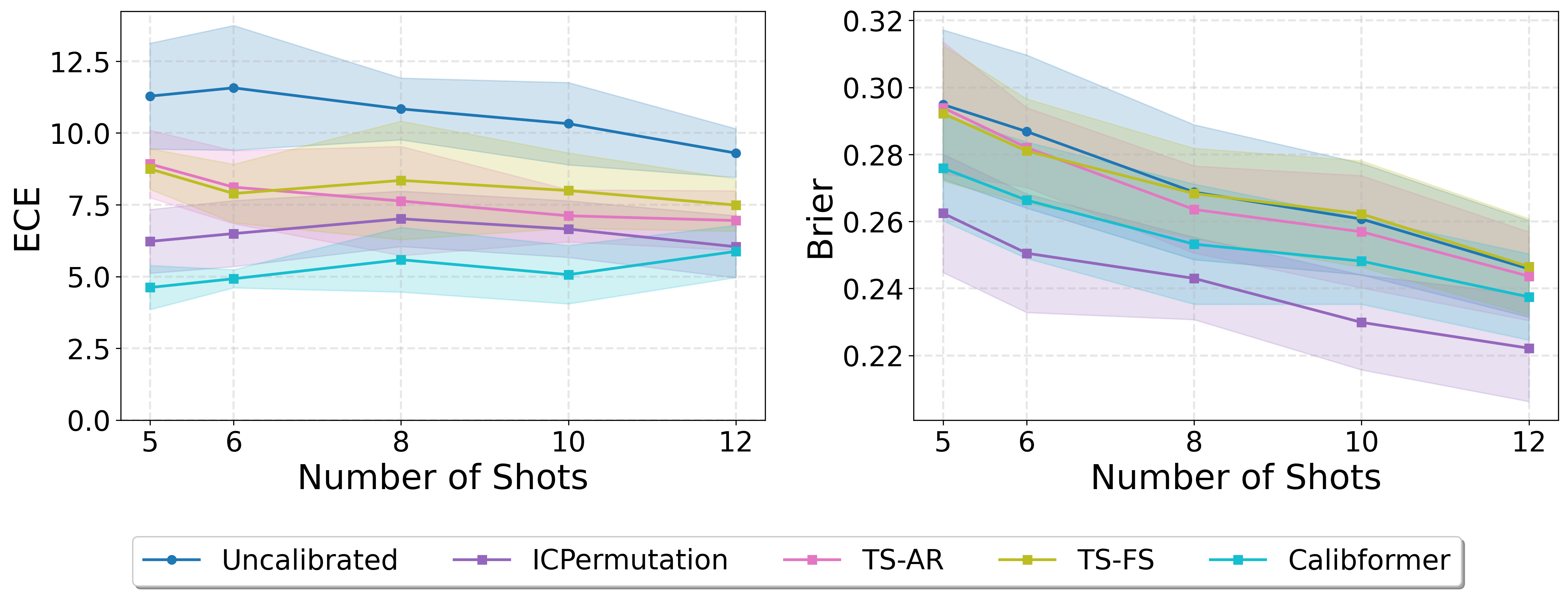}
        \caption{Similarity sampling}
    \end{subfigure}
    \caption{Calibration performance | Phi 4 mini instruct | Massive Intent}
    \label{fig:massive_intent}
\end{figure*}

\begin{figure*}[t]
    \begin{subfigure}{\textwidth}
        \includegraphics[width=\linewidth]{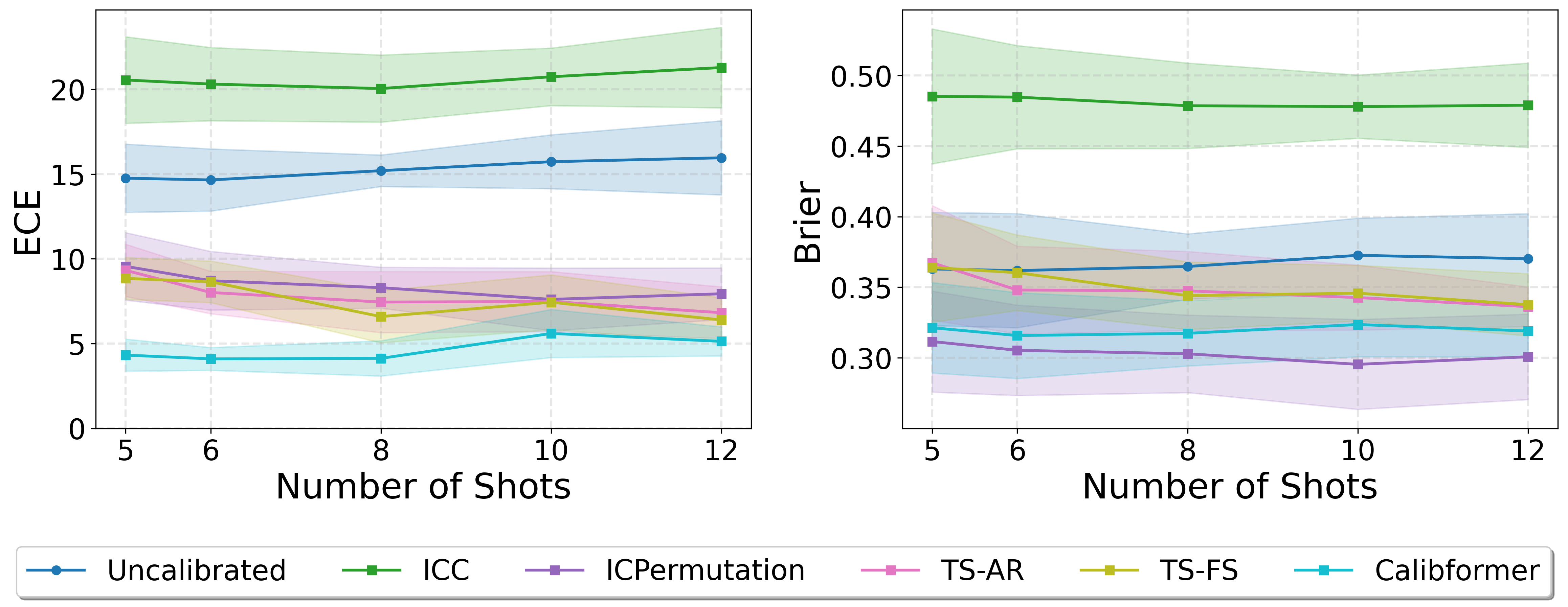}
        \caption{Random sampling}
    \end{subfigure}
    \vspace{0.5em}
    \begin{subfigure}{\textwidth}
        \includegraphics[width=\linewidth]{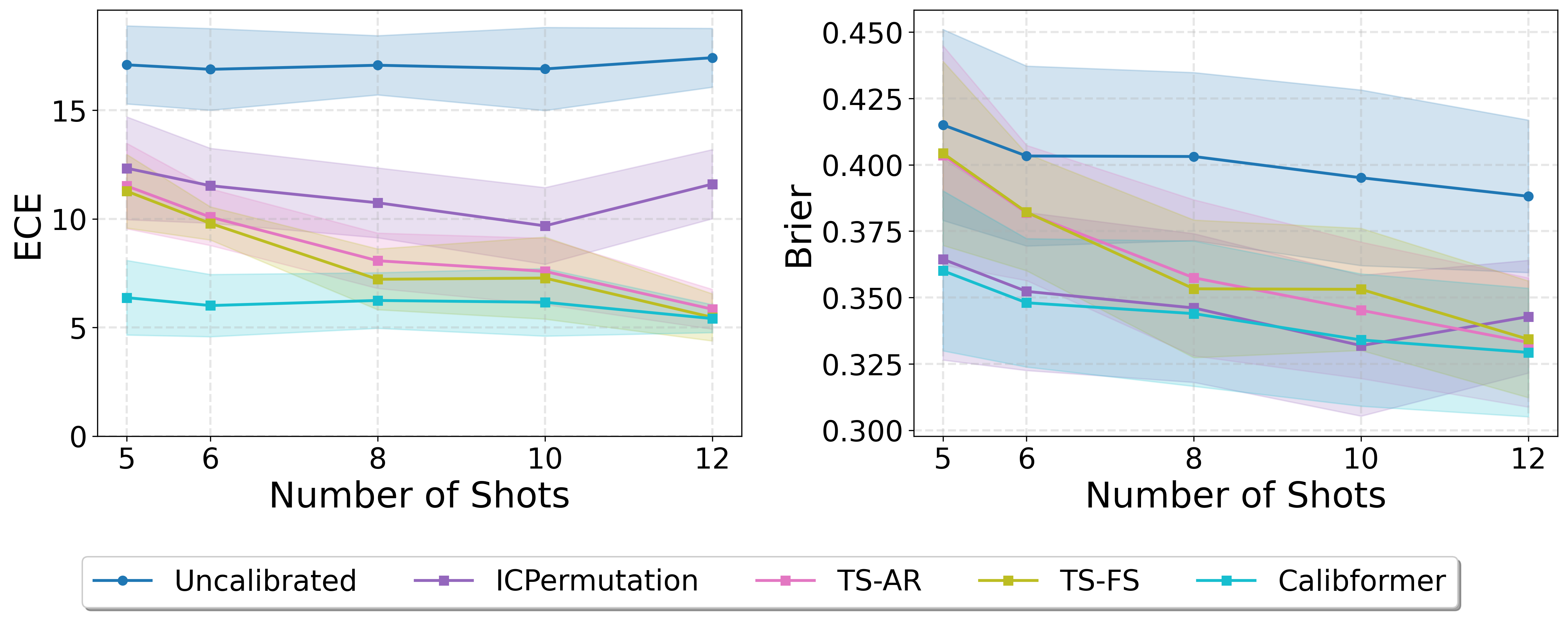}
        \caption{Similarity sampling}
    \end{subfigure}
    \caption{Calibration performance | Phi 4 mini instruct | Amazon Counterfactual(Amazon\_CF)}
    \label{fig:amazon_counterfactual}
\end{figure*}

\begin{figure*}[t]
    \begin{subfigure}{\textwidth}
        \includegraphics[width=\linewidth]{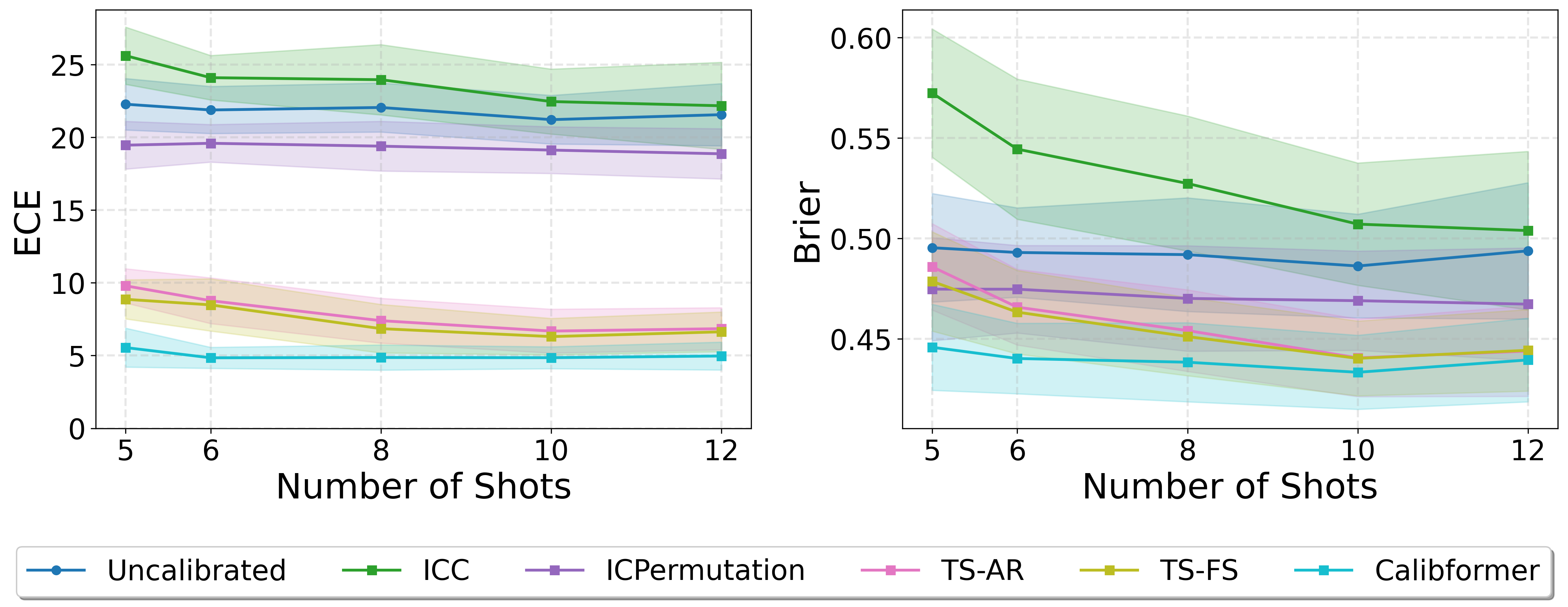}
        \caption{Random sampling}
    \end{subfigure}
    \vspace{0.5em}
    \begin{subfigure}{\textwidth}
        \includegraphics[width=\linewidth]{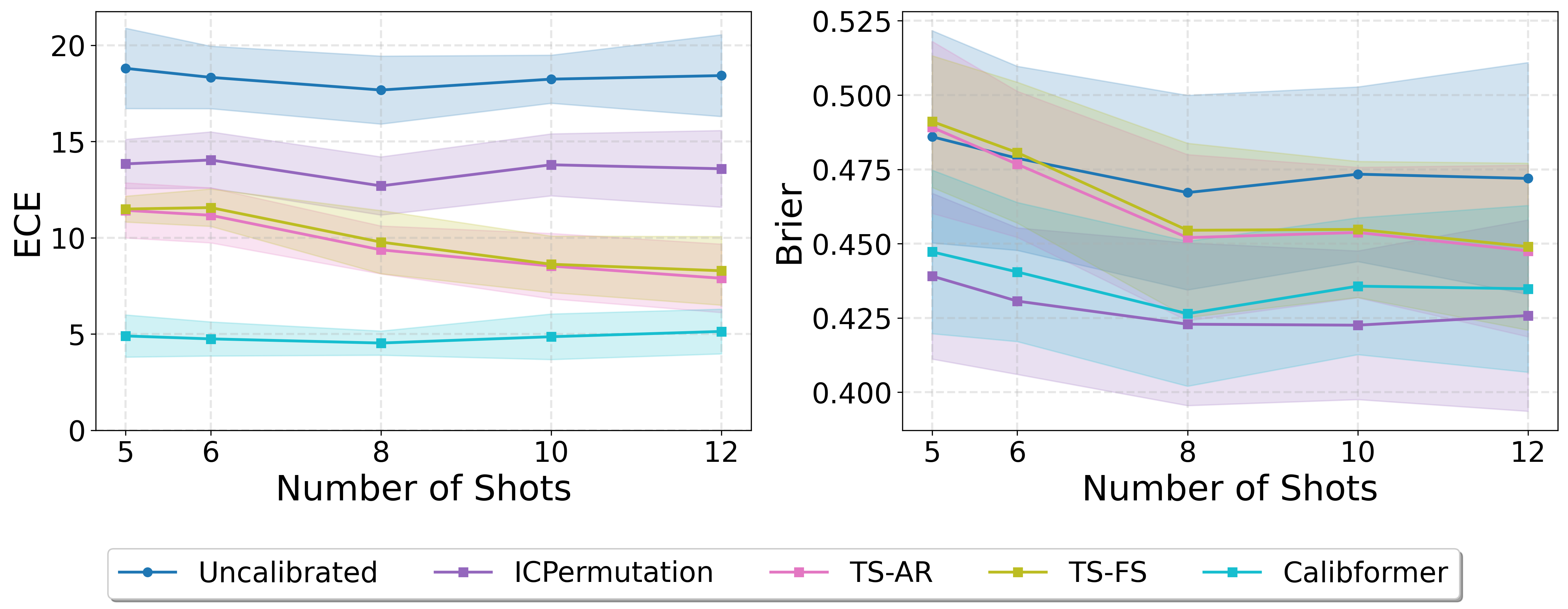}
        \caption{Similarity sampling}
    \end{subfigure}
    \caption{Calibration performance | Phi 4 mini instruct | CommonsenseQA(CS\_QA)}
    \label{fig:commonsense_qa}
\end{figure*}

\section{Additional Implementational Details}
\label{sec:implementation-details}

\subsection{Prompts}
We include in our prompts a base prompt that tells the LLM the label space for proper usage of the LLM's ICL ability. 
Tables \ref{tab:prompts_medqa} and \ref{tab:prompts_counterfactual} show some examples of what 1-shot prompts would look like with \texttt{phi-4-mini-instruct} and the chat template (as provided by the tokenizer) applied.

\begin{table*}[h]
\centering
\small
\begin{tabular}{|>{\centering\arraybackslash}m{1.8cm}|m{13cm}|}
\hline
\textbf{Sampling Strategy} & \textbf{Prompt} \\
\hline
Random Sampling & 
\texttt{<|system|>You are a helpful assistant.<|end|><|user|>Your task is to answer the given medical question by specifying the correct option(A, B, C, D)} \newline
\newline
\texttt{Question: A 43-year-old male presents to a clinic for routine follow-up. He was diagnosed with hepatitis B several months ago. He does not have any complaints about his health, except for poor appetite. The general physical examination is normal. The laboratory investigation reveals mildly elevated aminotransferases. Which of the following findings indicate that the patient has developed a chronic form of his viral infection?} \newline
\texttt{A. HbsAg -, Anti-HbsAg -, Anti-HbcAg IgM +, Anti-HbcAg IgG -, HbeAg -, Anti-HbeAg +} \newline
\texttt{B. HbsAg +, Anti-HbsAg -, Anti-HbcAg IgM -, Anti-HbcAg IgG +, HbeAg +, Anti-HbeAg -} \newline
\texttt{C. HbsAg -, Anti-HbsAg +, Anti-HbcAg IgM -, Anti-HbcAg IgG -, HbeAg -, Anti-HbeAg -} \newline
\texttt{D. HbsAg -, Anti-HbsAg +, Anti-HbcAg IgM -, Anti-HbcAg IgG +, HbeAg -, Anti-HbeAg +} \newline
\texttt{<|end|><|assistant|>Answer: B} \newline
\newline
\texttt{<|end|><|user|>Question: A 62-year-old man presents to the physician because of incomplete healing of a chest wound. He recently had a triple coronary artery bypass graft 3 weeks ago. His past medical history is significant for type 2 diabetes mellitus and hypertension for the past 25 years. Clinical examination shows the presence of wound dehiscence in the lower 3rd of the sternal region. The wound surface shows the presence of dead necrotic tissue with pus. Computed tomography (CT) of the thorax shows a small fluid collection with fat stranding in the perisurgical soft tissues. What is the most appropriate next step in the management of the patient?} \newline
\texttt{A. Surgical debridement} \newline
\texttt{B. Negative pressure wound management} \newline
\texttt{C. Sternal wiring} \newline
\texttt{D. Sternal fixation} \newline
\texttt{<|end|><|assistant|>Answer:} \\
\hline
Similarity Sampling & 
\texttt{<|system|>You are a helpful assistant.<|end|><|user|>Your task is to answer the given medical question by specifying the correct option(A, B, C, D)} \newline
\newline
\texttt{Question: A 78-year-old male presents to the emergency department after passing out. His wife reports that she and the patient were walking their dog when he suddenly lost consciousness. On physical exam, he has a loud crescendo-decrescendo systolic murmur and is subsequently diagnosed with severe aortic stenosis. The patient undergoes open aortic valve replacement and has an uncomplicated postoperative course. His sternal wound drain is pulled for low output on post-operative day three. On post-operative day five, the patient complains of pain during deep inspiration and retrosternal chest pain. His temperature is 101.7°F (38.7°C), blood pressure is 125/81 mmHg, pulse is 104/min, and respirations are 18/min. On physical exam, the patient is tender to palpation around his sternal wound, and there is erythema around the incision without dehiscence. His chest radiograph shows a widened mediastinum with a small pleural effusion on the left. CT angiography shows stranding in the subcutaneous tissue and a fluid collection below the sternum.} \newline
\texttt{Which of the following is the best next step in management?} \newline
\texttt{A. Surgical repair of esophageal perforation} \newline
\texttt{B. Surgical repair of aortic injury} \newline
\texttt{C. Intravenous antibiotics and observation} \newline
\texttt{D. Intravenous antibiotics and debridement of surgical wound} \newline
\texttt{<|end|><|assistant|>Answer: D} \newline
\newline
\texttt{<|end|><|user|>Question: A 62-year-old man presents to the physician because of incomplete healing of a chest wound. He recently had a triple coronary artery bypass graft 3 weeks ago. His past medical history is significant for type 2 diabetes mellitus and hypertension for the past 25 years. Clinical examination shows the presence of wound dehiscence in the lower 3rd of the sternal region. The wound surface shows the presence of dead necrotic tissue with pus. Computed tomography (CT) of the thorax shows a small fluid collection with fat stranding in the perisurgical soft tissues. What is the most appropriate next step in the management of the patient?} \newline
\texttt{A. Surgical debridement} \newline
\texttt{B. Negative pressure wound management} \newline
\texttt{C. Sternal wiring} \newline
\texttt{D. Sternal fixation} \newline
\texttt{<|end|><|assistant|>Answer:} \\
\hline
\end{tabular}
\caption{MedQA 1-shot prompt with chat template applied}
\label{tab:prompts_medqa}
\end{table*}

\begin{table*}[t]
\centering
\small
\begin{tabular}{|>{\centering\arraybackslash}m{3cm}|m{12.5cm}|}
\hline
\textbf{Sampling Strategy} & \textbf{Prompt} \\
\hline
Random Sampling & 
\texttt{<|system|>You are a helpful assistant.<|end|><|user|>Your task is to classify the given Amazon customer review as one of the following: not-counterfactual, counterfactual} \newline\newline
\texttt{User Review: The women suffered badly - the rejection and the unfaithfulness must have been hard to bear.} \newline
\texttt{<|end|><|assistant|>Category: not-counterfactual} \newline
\newline
\texttt{<|end|><|user|>User Review: But even without the seam, these are luscious stockings.} \newline
\texttt{<|end|><|assistant|>Category: not-counterfactual} \newline
\newline
\texttt{<|end|><|user|>User Review: "The lid latches could be slightly better (I've damaged a couple with repeated open/close cycles and heavy container stacking) but I have yet to find the \textbackslash\textbackslash\textbackslash\textbackslash perfect\textbackslash\textbackslash\textbackslash\textbackslash" container/lid combination among many product types and these certainly fit nearly all my requirements.""} \newline
\texttt{<|end|><|assistant|>Category:} \\
\hline
Similarity Sampling & 
\texttt{<|system|>You are a helpful assistant.<|end|><|user|>Your task is to classify the given Amazon customer review as one of the following: not-counterfactual, counterfactual} \newline\newline
\texttt{User Review: I am not sure the rubbermaid can handle more weight and I am a bit affraid to stack more stuffs on the top as one of the rail lightly bended when it arrived.} \newline
\texttt{<|end|><|assistant|>Category: not-counterfactual} \newline
\newline
\texttt{<|end|><|user|>User Review: They arrived with the tips in a small block of styrofoam-like material - which works well - but it would be great if there was some sort of cap (nothing fancy - even just a tiny plastic tip or something) since the block makes the very clumsy to store.} \newline
\texttt{<|end|><|assistant|>Category: counterfactual} \newline
\newline
\texttt{<|end|><|user|>User Review: "The lid latches could be slightly better (I've damaged a couple with repeated open/close cycles and heavy container stacking) but I have yet to find the \textbackslash\textbackslash\textbackslash\textbackslash perfect\textbackslash\textbackslash\textbackslash\textbackslash" container/lid combination among many product types and these certainly fit nearly all my requirements.""} \newline
\texttt{<|end|><|assistant|>Category:} \\
\hline
\end{tabular}
\caption{Amazon Counterfactual 2-shot prompts with chat template applied}
\label{tab:prompts_counterfactual}
\end{table*}

\subsection{LLM inference}
vLLM \cite{vllm} has been a popular inference framework for LLM serving that provides performance speed up over Huggingface through optimized CUDA kernels, prefix caching, PagedAttention, efficient batching and so forth. We use vLLM 0.19.0 installed with pytorch 2.10 and cuda 12.8 for all our open-weights LLM experiments. Each call to the LLM is made with some non default arguments in \texttt{SamplingParams} of which the following may be of interest:

\begin{itemize}
    \item \texttt{max\_tokens}=1
    \item \texttt{logprobs}=\texttt{len(label\_tokens\_ids)}
    \item \texttt{skip\_special\_tokens}=\texttt{True}
    \item \texttt{allowed\_token\_ids}=\texttt{label\_token\_ids}
\end{itemize}

It has been reported\footnote{ \hyperref[]{https://github.com/vllm-project/vllm/issues/14300}} that there is an $O(N)$ time complexity for inference time w.r.t the number of \texttt{logprobs} requested. The \texttt{allowed\_token\_ids} parameter out of the box only applies to the actual generated text and not the \texttt{raw\_logprobs} returned by vLLM for each prompt/generated token, so we added a two line modification to the source code (v1/sampler.py) to also apply the parameter to \texttt{raw\_logprobs}, masking the non-label tokens in the raw logits to $-inf$ which makes sure the top \texttt{len(label\_tokens\_ids)} tokens are the label tokens. This minimal change completely circumvents the issue and allows efficient inference. 

      

Transformers naturally provide token logits at all positions, including the prompt tokens, and this makes getting the predictions over the in context examples possible without any extra LLM calls. However, this is not viable when using black box LLMs that do not return prompt logprobs, and additional $k$ calls must be made to calibrate a k-shot prompt in such cases, removing the computational edge over the other approaches. Additional calls must also be made when the labels are not distinguishable by their first tokens, although most tasks tend to have simple label spaces.  

With regards to vLLM, while returning prompt logprobs is supported and improved by the simple source code modification that lets us work exclusively with label tokens, the prefix caching feature is disabled when requesting prompt logprobs due to absence of optimization demand in this setting \footnote{\href{https://github.com/vllm-project/vllm/issues/13414}{github.com/vllm-project/vllm/issues/13414}}. However, our goal is to provide a proof of concept and source code modifications/feature requests can be made if needed. We also provide a Huggingface implementation for LLM inference which supports efficient prompt logprobs extraction.

We ran all our training and evaluation experiments on Nvidia B200 GPUs. For the gemini experiments, we make calls to \texttt{gemini-2.5-flash-preview-09-2025}, requesting the top 20 (maximum allowed) logprobs and generating a single output token per call.

\subsection{Ablation Statistical Significance}

To verify that the performance degradations observed in Tables~\ref{tab:ablation} and ~\ref{tab:calib_choice} reflect genuine differences rather than random variation across training runs, we 
conduct paired statistical significance tests across 30 random seeds. For each 
ablation, we compute the per-seed difference in ECE and Brier score relative to 
\sysname\, and test the one-sided hypothesis that the ablation performs 
worse than the baseline (i.e., produces higher ECE or Brier score). Since the 
direction of degradation is pre-specified by construction---each ablation removes 
or degrades a deliberately designed component---a one-sided test is appropriate. 
We first apply the Shapiro--Wilk normality test to the paired differences; if 
normality holds ($p \geq 0.05$), we use a paired $t$-test, otherwise we fall back 
to the Wilcoxon signed-rank test. This fallback was triggered for the ECE 
differences of Half Train Data and Temperature Output.

All ablations produce statistically significant degradation in calibration error 
($p < 0.05$). The statistically weakest degradation happens when reducing feature set to just confidence and correctness ($p = 0.03$ for ECE), which suggests that the confidence and correctness alone carry most of the signal; although the corresponding Brier degradation is still highly statistically significant ($p < 0.001$) and the gaps widen in lower-shot and higher class count regimes. The strongest degradations are observed with permuted embeddings and Temperature Output, consistent with their mean differences in Table~\ref{tab:ablation}. These results confirm that each design choice provides statistically significant gains to \sysname\'s calibration performance.

\phantomsection\label{note:mlp-logistic}
Note that in Table~\ref{tab:calib_choice}, the MLP and Logistic regression models were trained on 7 dimensional prediction summary statistics, which are a concatenation of two features at the query position (top 2 probabilities) with the vector of five features averaged over the in-context examples. We use the five features described in Section~\ref{sec:feat}, but without the causal shifting of correctness indicator and ground truth probability related features (features 3-5). 

\section{LLM Usage Declaration}
We acknowledge that LLMs were used in the refinement of parts of the article and in the implementation of the code.

\end{document}